\documentclass[journal]{IEEEtran}

\usepackage[nocompress]{cite}
\usepackage{booktabs} 
\usepackage{xcolor}
\usepackage{soul}
\usepackage{epsfig}
\usepackage{graphicx}
\usepackage{amsmath}
\usepackage{amssymb}
\usepackage{courier}
\usepackage{helvet}
\usepackage{courier}
\usepackage{algorithm}
\usepackage{algorithmic}
\usepackage{multirow}
\usepackage{url}
\usepackage{booktabs}
\usepackage{array}
\usepackage{paralist,algorithmic,algorithm}
\usepackage{balance}
\usepackage[numbers]{natbib}

\newcommand{\p}{{\bf p}}
\newcommand{\w}{{\bf w}}

\newcommand{\vvv}{{\bf v}}

\newcommand{\y}{{\bf y}}

\newcommand{\D}{\mathcal{D}}

\newcommand{\R}{\mathcal{R}}

\newcommand{\EE}{\mathbb{E}}

\newtheorem{defn}{Definition}

\ifCLASSINFOpdf
\else
\fi
\begin{document}
%
\title{Exploiting Cross-Modal Prediction and Relation Consistency for Semi-Supervised Image Captioning}
%
%
%

\author{Yang Yang,~\IEEEmembership{}
	Hongchen Wei,~\IEEEmembership{}
	Hengshu Zhu~\IEEEmembership{Senior Member, IEEE},
	Dianhai Yu,~\IEEEmembership{}
	Hui Xiong,~\IEEEmembership{Fellow, IEEE}
	and Jian Yang~\IEEEmembership{Member, IEEE}
	\IEEEcompsocitemizethanks{\IEEEcompsocthanksitem Yang Yang, Hongchen Wei and Jian Yang are with the Nanjing University of Science and Technology, Nanjing 210094, China.\protect\\
		E-mail: {yyang,weihc,csjyang}@njust.edu.cn}
	\IEEEcompsocitemizethanks{\IEEEcompsocthanksitem Hengshu Zhu, Dianhai Yu are with Baidu Inc, Beijing 100000, China.\protect\\
		E-mail:{zhuhengshu,yudianhai}@baidu.com}
	\IEEEcompsocitemizethanks{\IEEEcompsocthanksitem  Hui Xiong is with the Management Science and Information Systems Department, Rutgers Business School, Rutgers University, Newark, NJ 07102, USA.\protect\\
		E-mail: hxiong@rutgers.edu}
	\thanks{Yang Yang, Hongchen Wei and Jian Yang are with PCA Lab, Key Lab of Intelligent Perception and Systems for High-Dimensional Information of Ministry of Education, and Jiangsu Key Lab of Image and Video Understanding for Social Security, School of Computer Science and Engineering, Nanjing University of Science and Technology. Yang Yang is the corresponding author.}}

%
%

\markboth{Journal of \LaTeX\ Class Files,~Vol.~14, No.~8, August~2015}%
{Shell \MakeLowercase{\textit{et al.}}: Bare Demo of IEEEtran.cls for IEEE Journals}
%



\maketitle

\begin{abstract}
The task of image captioning aims to generate captions directly from images via the automatically learned cross-modal generator. To build a well-performing generator, existing approaches usually need a large number of described images (i.e., supervised image-sentence pairs), which requires a huge effects on manual labeling. However, in real-world applications, a more general scenario is that we only have limited amount of described images and a large number of undescribed images. Therefore, a resulting challenge is how to effectively combine the undescribed images into the learning of cross-modal generator (i.e., \emph{semi-supervised image captioning}). To solve this problem, we propose a novel  image captioning method by exploiting the Cross-modal Prediction and Relation Consistency (CPRC), which aims to utilize the raw image input to constrain the generated sentence in the commonly semantic space. In detail, considering that the heterogeneous gap between modalities always leads to the supervision difficulty of using the global embedding directly, CPRC turns to transform both the raw image and corresponding generated sentence into the shared semantic space, and measure the generated sentence from two aspects: 1) Prediction consistency. CPRC utilizes the prediction of raw image as soft label to distill useful supervision for the generated sentence, rather than employing the traditional pseudo labeling; 2) Relation consistency. CPRC develops a novel relation consistency between augmented images and corresponding generated sentences to retain the important relational knowledge. In result, CPRC supervises the generated sentence from both the informativeness and representativeness perspectives, and can reasonably use the undescribed images to learn a more effective generator under the semi-supervised scenario. The experiments show that our method outperforms state-of-the-art comparison methods on the MS-COCO ``Karpathy'' offline test split under complex non-parallel scenarios, e.g., CPRC achieves at least 6$\%$ improvements on CIDEr-D score considering different losses.
\end{abstract}

\begin{IEEEkeywords}
Cross-modal Learning, Image Captioning, Semi-Supervised Learning, Relation Consistency
\end{IEEEkeywords}

%
\IEEEpeerreviewmaketitle

\begin{figure}[htb]\centering
	\centering
	\includegraphics[width = 85mm]{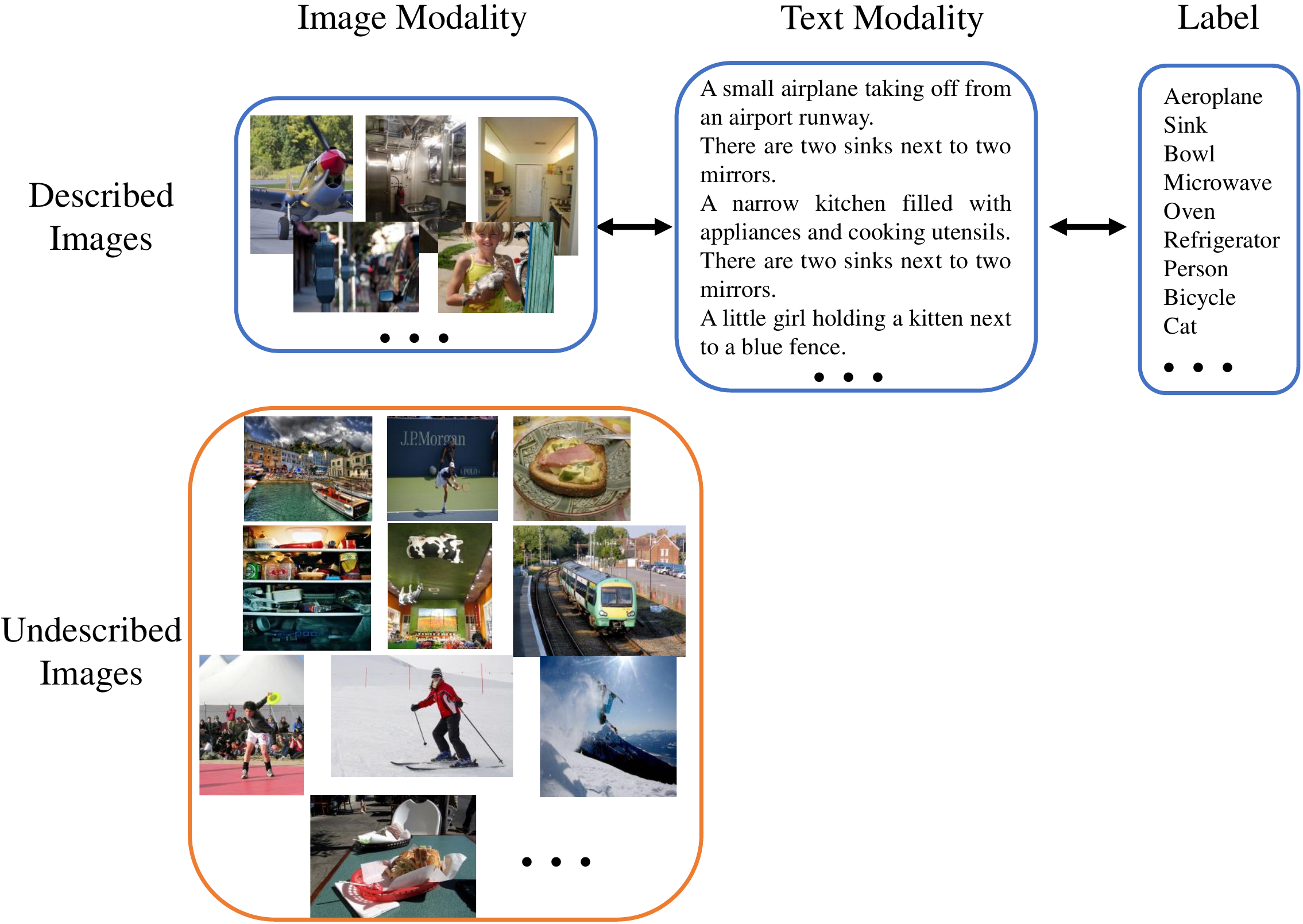}\\
	\caption{Semi-supervised image-sentence pairs, which include limited described images and a huge number of undescribed images. It is notable that we have two types of supervision: text and label ground-truths. Described images have all the supervisions, whereas the undescribed images do not have any kind of supervision information here.}\label{fig:data}
\end{figure}
\section{Introduction}
In real-world applications, object can always be represented by multiple source information, i.e., multiple modalities~\cite{BaltrusaitisAM19,DebieRFBKAGA21}. For example, the news always contains image and text information, the video can be divided into image, audio and text information. Along this line, the study of cross-modal learning has emerged for bridging the connections among different modalities, so as to better perform downstream tasks, in which the image captioning is one of the important research directions. Specifically, image captioning aims to automatically generate natural language descriptions for images, and has emerged as a prominent research problem in both academia and industry~\cite{KarpathyL15,XuBKCCSZB15,SammaniE19,BinYSXSL19}. For example, we can automatically broadcast road conditions by learning visual images to assist driving, and can also help visually impaired users to read more conveniently. In fact, the challenge of image captioning is to learn the generator between two heterogeneous modalities (i.e., the image and text modalities), which needs to recognize salient objects in an image using computer vision techniques and generate coherent descriptions using natural language processing.

To solve this problem, researchers firstly explored the neural encoder-decoder models~\cite{KarpathyL15,YangYWCS16}, which are composed of a CNN encoder and a LSTM (or Transformer) decoder. In detail, these methods firstly encode the image into a set of feature vectors using a CNN based model, each segmentation captures semantic information about an image region, then decode these feature vectors to words sequentially via a LSTM-based or Transformer-based network. Furthermore, \cite{XuBKCCSZB15,LuXPS17,HuangWCW19} adopted the single or hierarchical attention mechanism that enables the model to focus on particular image regions during decoding process. To mitigate the incorrect or repetitive content, several researches consider to edit inputs independently from the problem of generating inputs~\cite{HashimotoGOL18,SammaniE19}. However, note that all these methods require full image-sentence pairs in advance, i.e., all the images need to be described manually, which is hard to accomplish in real-world applications. A more general scenario is shown in Figure \ref{fig:data}, we have limited described images with corresponding label ground-truths, and a large number of undescribed images. Therefore, a resulting challenge is the ``\emph{Semi-Supervised Image Captioning}'', which aims to conduct the captioning task by reasonably using the huge number of undescribed images and limited supervised data. 

The key difficulty of semi-supervised image captioning is to design the pseudo supervision for the generated sentences. Actually, there have been some preliminary attempts recently. For example, \cite{Feng00L19a,GuJCZYW19} proposed unsupervised captioning methods, which combined the adversarial learning~\cite{GoodfellowPMXWOCB14} with traditional encoder-decoder models to evaluate the quality of generated sentences. In detail, based on the traditional encoder-decoder models, these approaches employ adversarial training to generate sentences such that they are indistinguishable from the sentences within  auxiliary corpus. In order to ensure that the generated captions contain the visual concepts, they additionally distill the knowledge provided by a visual concept detector into the image captioning model. However, the domain discriminator and visual concept distiller do not fundamentally evaluate the matching degree and structural rationality of the generated sentence, so the captioning performance is poor. As for semi-supervised image captioning, a straightforward way is directly utilizing the undescribed images together with their machine-generated sentences~\cite{MithunPPR18,HuangKLCH19} as the pseudo image-sentence pair, to fine-tune the model. However, limited amount of parallel data can hardly establish a proper initial generator to generate precisely pseudo descriptions, which may have negative affection to the fine-tuning of visual-semantic mapping function. 


To circumvent this issue, we attempt to utilize the raw image as pseudo supervision. However, heterogeneous gap between modalities always leads the supervision difficulty if we directly constrain the consistency between global embedding of image and sentence. Thereby, we switch to use the broader and effective semantic prediction information, rather than directly utilize the embedding, and introduce a novel approach, dubbed \emph{semi-supervised image captioning by exploiting the Cross-modal Prediction and Relation Consistency} (CPRC). In detail, there are two common approaches for traditional semi-supervised learning: 1) Pseudo labeling: it minimizes the entropy of unlabeled data using predictions; 2) Consistency regularization: it transforms the unlabeled raw images using data augmentation techniques, then constrains the consistency of transformed instances' outputs. Different form these two techniques, we design cross-modal prediction and relation consistency by comprehensively considering the informativeness and representativeness: 1) Prediction consistency: we utilize the soft label of image to distill effective supervision for generated sentence; 2) Relation consistency: we work on encouraging the generated sentences to have similar relational distribution to the augmented image inputs. The central tenet is that the relations of learned representations can better present the consistency than individual data instance~\cite{ParkKLC19}. Consequently, CPRC can effectively qualify the generated sentences from both the prediction confidence and distribution alignment perspectives, thereby to learn more robust mapping function. \emph{Note that CPRC can be implemented with any current captioning model}, and we adopt several typical approaches for verification~\cite{RennieMMRG17,ZhouWLHZ20}. Source code is available at \url{https://github.com/njustkmg/CPRC}. 

In summary, the contributions in this paper can be summarized as follows:
\begin{itemize}
	\item We propose a novel semi-supervised image captioning framework for processing undescribed images, which is universal for any captioning model;
	\item We design the cross-modal prediction and relation consistency to measure the undescribed images, which maps the raw image and corresponding generated sentence into the shared semantic space, and supervises the generated sentence by distilling the soft label from image prediction and constraining the cross-modal relational consistency;
	\item In experiments, our approach improves the performance under semi-supervised scenario, which validates that knowledge hidden in the content and relation is effective for enhancing the generator.
\end{itemize}

\section{Related Work}
\subsection{Image Captioning}
Image captioning approaches can be roughly divided into three categories: 1) Template based methods, which generate slotted captioning templates manually, and then utilize the detected keywords to fill the templates~\cite{YaoYLLZ10}, but their expressive
power is limited because of the need for designing templates manually; 2) Encoder-decoder based methods, which are inspired by the neural machine translation~\cite{ChoMGBBSB14}. For example, \cite{VinyalsTBE15} proposed an end-to-end framework with a CNN encoding the image to feature vector and a LSTM decoding to caption; \cite{HuangWCW19} added an attention-on-attention module after both the LSTM and the attention mechanism, which can measure the relevance between attention result and query; and 3) Editing based methods, which consider editing inputs independent from generating inputs. For example, \cite{HashimotoGOL18} learned a retrieval model that embeds the input in a task-dependent way for code generation; \cite{SammaniE19} introduced a framework that learns to modify existing captions from a given framework by modeling the residual information. However, all these methods need huge amount of supervised image-sentence pairs for training, whereas the scenario with large amount of undescribed images is more general in real applications. To handle the undescribed images, several attempts propose unsupervised image captioning approaches. \cite{Feng00L19a} distilled the knowledge in visual concept detector into the captioning model to recognize the visual concepts, and adopted sentence corpus to teach the captioning model; \cite{GuJCZYW19} developed an unsupervised feature alignment method with adversarial learning that maps the scene graph features from the image to sentence modality. Nevertheless, these methods mainly depend on employing the domain discriminator for learning plausible sentences, that are difficult for generating matched sentences. On the other hand, considering the semi-supervised image captioning, \cite{MithunPPR18,HuangKLCH19} proposed to extract regional semantics from un-annotated images as additional weak supervision to learn visual-semantic embeddings. However, the generated pseudo sentences are always unqualified to fine-tune the generator in real experiments.

\begin{figure*}[t]\centering
	\centering
	\includegraphics[width = 170mm]{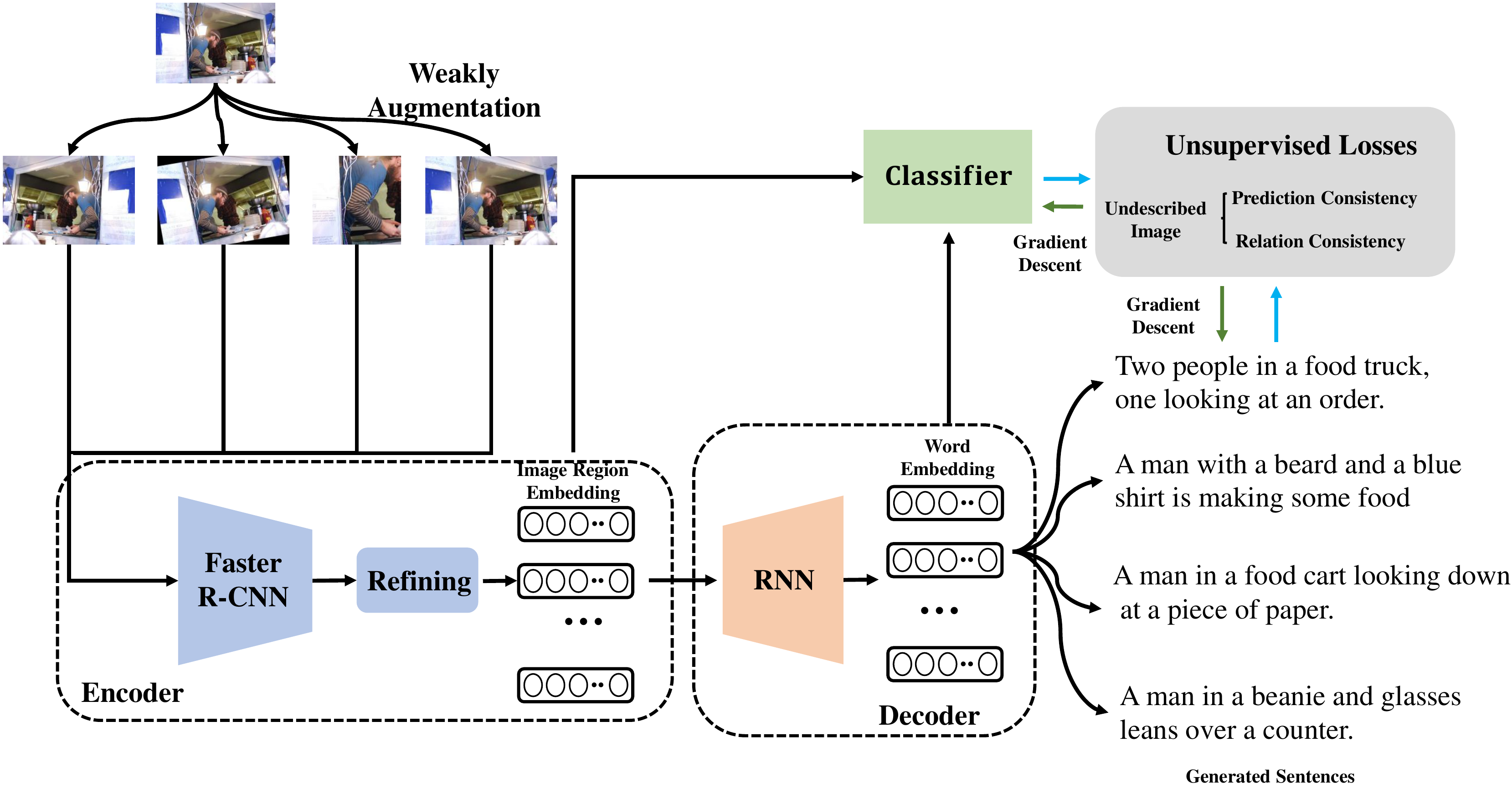}\\
	\caption{Diagram of the proposed unsupervised loss. For example, three weakly-augmented images and the raw image are fed into the encoder to obtain image region embeddings, then four corresponding sentences are generated by the decoder. Then, the embeddings of image inputs and generated sentences are fed into the shared classifier to obtain the predictions. The model is trained by considering two objectives: 1) \emph{supervised loss} includes the generation cross-entropy and prediction cross-entropy for described images. In detail, generation cross-entropy measures the quality of generated sentence sequence, and prediction cross-entropy considers the multi-label prediction loss of generated sentence. 2) \emph{unsupervised loss} includes the prediction consistency and relation consistency for undescribed images. In detail, prediction consistency utilizes the image's prediction as pseudo labels for corresponding generated sentence, and relation consistency consist the generated sentences' distribution with image inputs' distribution.}\label{fig:framework}
\end{figure*}

\subsection{Semi-Supervised Learning}
Recently, deep networks achieve strong performance by supervised learning, which requires a large number of labeled data. However, it comes at a significant cost when labeling by human labor, especially by domain experts. To this end, 
semi-supervised learning, which concerns combining supervised and unsupervised learning techniques to perform certain learning tasks and permits harnessing the large amounts of unlabeled data in combination with typically smaller sets of labeled data, attracts more and more attention. Existing semi-supervised learning mainly considers two aspects: 1) Self-training~\cite{GrandvaletB04}. The generality of self-training is to use a model’s predictions to obtain artificial labels for unlabeled data. A specific variant is the pseudo-labeling, which converts the model predictions of unlabeled data to hard labels for calculating the cross-entropy. Besides, pseudo-labeling is often used along with a confidence thresholding that retains sufficiently confident unlabeled instances. In result, pseudo-labeling results in entropy minimization, which has been used as a component for many semi-supervised algorithms, and has been validated to produce better results~\cite{ArazoOAOM20}. 2) Consistency regularization~\cite{BachmanAP14}. Early extensions include exponential moving average of model parameters~\cite{TarvainenV17} or using previous model checkpoints~\cite{LaineA17}. Recently, data augmentation, which integrates these techniques into the self-training framework, has shown better results~\cite{XieDHL020,BerthelotCCKSZR20}. A mainstream technology is to produce random perturbations with data augmentation~\cite{FrenchMF18}, then enforce consistency between the augmentations. For example, \cite{XieDHL020} proposed unsupervised data augmentation with distribution alignment and augmentation anchoring, which encourages each output to be close to the weakly-augmented version of the same input; \cite{BerthelotCCKSZR20} used a weakly-augmented example to generate an artificial label and enforce consistency against strongly-augmented example. Furthermore, \cite{Kihyuk2020} combined the pseudo labeling and consistency regularization into a unified framework, which generates pseudo-labels using the model’s predictions on weakly-augmented unlabeled images, and constrain the prediction consistency between weakly-augmented and strongly-augmented version. Note that the targets in previous semi-supervised methods are uniform and simple, i.e., the label ground-truths. However, cross-modal semi-supervised learning is more complicated, e.g., each image has the corresponding sentence and label ground-truth. It is more difficult for building cross-modal generator than single modal classifier with limited supervised data, thereby it may causes noise accumulation if we directly employ the traditional semi-supervised technique for the generated sentences.

The remainder of this paper is organized as follows. Section \ref{sec:s1} presents the proposed method, including the model, solution, and extension. Section \ref{sec:s2} shows the experimental results on COCO dataset, under different semi-supervised setting. Section \ref{sec:s3} concludes this paper.

\section{Proposed Method}\label{sec:s1}
\subsection{Notations}
Without any loss of generality, we define the semi-supervised image-sentence set as: $\D = \{\{\vvv_i,\w_i,\y_i\}_{i=1}^{N_l}, \{\vvv_j\}_{j=1}^{N_u}\}$, where $\vvv_i \in \R^{d_v}$ denotes the $i-$th image instance, $\w_i \in \R^{d_w}$ represents the aligned sentence instance, $\y_i \in \R^{C}$ denotes the instance label, $\y_{i,k} = 1$ if $i-$th instance belongs to the $k-$th label, otherwise is $0$. $\vvv_j$ is the $j-$th undescribed image. $N_l$ and $N_u$ ($N_l \ll N_u$) are the number of described and undescribed instances, respectively. 
\begin{defn}\label{def:d1}
	\textbf{Semi-Supervised Image Captioning.} Given limited parallel image-sentence pairs $\{\vvv_i,\w_i,\y_i\}_{i=1}^{N_p}$ and a huge number of undescribed images $\{\vvv_j\}_{j=1}^{N_u}$, we aim to construct a generator $G$ for image captioning by reliably utilizing the undescribed images.
\end{defn}
\subsection{The Framework}
It is notable that CPRC focuses on employing the undescribed images, and is a general semi-supervised framework. Thereby the image-sentence generator, i.e., $G: \vvv \rightarrow \w $, can be represented as any state-of-the-art captioning model. In this paper, considering the effectiveness and reproducibility, we adopt the attention model, i.e., AoANet~\cite{HuangWCW19}, for $G$ as base model. In detail, the $G$ is an encoder-decoder based captioning model, which always includes an image encoder and a text decoder. Given an image $\vvv$, the target of $G$ is to generate a natural language sentence $\hat{\w}$ describing the image. The formulation can be represented as: $\hat{\w} = D(E(\vvv))$, where the encoder $E$ is usually a convolutional neural network~\cite{HeZRS16,RenHG017} for extracting the embedding of raw image input. Note that $E$ usually includes refining module such as attention mechanism~\cite{BahdanauCB14}, which aims to refine the visual embedding for suiting the language generation dynamically. The decoder $D$ is widely used RNN-based model for the sequence prediction $\hat{\w}_i$.

The learning process of CPRC is shown in Figure \ref{fig:framework}. Specifically, CPRC firstly samples a mini-batch of images from the dataset $\D$ (including described and undescribed images), and adopts the data augmentation techniques for each undescribed image (i.e., each image has $K$ variants). Then we can acquire the generated sentences for both augmented images and the raw image using the $G$, and compute the predictions for image inputs and generated sentences using the shared prediction classifier $f$. The model is trained through two main objects: 1) \emph{supervised loss}, which is designed for described images, i.e., supervised image-sentence pairs. In detail, supervised loss considers both the label and sentence predictions, including: a) \emph{generation cross-entropy}, which employs the cross-entropy loss or reinforcement learning based reward~\cite{RennieMMRG17} for generated sentence sequence and ground-truth sentence. b) \emph{prediction cross-entropy}, which calculates the multi-label loss between image/sentence's prediction and label ground-truth. 2) \emph{unsupervised loss}, which is designed for undescribed images. In detail, unsupervised loss considers both the informativeness and representativeness: a) \emph{prediction consistency}, which uses the image's prediction as pseudo label to distill effective information for generated sentence, so as to measure the instance's informativeness; b) \emph{relation consistency}, which adopts the relational structure of the augmented images as the supervision distribution for generated sentences, so as to measure the instance's representativeness. Therefore, in addition to the traditional loss for described images, we constrain the sentences generated from undescribed images by comprehensively using the raw image inputs as pseudo labels. The details are described as follows.

\subsection{Supervised Loss}
\subsubsection{Generation Loss}
Given an image $\vvv$, the decoder (Figure \ref{fig:framework}) generate a sequence of sentence $\hat{\w} = \{w_{1}, w_{2}, \cdots , w_{T}\}$ describing the image, $T$ is the length of sentence. Then, we can minimize the cross-entropy loss (i.e., $\ell_{XE}$) or maximize a reinforcement learning based reward~\cite{RennieMMRG17} (i.e., $\ell_{RL}$), according to ground truth caption $\w$: 
\begin{equation}\label{eq:e0}
\begin{split}
\ell_{XE}& = - \sum_{t = 1}^T \log p(\w_{t} | \w_{{1:t-1}}), \\
\ell_{RL}&= - \EE_{\w_{1:T}}~ p [r(\w_{1:T})], \\
\end{split}
\end{equation} 
where $\w_{1:T}$ denotes the target ground truth sequence, $p(\cdot)$ is the prediction probability. the reward $r(\cdot)$ is a sentence-level metric for the sampled sentence and the ground-truth, which always uses the score of some metric (e.g. CIDEr-D~\cite{VedantamZP15}). In detail, as introduced in~\cite{RennieMMRG17}, captioning approaches traditionally train the models using the cross entropy loss. On the other hand, to directly optimize NLP metrics and address the exposure bias issue. \cite{RennieMMRG17} casts the generative models in the Reinforcement Learning terminology as~\cite{RanzatoCAZ15}. In detail, traditional decoder (i.e., LSTM) can be viewed as an “agent” that interacts with the “environment” (i.e., words and image features). The parameters of the network define a policy, that results in an “action” (i.e.,  the prediction of the next word). After each action, the agent updates its internal “state” (i.e., parameters of the LSTM, attention weights etc). Upon generating the end-of-sequence (EOS) token, the agent observes a “reward” that is, e.g., the CIDEr score of the generated sentence.

\subsubsection{Prediction Loss}
On the other hand, we can measure the generation with classification task using label ground-truth $\y$. We extract the embeddings of image input and generated sentence from the representation output layer. Considering that the image and corresponding sentence share the same semantic representations, the embeddings of image input and generated sentence can be further put into the shared classifier $f$ for predicting. Thereby, the forward prediction process can be represented as:
\begin{equation}
\begin{split}
\p^v = f(E_e(\vvv)), \quad \p^w = f(D_e(E(\vvv))), \nonumber
\end{split}
\end{equation} 
where $\p^v$ and $\p^w$ are normalized prediction distribution of image input and generated sentence. $f(\cdot)$ denotes the shared classification model for text and image modalities. Without any loss of generality, we utilize three fully connected layer network here. $E_e(\vvv), D_e(E(\vvv)) \in \R^d$ represents the embeddings of image input and generated sentence. Note that $E_e(\vvv)$ and $D_e(E(\vvv))$ are the final embeddings of image/text region embedding with $mean(\cdot)$ operator. The commonly used image captioning dataset (i.e., COCO dataset) is a multi-label dataset, i.e., different from multi-class dataset that each instance only has one ground-truth, each instance has multiple labels. Therefore, we utilize the binary cross entropy loss (BCELoss) here:
\begin{equation}\label{eq:e}
\begin{split}
\ell_{p} = & \sum_{m \in \{v,w\}} H(\p^m, \y^m)\\
H(\p^m, \y^m) = & - \sum_j (y_j^m \log p_j^m + (1-y_j^m)\log (1-p_j^m),
\end{split}
\end{equation}
where $H(\cdot)$ denotes the BCELoss for multi-label prediction, and the model’s predictions are encouraged to be low-entropy (i.e., high-confidence) on supervised data. 

\subsection{Unsupervised Loss}
\subsubsection{Prediction Consistency}
First, we introduce the augmentation technique for transforming the images.  Existing methods usually leverage two kinds of augmentations: a) Weak augmentation is a standard flip-and-shift strategy, which does not significantly change the content of the input. b) Strong augmentation always refers to the AutoAugment~\cite{Cubuk20} and its variant, which uses reinforcement learning to find an augmentation
strategy comprising transformations from the Python Imaging Library\footnote{https://www.pythonware.com/products/pil/}. Considering that ``strong'' augmented (i.e., heavily-augmented) instances are almost certainly outside the data distribution, which leads to the low quality of generated sentence, we leverage the ``weak'' augmentation instead. In result, each image can be expanded to $K+1$ variants, i.e., $\Psi(\vvv) = \{\vvv_{0},\vvv_{1},\cdots,\vvv_{K}\}$, $0$ denotes the raw input.

Then, we input the augmented image set to the image-sentence generator $G$, and extract the embeddings of generated sentences from the representation output layer. The embeddings are further put into the shared classifier for prediction. Thereby, the prediction process can be represented as:
\begin{equation}\label{eq:e1}
\begin{split}
\p_{k}^w = f(D_e(E(\vvv_k))), \quad k \in \{0,1,\cdots,K\}, \\
\end{split}
\end{equation} 
where $f(\cdot)$ denotes the shared classification model for text and image modalities. $D_e(E(\vvv_k)) \in \R^d$ represents the embedding of generated sentence. Similarly, we can acquire the prediction of image inputs: $\p_{k}^v = f(E_e(\vvv_k)), \quad k \in \{0,1,\cdots,K\}$, $E_e(\vvv_k) \in \R^d$ represents the embedding of image. The commonly used image captioning dataset (i.e., COCO dataset) is a multi-label dataset, i.e., different from multi-class dataset that each instance only has one ground-truth, each instance in COCO has multiple labels. Therefore, traditional pseudo-labeling that leverages ``hard'' labels (i.e., the $\arg\max$ of model’s output) is inappropriate, because it is difficult to determine the number of ``hard'' label for each instance. As a consequence, we directly utilize the prediction of image for knowledge distillation~\cite{LinWCS21} in the multi-label BCEloss:
\begin{equation}\label{eq:e2}
\begin{split}
\ell_{pc} = &\sum_{k \in \{0,1,\cdots,K\}} H(\p_{k}^v,\p_{k}^w)\\
H(\p_{k}^v,\p_{k}^w) = & - \sum_j (p_{k_j}^v \log p_{k_j}^w + (1-p_{k_j}^v)\log (1-p_{k_j}^w)),
\end{split}
\end{equation}
where $H(\cdot)$ denotes the binary cross entropy loss (BCELoss), and the model’s predictions are encouraged to be low-entropy (i.e., high-confidence) on unsupervised data. 


\begin{figure}[htb]
	\centering
	\includegraphics[width = 80mm]{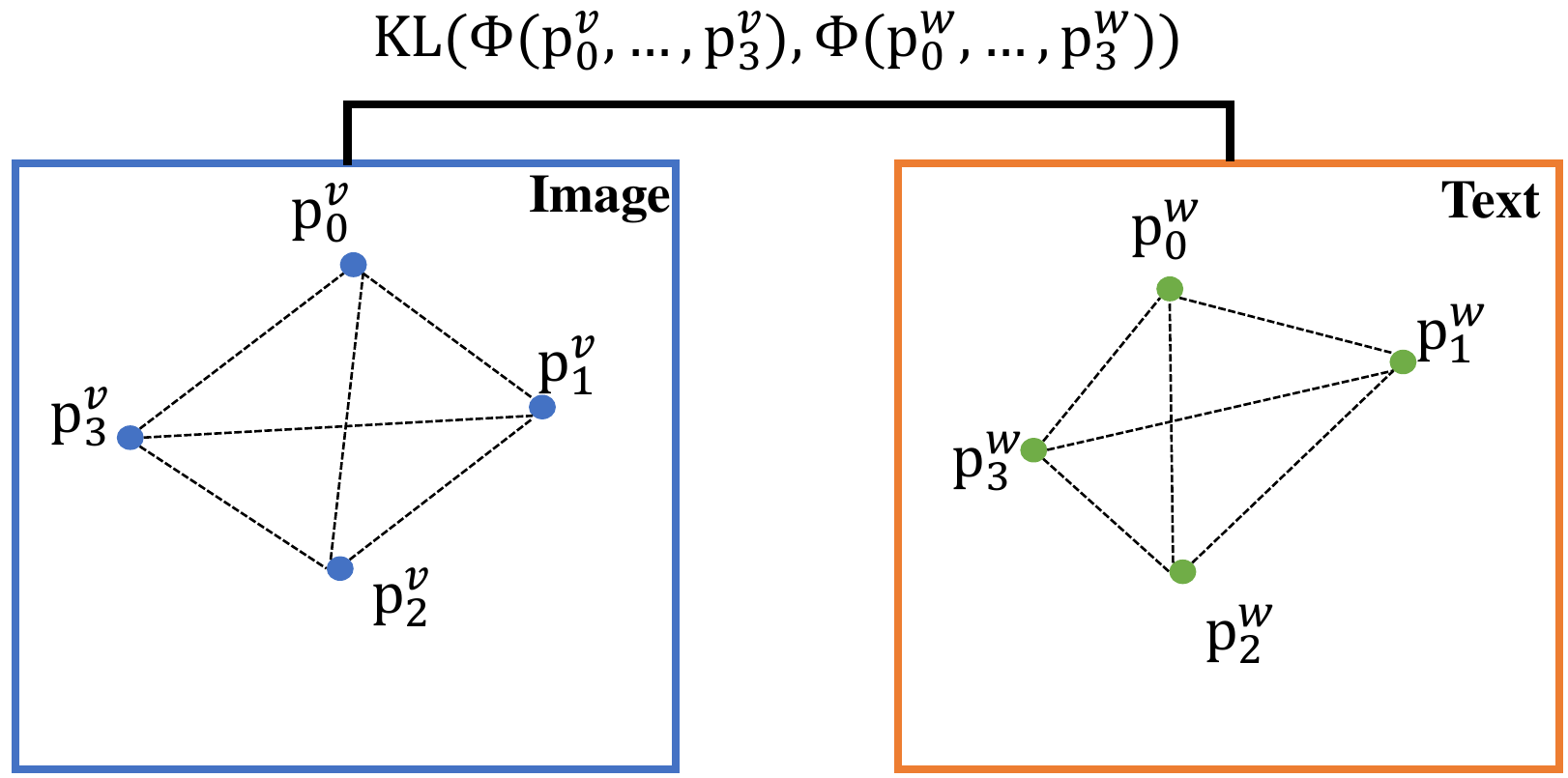}\\
	\caption{The relational consistency. The blue and orange rectangles represent image domain and text domain, respectively. Any point inside the rectangles represents a specific instance in that domain. Relational Consistency: for example, given a tuple of image instances $\{\vvv_{0}, {\vvv}_{1}, {\vvv}_{2}, {\vvv}_{3}, {\vvv}_{4}\}$, relational consistency loss requires that the generated sentences, $\{\w_{0}, {\w}_{1}, {\w}_{2}, {\w}_{3}, {\w}_{4}\}$, should share the similar relation structure with the raw inputs.}\label{fig:consistency}
\end{figure}

\subsubsection{Relation Consistency}
Inspired by the linguistic structuralism~\cite{Matthews2001A} that relations can better present the knowledge than individual example, the primary information actually lies in the structure of the data space. Therefore, we define a new relation consistency loss, $\ell_{rc}$, using a metric learning-based constraint, which calculates the KL divergence of the similarity vectors between the image inputs and generated sentences. The relation consistency aims to ensure the structural knowledge using mutual relations of data examples in the raw inputs. Specifically, each image input can be denoted as a bag of $K+1$ instances, i.e., $\Psi(\vvv)$, while the corresponding generated sentences can also be represented as a bag of instances, i.e., $G(\Psi(\vvv))$. With the shared classifier, the image and sentence prediction can be formulated as:
\begin{equation}\label{eq:e3}
\begin{split}
\p_{k}^v = &f(E_e(\vvv_k)), \quad k \in \{0,1,\cdots,K\} \\
\p_{k}^w = &f(D_e(E(\vvv_k))), \quad k \in \{0,1,\cdots,K\}, \nonumber
\end{split}
\end{equation}
With the predictions of image inputs and generated sentences, the objective of relational consistency can be formulated as:
\begin{equation}\label{eq:e4}
\begin{split}
\ell_{rc} = &  KL(\Phi(\p_{0}^v,\p_{1}^v, \cdots,\p_{K}^v), \Phi(\p_{0}^w,\p_{1}^w,\cdots,\p_{K}^w)),  
\end{split}
\end{equation}
$KL(a,b) = a \log \frac{a}{b}$ is the KL divergence that penalizes difference between the similarity distributions of image inputs and the similarity distributions of generated sentences. $\Phi$ is a relation prediction function, which measures a relation energy of the given tuple. In detail, $\Phi$ aims to measure the similarities formed by the examples in semantic prediction space:
\begin{equation}\label{eq:e5}
\begin{split}
\Phi(\p_{0}^v,\p_{1}^v,\cdots,\p_{K}^v) &= [q_{{mn}}^v]_{m,n \in [0,\cdots,K]} \\
\Phi(\p_{0}^w,\p_{1}^w,\cdots,\p_{K}^w) &= [q_{{mn}}^w]_{m,n \in [0,\cdots,K]} \\
q_{{mn}}^v &= \frac{exp(d_{{mn}}^v)}{\sum exp(d_{\cdot}^v)} \\
q_{{mn}}^w &= \frac{exp(d_{{mn}}^w)}{\sum exp(d_{\cdot}^w)}, \\
\end{split}
\end{equation}
where $d_{{mn}}^v = -Dist(\p_{m}^v,\p_{n}^v), d_{{mn}}^w = -Dist(\p_{m}^w,\p_{n}^w)$ measures the distance between $(\p_{m}^v, \p_{n}^v)$ and between $(\p_{m}^w, \p_{n}^w)$ respectively, $Dist(\p_{m}^v,\p_{n}^v) = \| \p_{m}^v - \p_{n}^v\|_2$ and $ Dist(\p_{m}^w,\p_{n}^w) = \| \p_{m}^w - \p_{n}^w\|_2$. $q_{{mn}}^v$ and $q_{{mn}}^w$ denote the relative instance-wise similarity. Finally, we pull the $[q_{{mn}}^v]$ and $[q_{{mn}}^v]$ into vector form. In result, the relation consistency loss can deliver the relationship of examples by penalizing structure differences. Since the structure has higher-order properties than single output, it can transfer knowledge more effectively, and is more suitable for consistency measure.

\subsection{Overall Function}
In summary, with the limited amount of parallel image-sentence pairs and large amount of undescribed images, we define the total loss by combining the Eq. \ref{eq:e0}, Eq. \ref{eq:e}, Eq. \ref{eq:e2} and Eq. \ref{eq:e4}:
\begin{equation}\label{eq:e6}
\begin{split}
L = &\sum_{j=1}^{N_l} \ell_s(\vvv_i,\w_i,\y_i) + \sum_{j=1}^{N_u} \lambda_1 \ell_{pc}(\vvv_i) + \lambda_2 \ell_{rc}(\vvv_i)\\
\ell_s(\vvv_i,\w_i,\y_i) = & \ell_c (\vvv_i,\w_i) + \ell_p(\vvv_i,\w_i,\y_i) 
\end{split}
\end{equation}
where $\ell_c$ denotes the captioning loss, which can be adopted as $\ell_{XE}$ or $\ell_{RL}$ in Eq. \ref{eq:e0}. Note that $\ell_c$ and $\ell_p$ are with same order of magnitude, so we do not add hyper-parameter here. $\lambda_1$ and $\lambda_2$ are scale values that control the weights of different losses. In $\ell_{s}$, we use labeled images and sentences to jointly train the shared classifier $f$, which increases the amount of training data, as well as adjusts the classifier to better suit subsequent prediction of augmented images and generated sentences. Furthermore, considering that the pseudo labels $\p^v,\p^w$ may exist noises, we can also adopt a confidence threshold that retains confident generated sentences. The Eq. \ref{eq:e6} can be reformulated as:
\begin{equation}\label{eq:e7}
\begin{split}
L = &\sum_{j=1}^{N_l} \ell_s(\vvv_i,\w_i,\y_i) + \sum_{j=1}^{N_u} {\bf 1}(\max(\p_{j_0}^v) \ge \tau ) \\ &  \big\{ \lambda_1 \ell_{pc}(\vvv_i) + \lambda_2 \ell_{rc}(\vvv_i)\big \} \\
\ell_s(\vvv_i,\w_i,\y_i) = & \ell_{XE}(\vvv_i,\w_i) + \ell_p(\vvv_i,\w_i,\y_i) 
\end{split}
\end{equation}
where $\p_{j_0}^v$ denotes the prediction probability of the $j-$th raw image input, $\tau$ is a scalar hyperparameter denoting the threshold above which we retain the generated sentences.  The details are shown in Algorithm \ref{alg:alg1}. 

{\begin{algorithm}[htb]
		\caption{The Code of CPRC}
		\label{alg:alg1}
		\textbf{Input}:\\
		Data: $\D = \{\{\vvv_i,\w_i,\y_i\}_{i=1}^{N_p}, \{\vvv_j\}_{j=1}^{N_u}\}$ \\
		Parameters: $\lambda_1$, $\lambda_2$  \\
		\textbf{Output}:\\
		Image captioning mapping function: $G$ \\
		\begin{algorithmic}[1]{
				\STATE Initialize the $G$ and $f$ randomly;
				\WHILE {stop condition is not triggered}
				\FOR{mini-batch sampled from $\D$}
				\STATE Calculate $\ell_{s}$ according to Eq. \ref{eq:e0} and Eq. \ref{eq:e};
				\STATE Calculate $\ell_{pc}$ according to Eq. \ref{eq:e2};
				\STATE Calculate $\ell_{rc}$ according to Eq. \ref{eq:e4};
				\STATE Calculate $L$ according to Eq. \ref{eq:e6} or Eq. \ref{eq:e7};
				\STATE Update model parameters of $G,f$ using SGD;
				\ENDFOR
				\ENDWHILE
			}
		\end{algorithmic}
\end{algorithm}}

\begin{table*}[!htb]{
		\centering
		\caption{Performance of comparison methods on MS-COCO “Karpathy” test split, where B$@$N, M, R, C and S are short for BLEU@N, METEOR, ROUGE-L, CIDEr-D and SPICE scores. }
		\label{tab:tab1}
		\begin{tabular*}{1\textwidth}{@{\extracolsep{\fill}}@{}l|@{}c|@{}c|@{}c|@{}c|@{}c|@{}c|@{}c|@{}c|@{}c|@{}c|@{}c|@{}c|@{}c|@{}c|@{}c|@{}c}
			\toprule
			\multirow{2}{*}{Methods} & \multicolumn{8}{c|}{Cross-Entropy Loss} & \multicolumn{8}{c}{CIDEr-D Score Optimization} \\
			\cmidrule(l){2-17}
			& B$@$1 & B$@$2 & B$@$3 & B$@$4 & M & R & C & S & B$@$1 & B$@$2 & B$@$3 & B$@$4 & M & R & C & S\\
			\midrule
			SCST  &56.8&38.6&25.4&16.3&16.0 &42.4&38.9&9.3&59.4 &39.5&25.3 &16.3&17.0&42.9&43.7&9.9\\
			AoANet &67.9&49.8&34.7 &23.2&20.9&49.2&69.2&14.3&66.8 &48.6&34.1&23.6&21.8 &48.7&70.4&15.2\\
			AAT &63.2&45.8&31.7 &21.3&19.0&47.6&58.0&12.4&66.7 &48.1&33.3&22.7&20.4 &47.8&63.5&13.2\\
			ORT &63.6&45.8&31.7 &21.4&19.4&46.9&61.1&12.6&65.3 &46.5&31.9&21.3&20.3 &47.2&62.0&13.3\\
			GIC &63.0&46.8&33.2 &20.0&19.2&50.3&50.5&12.3&64.7 &46.9&32.0&20.7&19.0 &47.8&55.7&12.5\\
			\midrule
			Graph-align &-&-&-&-&-&-&-&-&67.1 &47.8&32.3&21.5&20.9&47.2&69.5&15.0\\
			UIC &-&-&-&-&-&-&-&-&41.0 &22.5&11.2&5.6&12.4&28.7&28.6&8.1\\
			\midrule
			A3VSE &68.0&50.0&34.9 &23.3&20.8&49.3&69.6&14.5&67.6 &49.6&35.2&24.5&22.1 &49.3&72.4&15.3\\
			\midrule
			AoANet+P &67.4&49.7&35.2 &24.3&22.3&49.1&71.7&14.9&67.2 &49.5&35.9&24.4&21.6 &50.1&74.2&15.7\\
			AoANet+C &67.1&49.4&35.2 &24.5&22.7&49.5&71.5&14.9&67.8 &49.4&35.5&24.7&22.0 &50.0&73.9&15.6\\
			PL &67.8&49.6&35.2&24.2&22.0&50.4&74.7&15.6&67.9 &50.0&35.6&24.3&22.2&49.7&76.6&16.1\\
			AC &67.8&48.8&34.6&23.7&21.9&49.1&69.7&14.5&67.9 &50.0&25.3&24.1&22.1&49.7&73.0&15.5\\
			Embedding+ &65.1&46.4&31.9 &21.5&20.7&47.6&65.1&14.1&65.6 &47.1&32.3&22.6&20.8 &47.8&69.1&14.5\\
			Semantic+ &68.3&49.9&34.9 &23.8&21.5&49.9&70.3&14.7&69.3 &50.8&35.5&24.1&21.6 &50.0&72.7&14.9\\
			Strong+ &68.4&50.8&35.4&24.8&22.5&\bf 50.6&77.8& 16.2&69.5 &51.5& 36.7& 25.5&23.3&50.6&78.6&16.7\\
			w/o Prediction &68.3&49.6&35.3 &24.4&22.2&49.6&70.5&15.0&68.2 &50.4&35.8&24.8&22.5 &50.1&73.6&15.6\\
			w/o Relation &68.1&50.0& 35.5 &24.8&22.4&50.5&75.2&15.8&68.3 &50.5&35.8&24.9&22.7 &50.4&76.9&16.3\\
			w/o $\tau$ &66.9&49.8&34.5&24.2&21.5&49.5&76.2&15.4&68.5 &50.8&36.2&25.0&22.5&49.8&77.5&16.2\\
			\midrule
			
			CPRC &\bf 68.8&\bf 51.1&\bf 35.5 &\bf 24.9&\bf 22.8&50.4&\bf 77.9&\bf 16.2&\bf 69.9 &\bf 51.8&\bf 36.7&\bf 25.5&\bf 23.4 &\bf 50.7&\bf 78.8&\bf 16.8\\
			\bottomrule
	\end{tabular*}}
\end{table*}

\begin{figure*}[t]
	\begin{center}
		\begin{minipage}[h]{43mm}
			\centering
			\includegraphics[width=43mm]{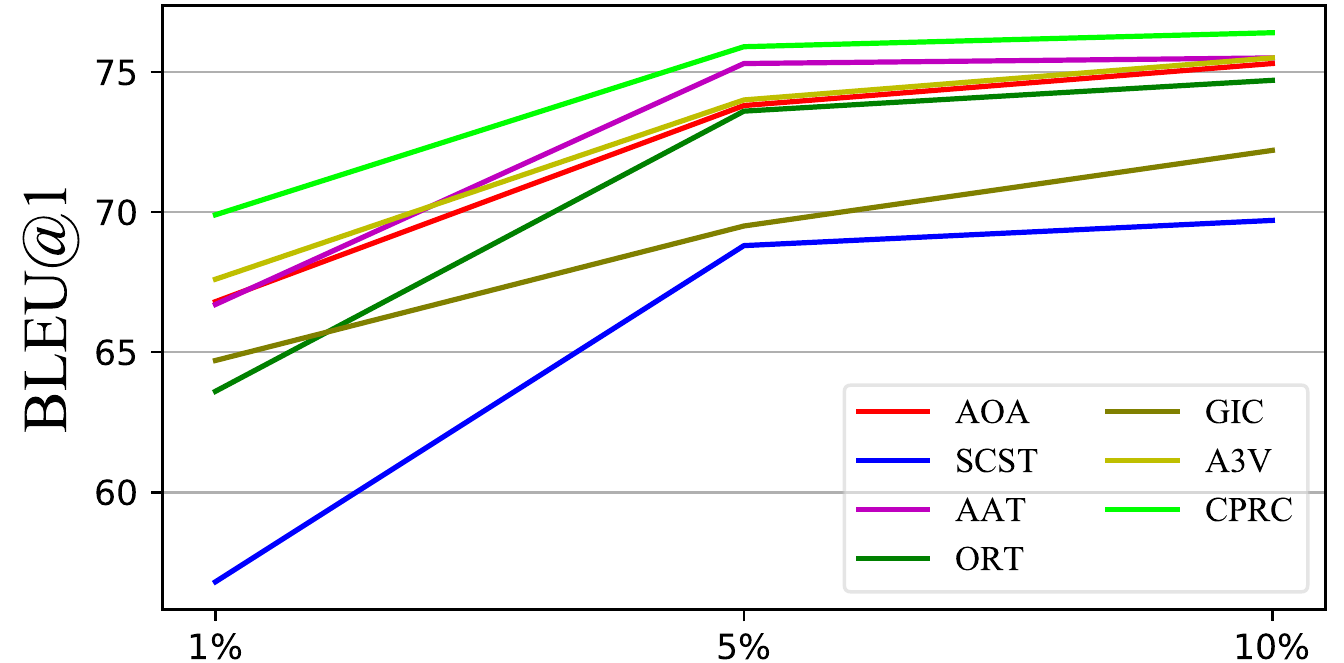}\\
			\mbox{ \;\;\;\; ({\it a}) {BLEU$@$1}}
		\end{minipage}
		\begin{minipage}[h]{43mm}
			\centering
			\includegraphics[width=43mm]{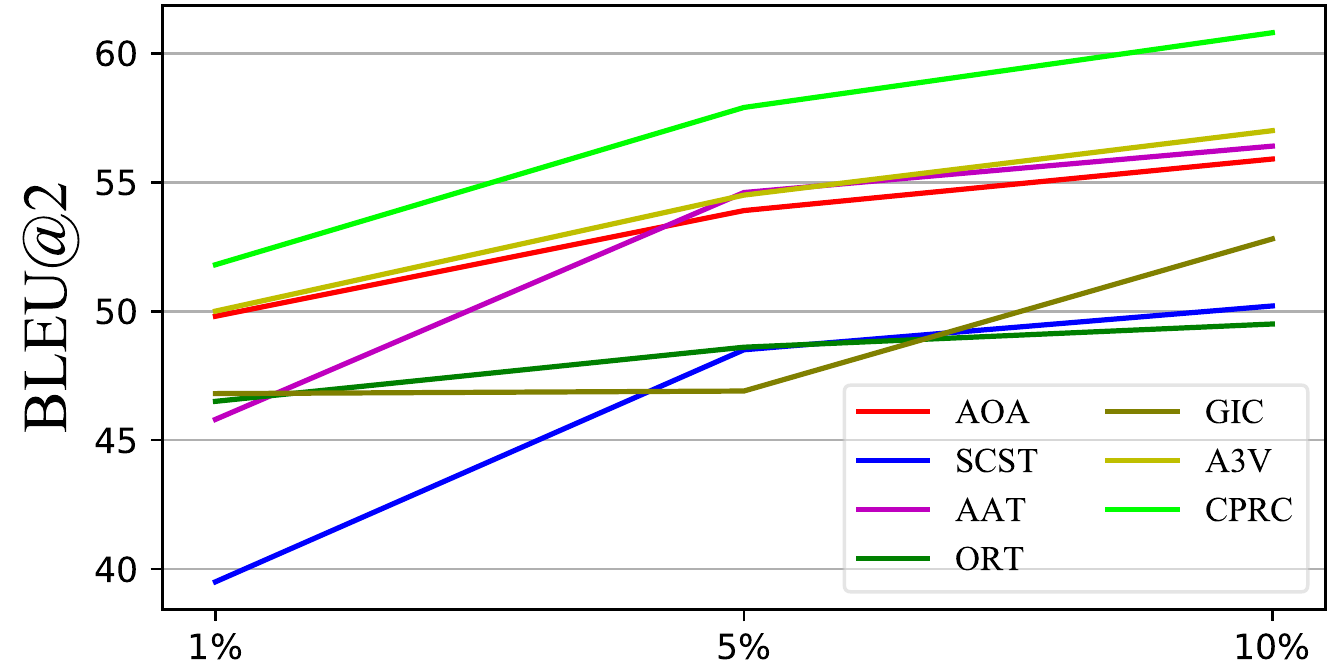}\\
			\mbox{ \;\;\;\; ({\it b}) {BLEU$@$2}}
		\end{minipage}
		\begin{minipage}[h]{43mm}
			\centering
			\includegraphics[width=43mm]{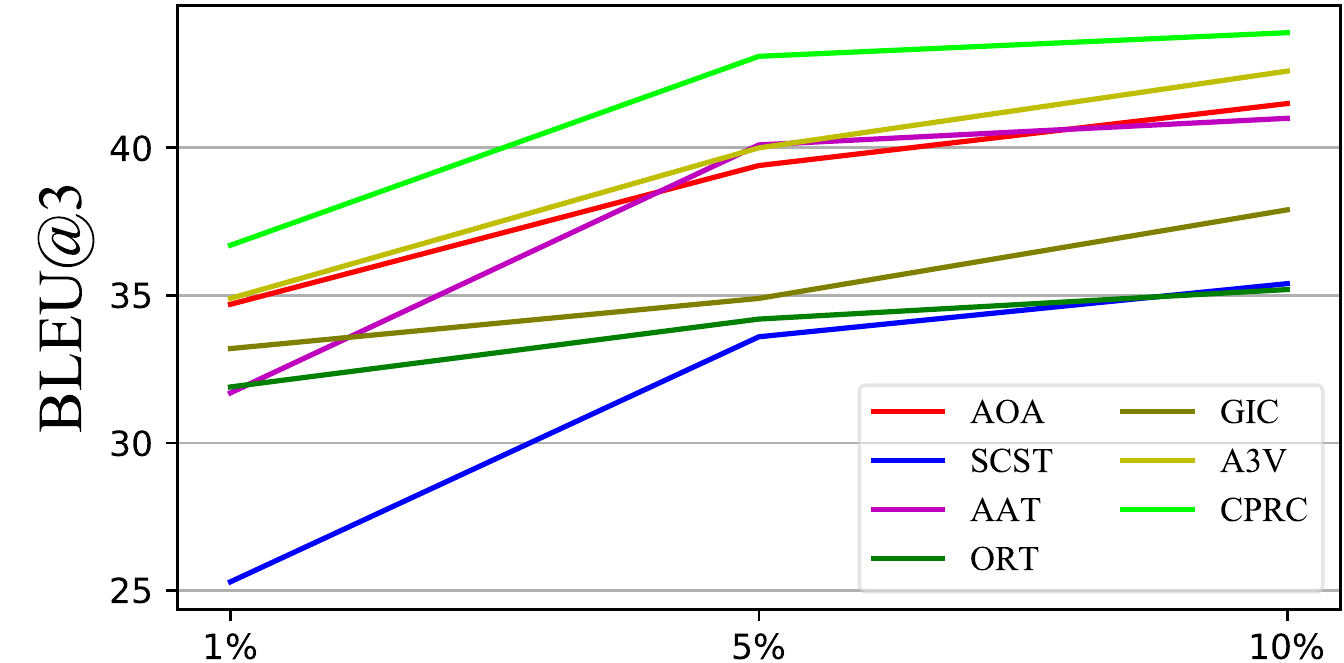}\\
			\mbox{ \;\;\;\; ({\it c}) {BLEU$@$3}}
		\end{minipage}
		\begin{minipage}[h]{43mm}
			\centering
			\includegraphics[width=43mm]{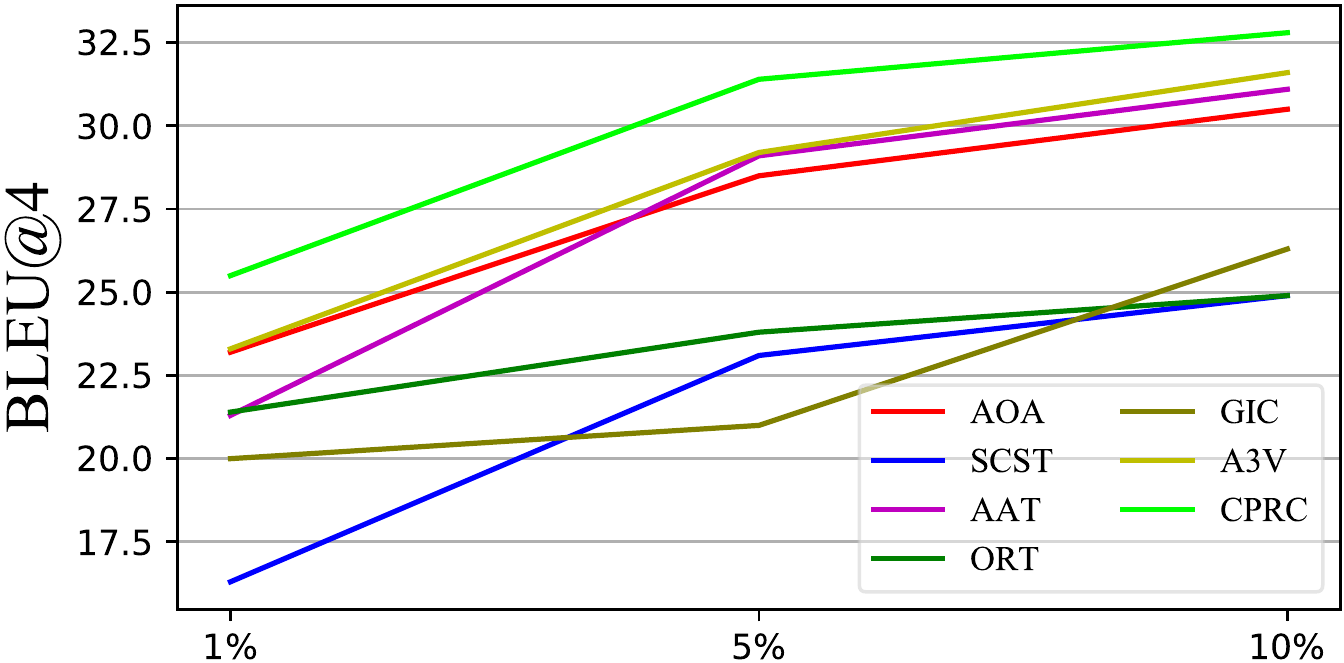}\\
			\mbox{ \;\;\;\; ({\it d}) {BLEU$@$4}}
		\end{minipage}\\
		\begin{minipage}[h]{43mm}
			\centering
			\includegraphics[width=43mm]{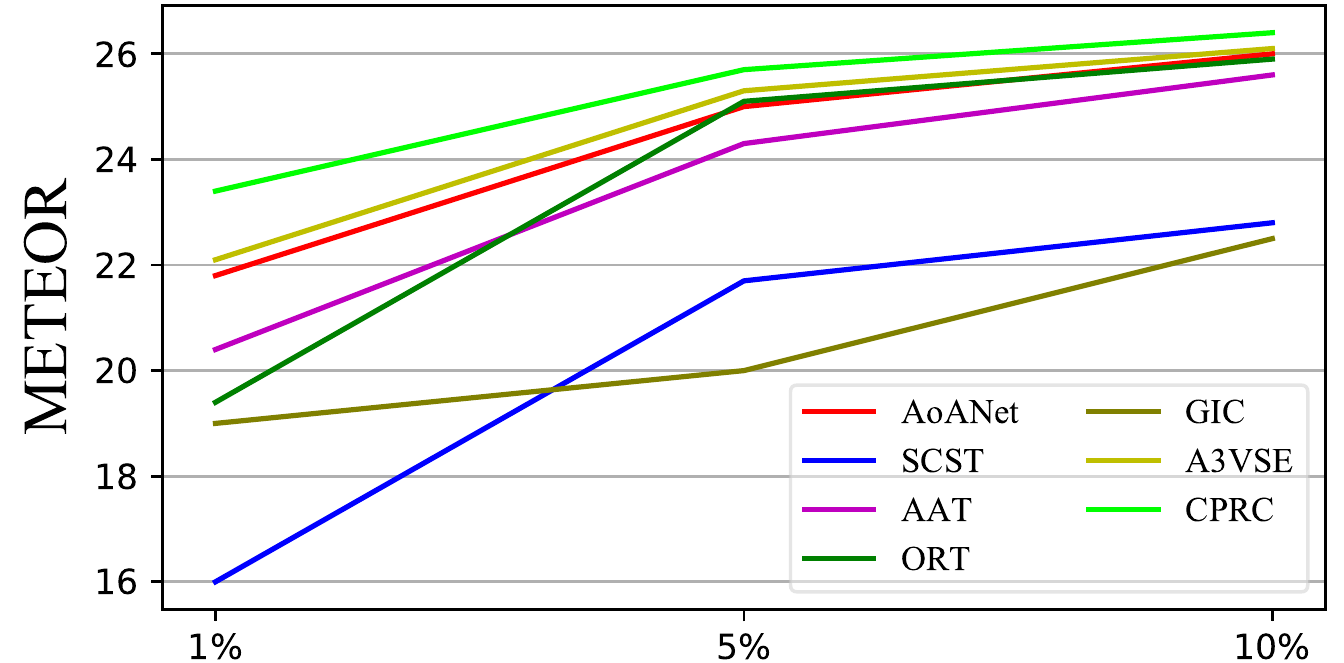}\\
			\mbox{ \;\;\;\; ({\it e}) {METEOR}}
		\end{minipage}
		\begin{minipage}[h]{43mm}
			\centering
			\includegraphics[width=43mm]{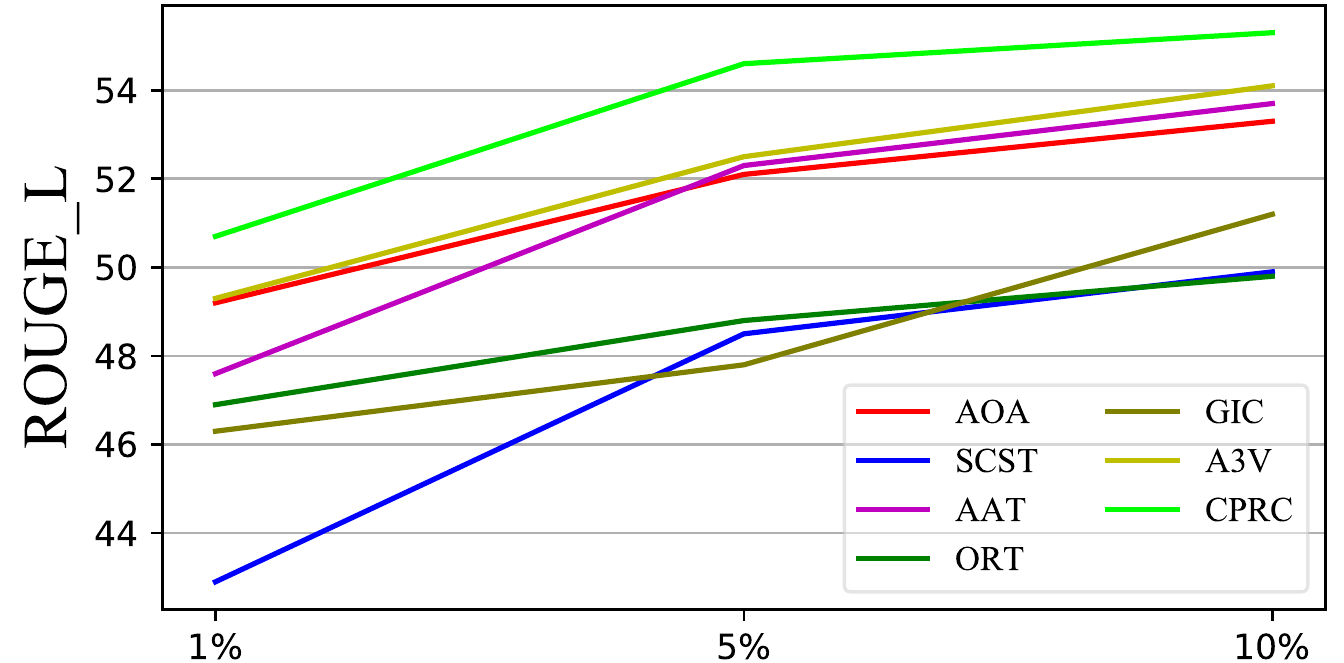}\\
			\mbox{ \;\;\;\; ({\it f}) {ROUGE-L}}
		\end{minipage}
		\begin{minipage}[h]{43mm}
			\centering
			\includegraphics[width=43mm]{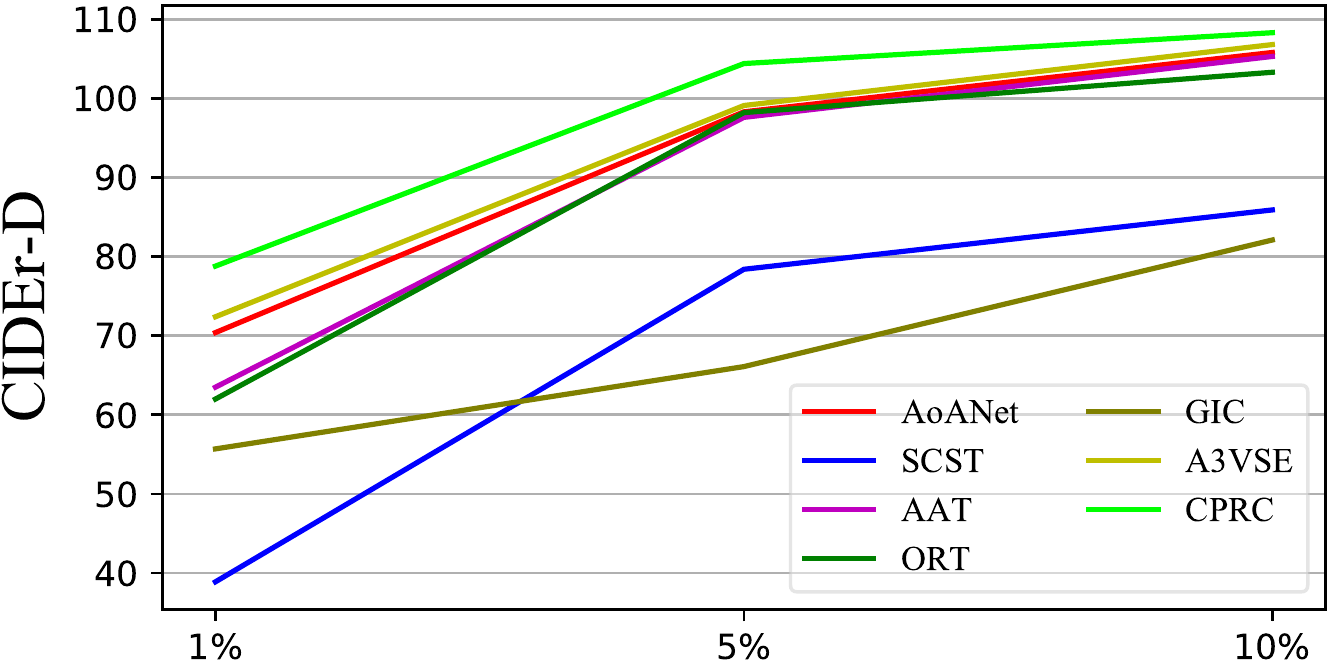}\\
			\mbox{ \;\;\;\; ({\it g}) {CIDEr-D}}
		\end{minipage}
		\begin{minipage}[h]{43mm}
			\centering
			\includegraphics[width=43mm]{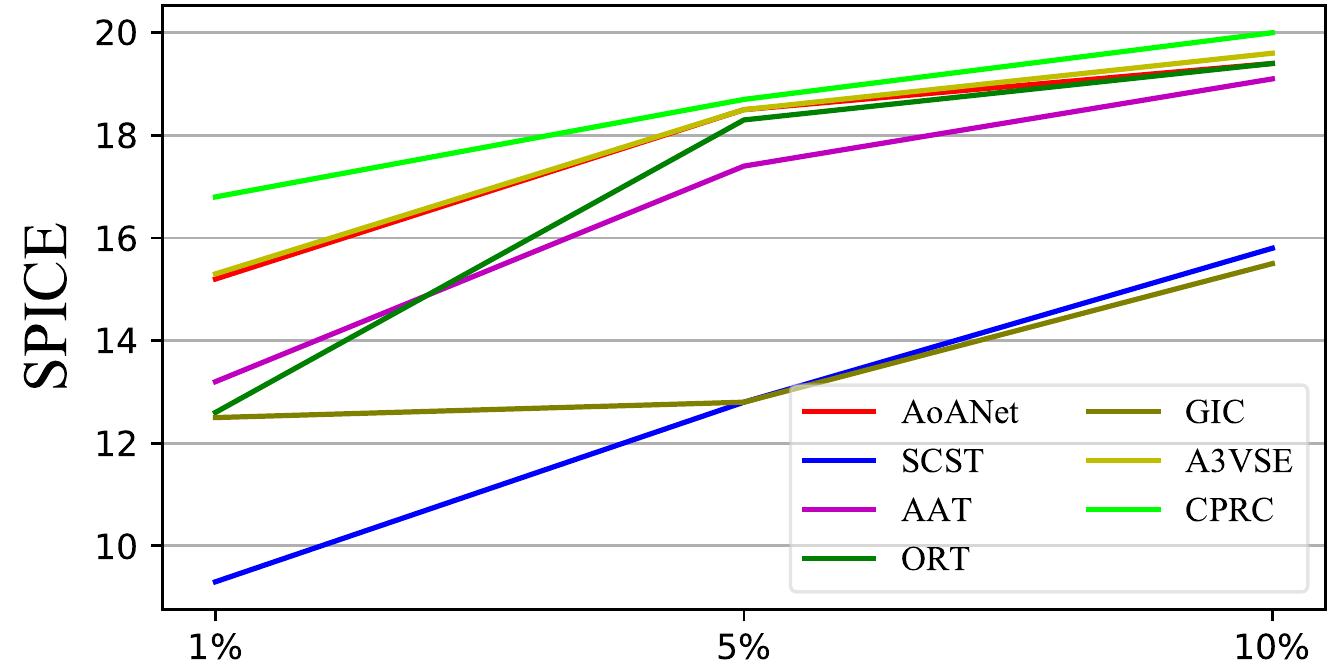}\\
			\mbox{ \;\;\;\; ({\it h}) {SPICE}}
		\end{minipage}
	\end{center}
	\caption{Relationship between captioning performance with different ratio of supervised data.}\label{fig:f1}
\end{figure*}

\begin{figure*}[htb]
	\begin{center}
		\begin{minipage}[h]{42mm}
			\centering
			\includegraphics[width=42mm]{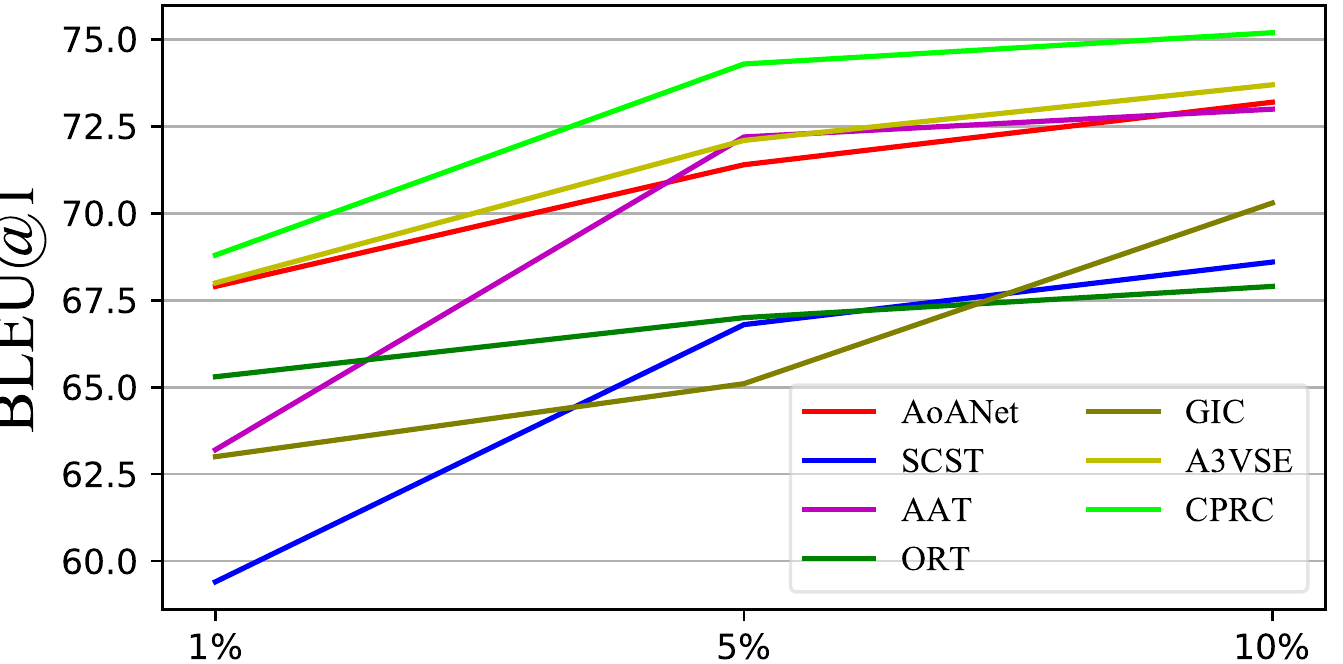}\\
			\mbox{ \;\;\;\; ({\it a}) {BLEU$@$1}}
		\end{minipage}
		\begin{minipage}[h]{42mm}
			\centering
			\includegraphics[width=42mm]{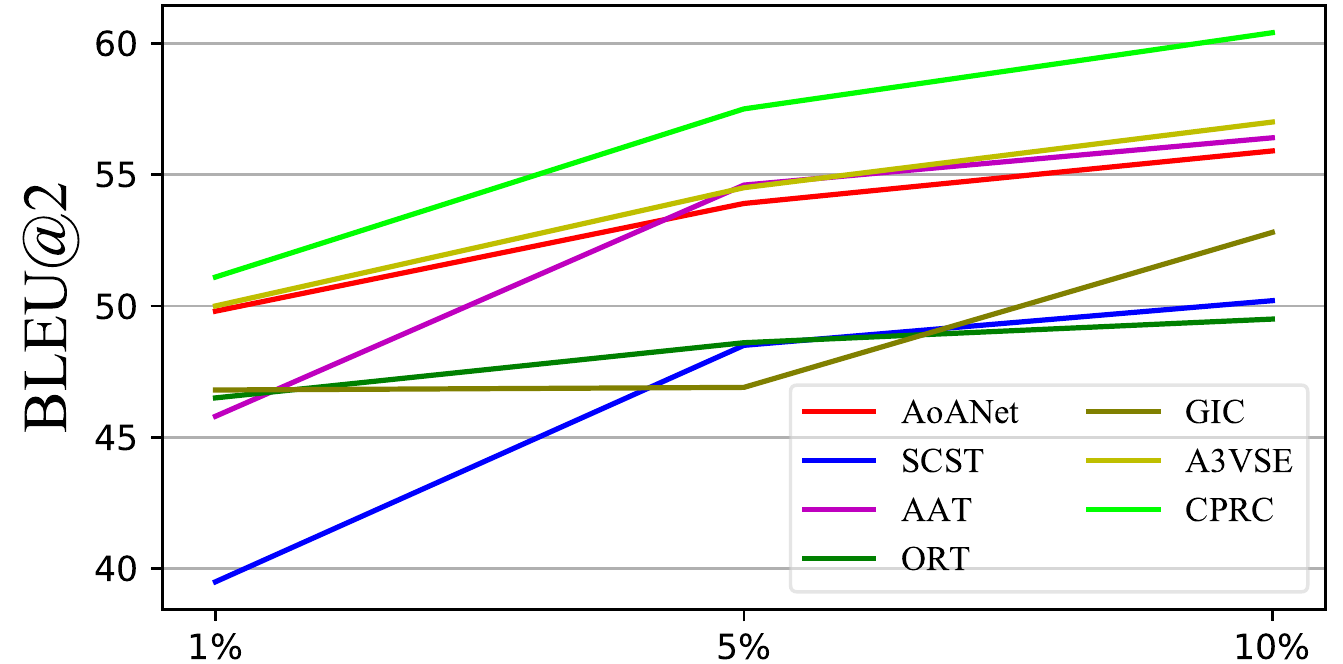}\\
			\mbox{ \;\;\;\; ({\it b}) {BLEU$@$2}}
		\end{minipage} 
		\begin{minipage}[h]{42mm}
			\centering
			\includegraphics[width=42mm]{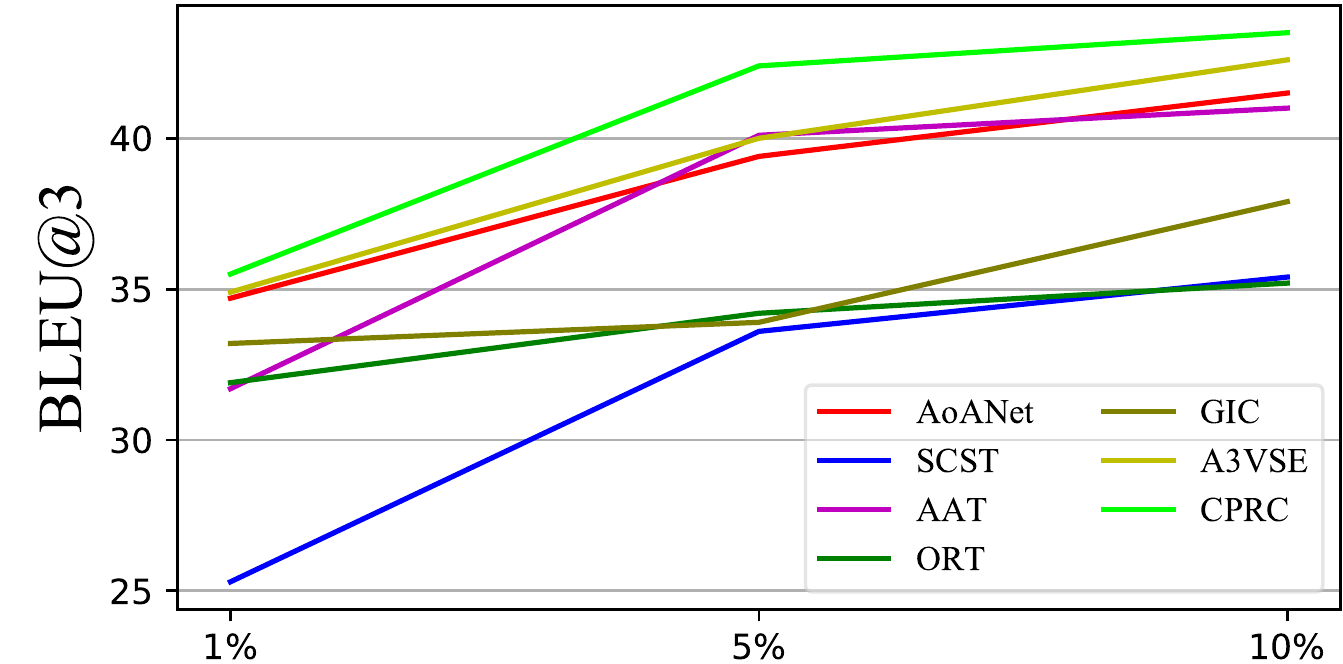}\\
			\mbox{ \;\;\;\; ({\it c}) {BLEU$@$3}}
		\end{minipage}
		\begin{minipage}[h]{42mm}
			\centering
			\includegraphics[width=42mm]{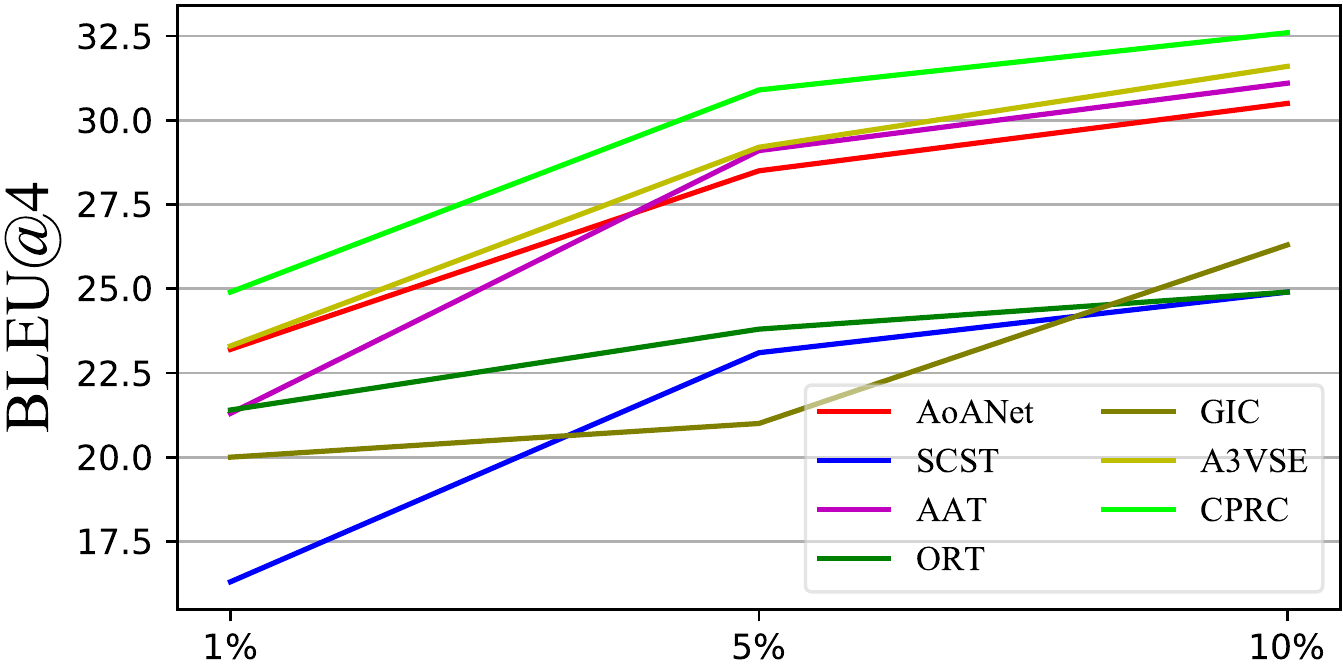}\\
			\mbox{ \;\;\;\; ({\it d}) {BLEU$@$4}}
		\end{minipage}\\
		\begin{minipage}[h]{42mm}
			\centering
			\includegraphics[width=42mm]{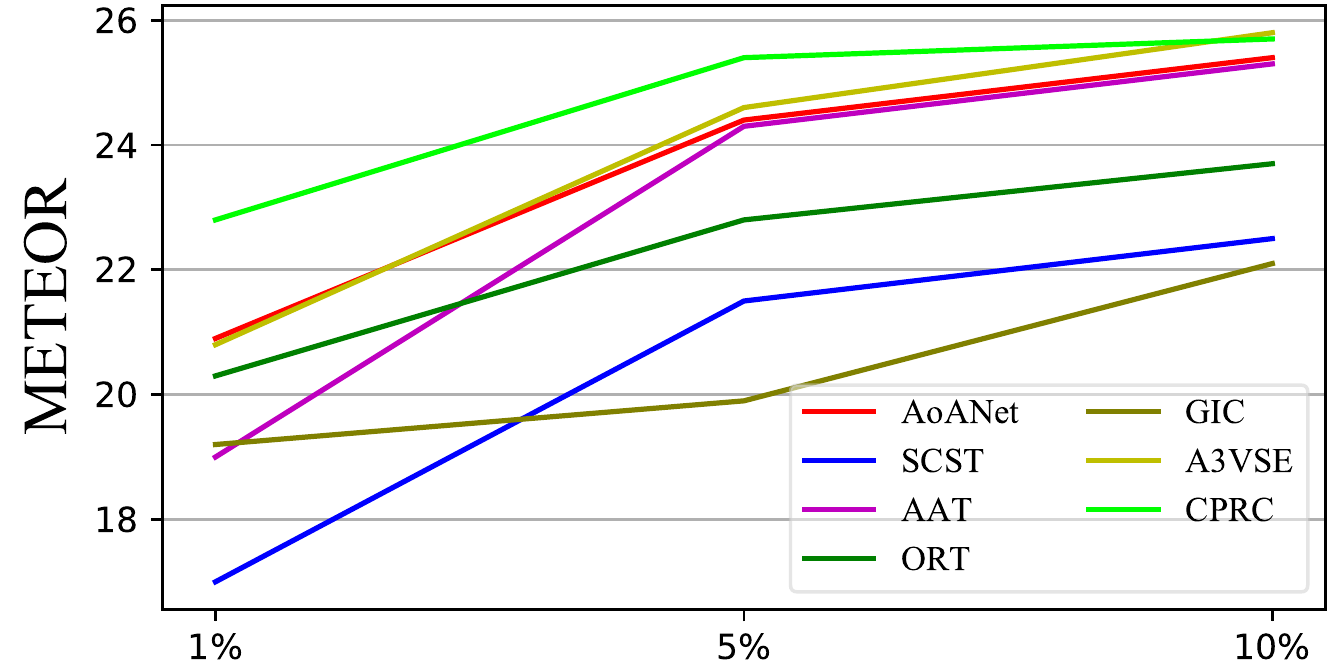}\\
			\mbox{ \;\;\;\; ({\it e}) {METEOR}}
		\end{minipage}
		\begin{minipage}[h]{42mm}
			\centering
			\includegraphics[width=42mm]{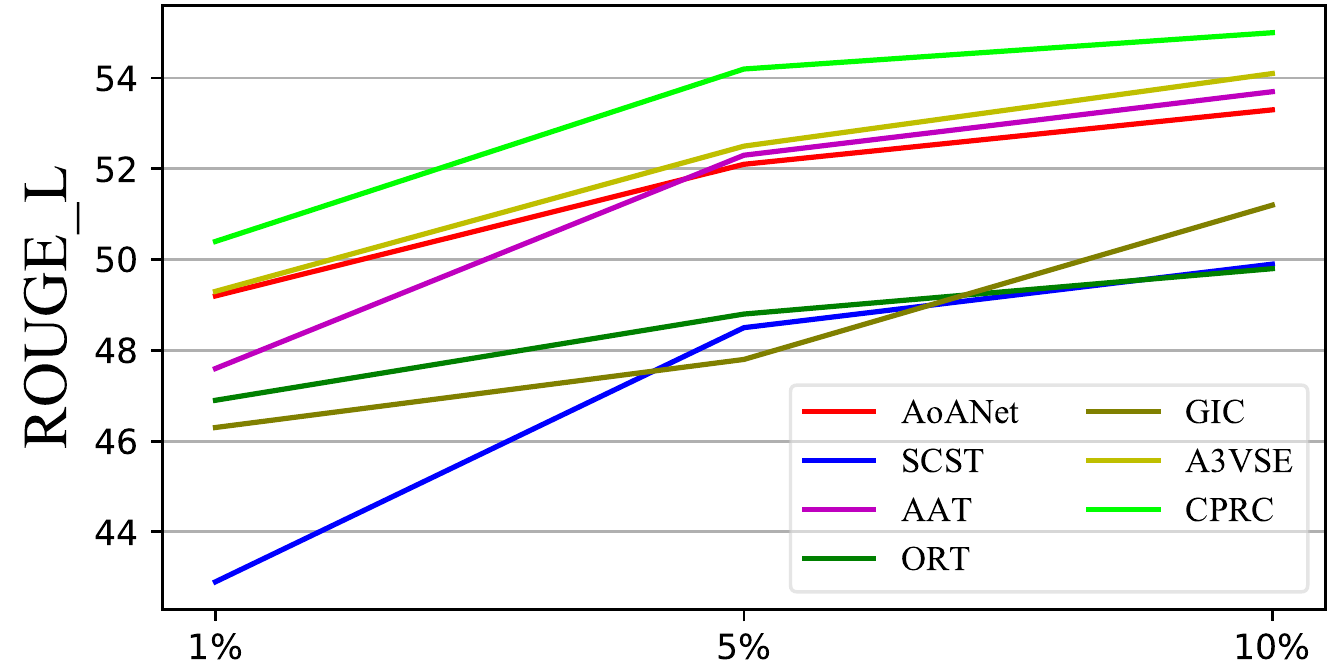}\\
			\mbox{ \;\;\;\; ({\it f}) {ROUGE-L}}
		\end{minipage}
		\begin{minipage}[h]{42mm}
			\centering
			\includegraphics[width=42mm]{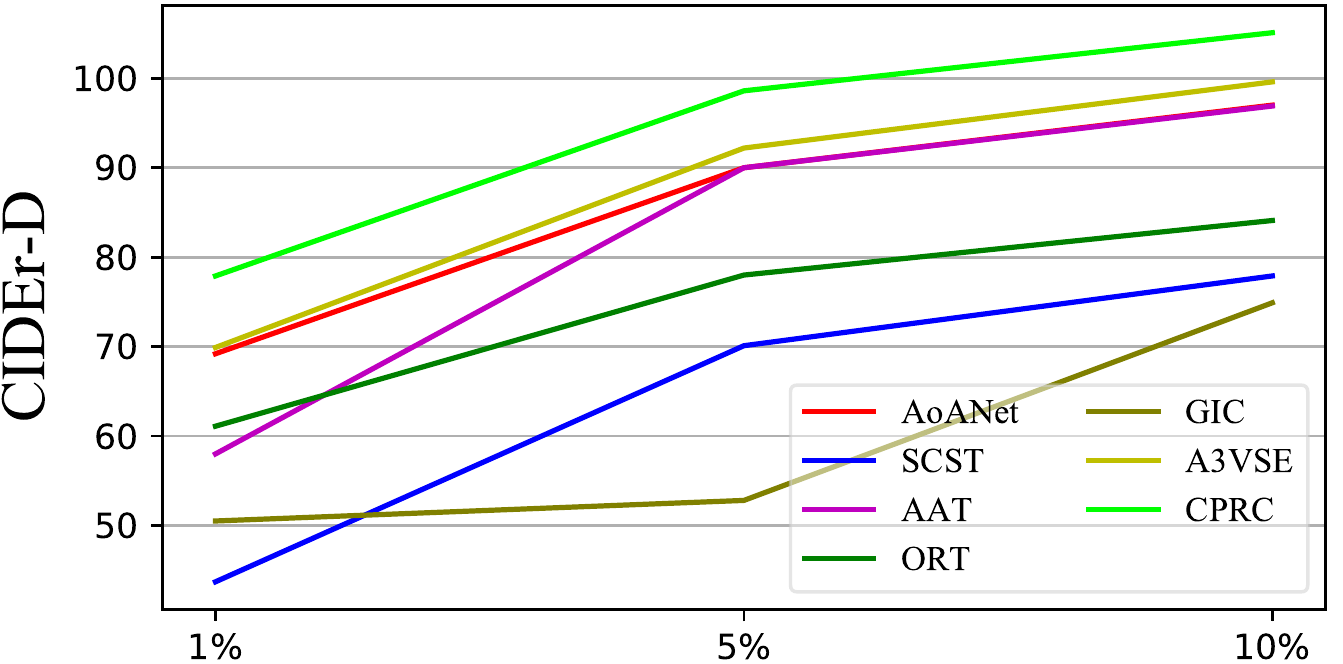}\\
			\mbox{ \;\;\;\; ({\it g}) {CIDEr-D}}
		\end{minipage}
		\begin{minipage}[h]{42mm}
			\centering
			\includegraphics[width=42mm]{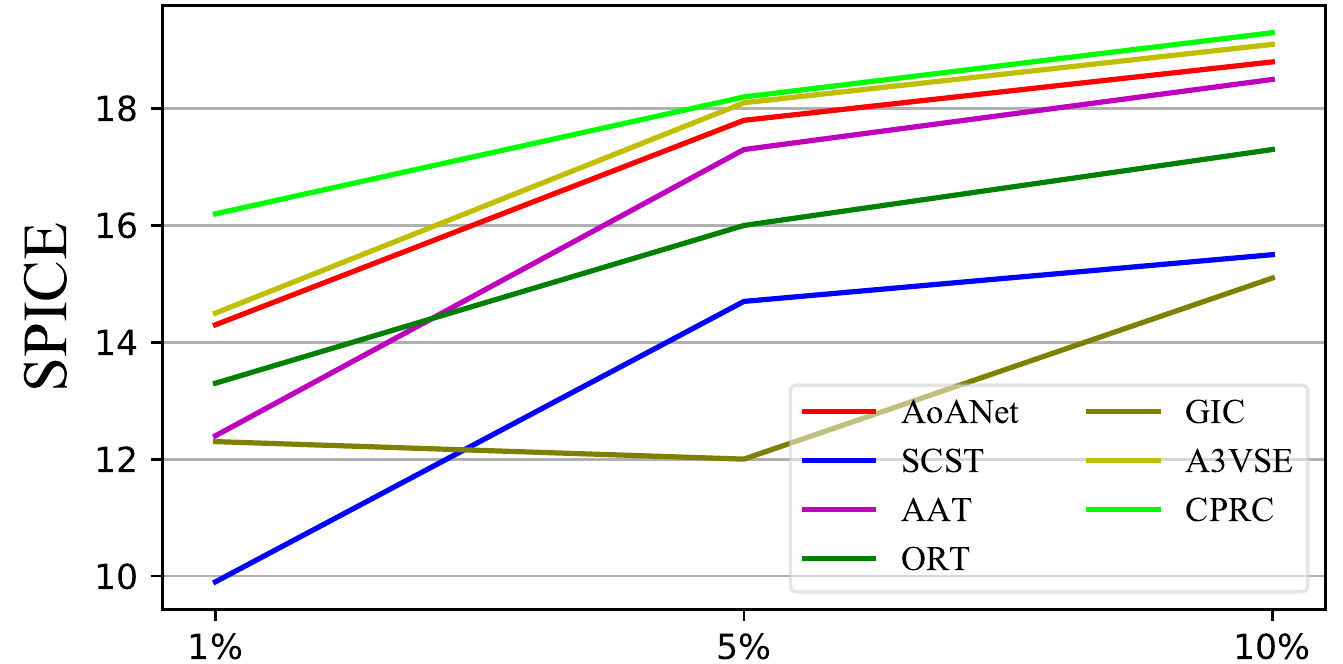}\\
			\mbox{ \;\;\;\; ({\it h}) {SPICE}}
		\end{minipage}
	\end{center}
	\caption{Relationship between caption performance with different ratio of supervised data (Cross-Entropy Loss).}\label{fig:f11}
\end{figure*}

\begin{figure*}[t]
	\centering
	\includegraphics[width=180mm]{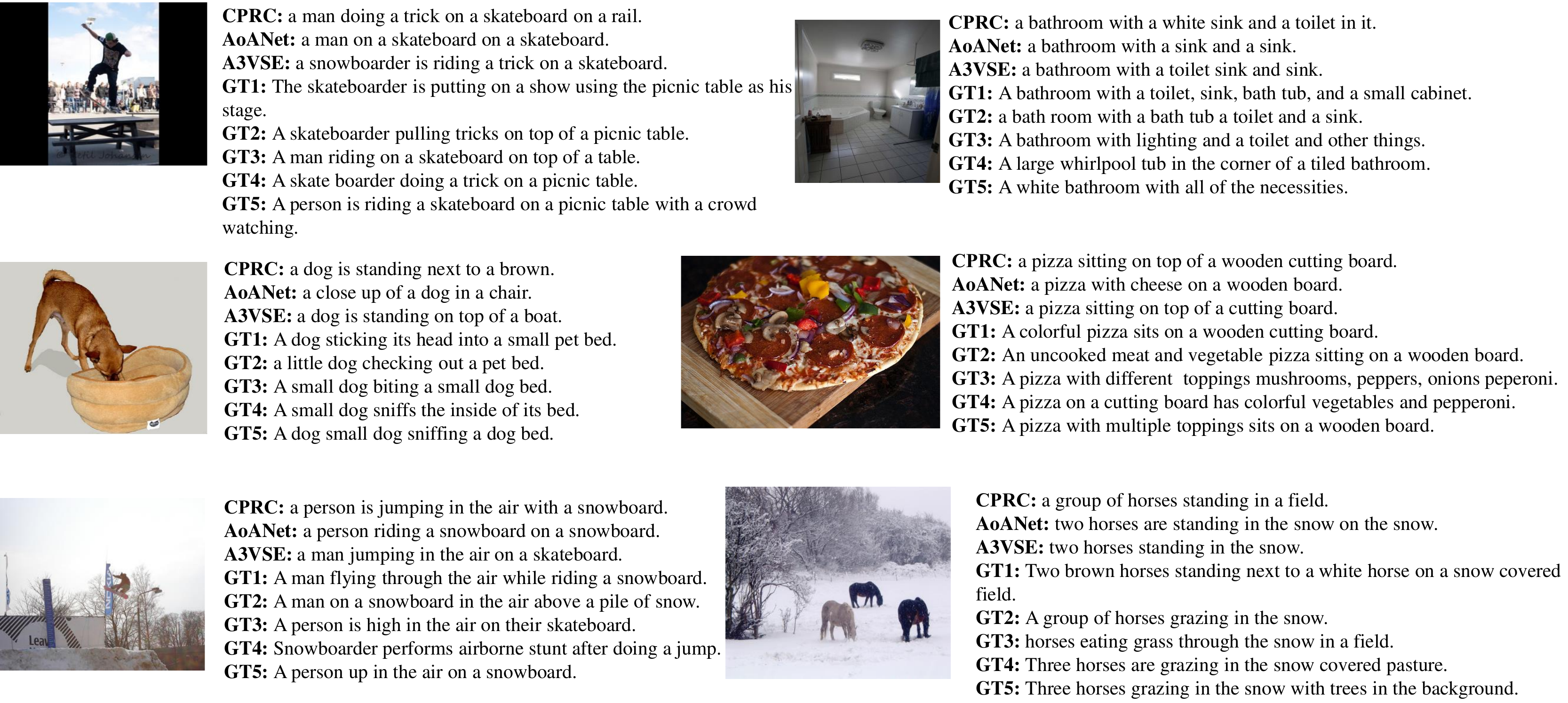}\\
	\caption{Examples of captions generated by CPRC and baseline models as well as the corresponding ground truths.}\label{fig:f3}
\end{figure*}

\begin{figure*}[t]
	\centering
	\includegraphics[width=180mm]{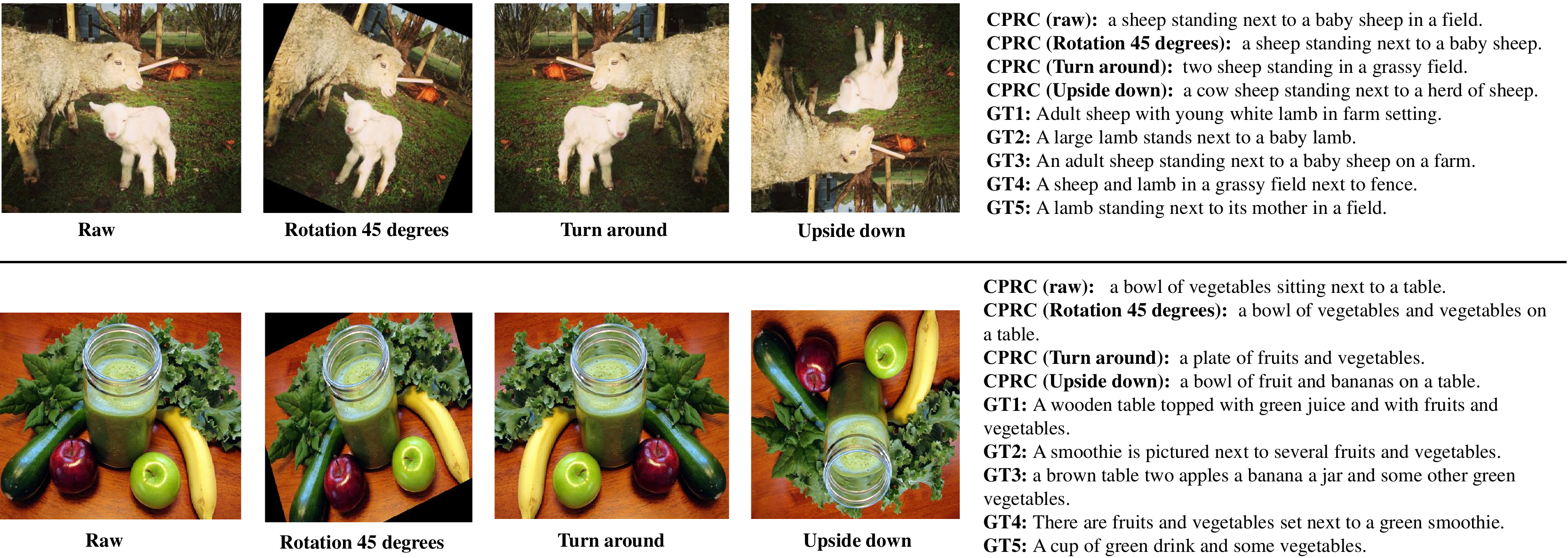}\\
	\caption{(Best viewed in color) Examples of captions generated by augmented images.}\label{fig:f4}
\end{figure*}

\begin{figure}[t]
	\begin{center}
		\begin{minipage}[h]{85mm}
			\centering
			\includegraphics[width=85mm]{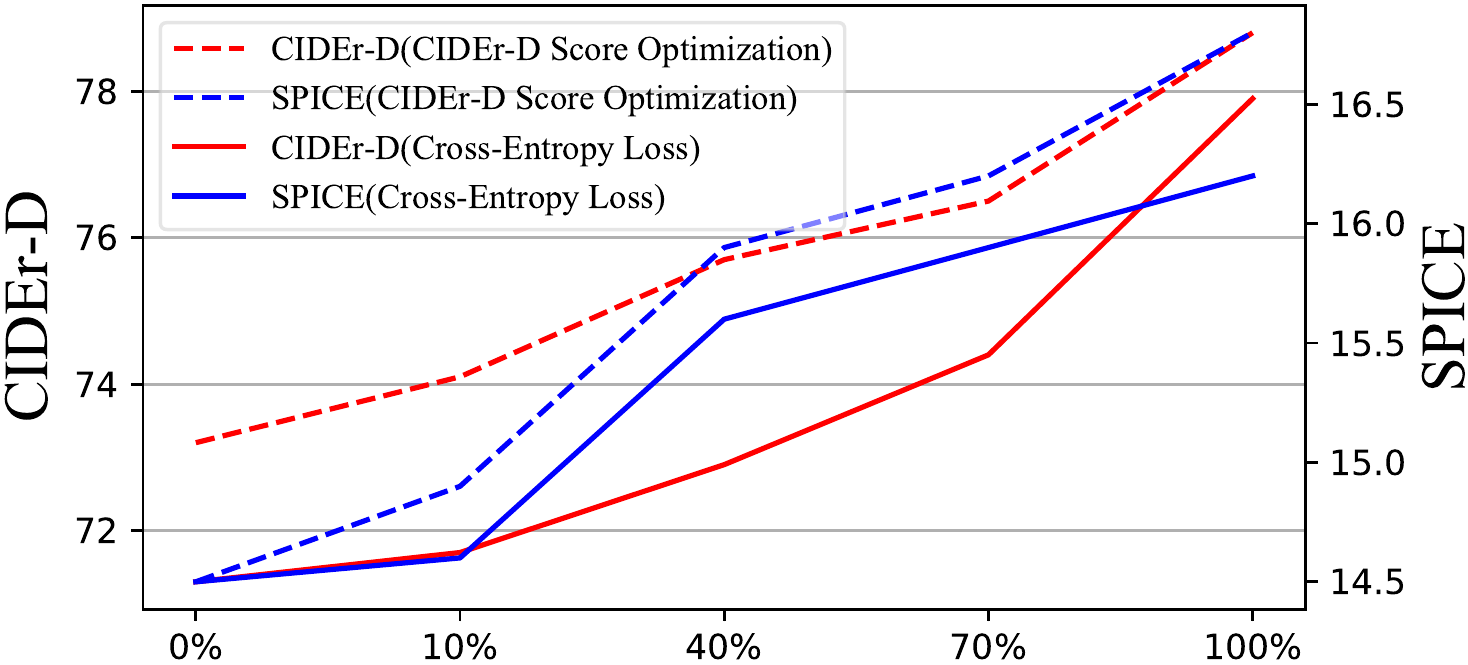}\\
		\end{minipage}
	\end{center}
	\caption{Relationship between captioning performance with different ratio of unsupervised data (CIDEr-D Score Optimization).}\label{fig:f2}
\end{figure}

\begin{figure*}[t]
	\centering
	\includegraphics[width=180mm]{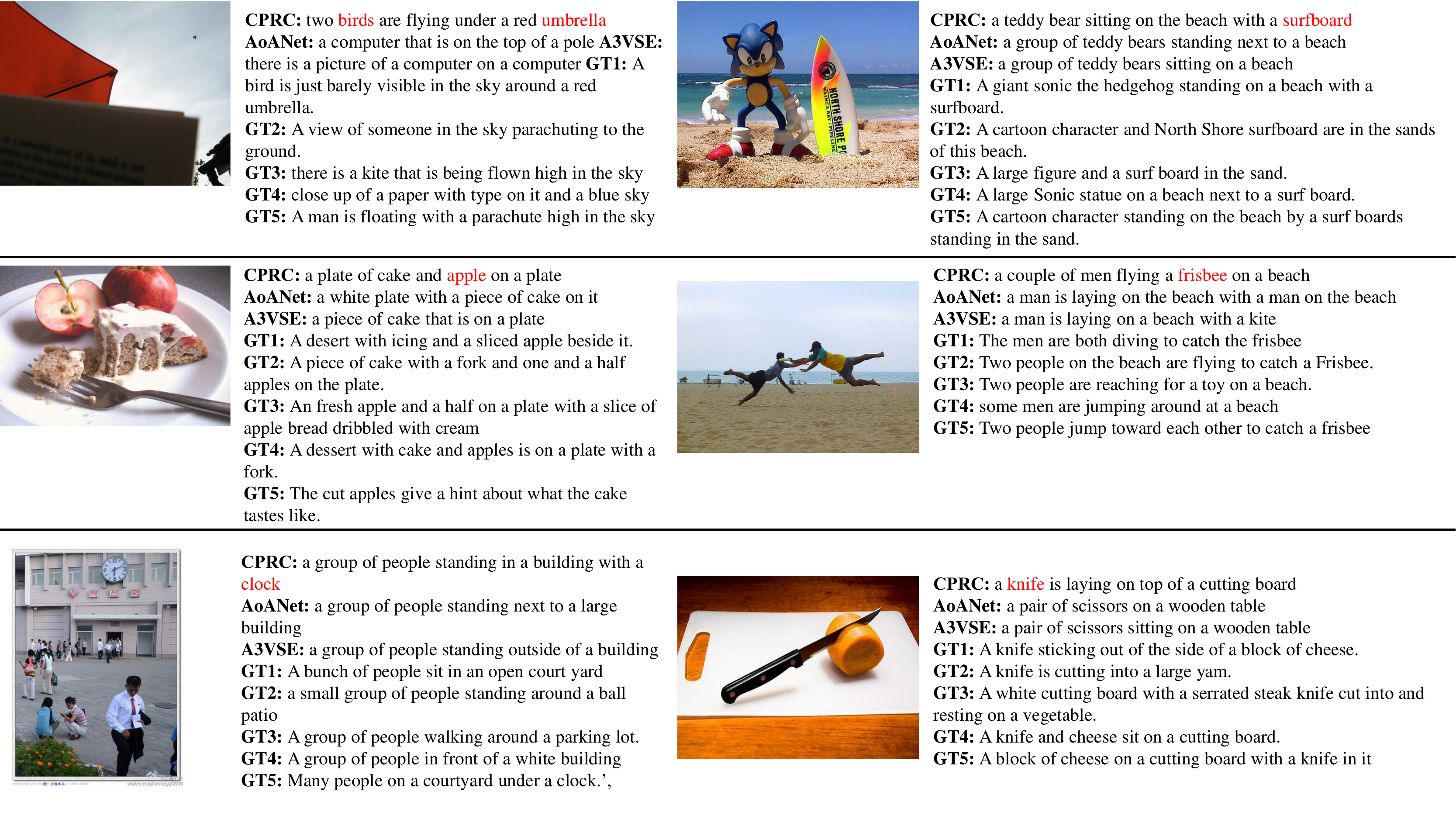}\\
	\includegraphics[width=180mm]{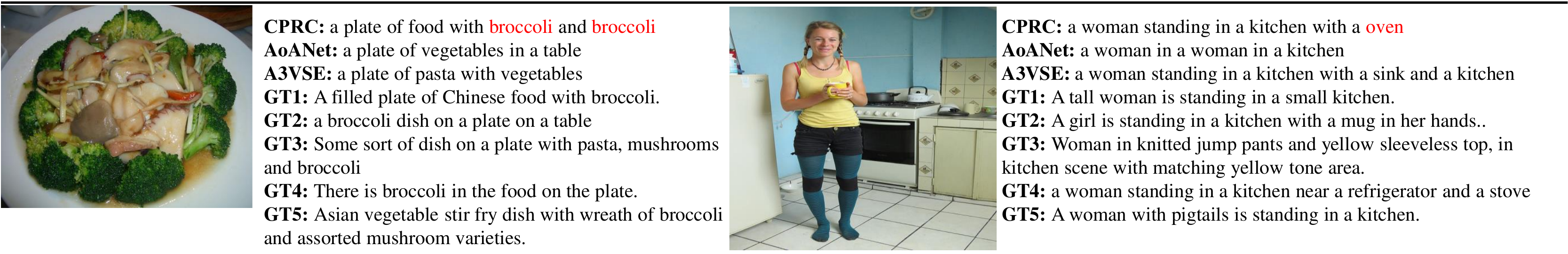}\\
	\caption{Examples of captions generated by CPRC and baseline models as well as the corresponding ground truths (GT1-GT5 are the 5 given ground-truth sentences).}\label{fig:f12}
\end{figure*}

\section{Experiments}\label{sec:s2}
\subsection{Datasets}
We adopt the popular MS COCO dataset~\cite{LinMBHPRDZ14} for evaluation, as former related methods are mostly practiced exclusively on this dataset ~\cite{HuangWCW19,HuangWXC19,HerdadeKBS19,ZhouWLHZ20,RennieMMRG17}. MS COCO dataset contains 123,287 images (82,783 training images and 40,504 validation images), each labeled with 5 captions. The popular test sets are divided into two categories: online evaluation and offline evaluation. Considering that all methods are evaluated under semi-supervised scenario, online evaluation cannot be used, so we only use offline evaluation. The offline “Karpathy” data split~\cite{KarpathyF17} contains 5,000 images for validation, 5,000 images for testing, and the rest for training. To construct the semi-supervised scenario, we randomly selected examples with artificially set proportions as supervised data from the training set, and the rest are unsupervised data.

\subsection{Implementation Details}
The target of CPRC is to train the generator $G$. In detail, we employ AoANet~\cite{HuangWCW19} structure for $G$ as base model. Meanwhile, we adopt fully connected networks for $f$ with three fully connected layers (with 1024 dimension for the hidden layers). The dimension of original image vectors is 2048 and we project them to a new space with the dimension of 1024 following~\cite{HuangWCW19}. The $K=3$, i.e., each image has three augmentations using random occlusion technique. As for the training process, we train AoANet for $40$ epochs with a mini-batch size of 16, and ADAM~\cite{KingmaB14} optimizer is used with a learning rate initialized by $10^{-4}$ and annealed by $0.8$ every $3$ epochs. The parameter $\lambda_1$ and $\lambda_2$ is tuned in $\{0.01, 0.1, 1, 10\}$, and $\tau = 0.1$. The entire network is trained on an Nvidia TITAN X GPU. 

\subsection{Baselines and Evaluation protocol}
The comparison models fall into three categories: 1) state-of-the-art supervised captioning methods: SCST~\cite{RennieMMRG17}, AoANet~\cite{HuangWCW19}, AAT~\cite{HuangWXC19}, ORT~\cite{HerdadeKBS19} and GIC~\cite{ZhouWLHZ20}. Note that these methods can only utilize the supervised image-sentence pairs. 2) state-of-the-art unsupervised captioning methods: Graph-align~\cite{GuJCZYW19} and UIC~\cite{Feng00L19a}. These approaches utilize the independent image set and corpus set for training. 3) state-of-the-art semi-supervised method: A3VSE~\cite{HuangKLCH19}. 

Moreover, we conduct extra ablation studies to evaluate each term in our proposed CPRC: 1) AoANet+P, we combine the label prediction consistency with the original AoANet generation loss as multi-task loss (only using the supervised data); 2) AoANet+C, we combine the relation consistency loss with the original AoANet generation loss as multi-task loss (only using the supervised data); 3) PL, we replace the prediction consistency with pseudo labeling as traditional semi-supervised methods; 4) AC, we replace the relation consistency with augmentation consistency as traditional semi-supervised methods; 5) Embedding+, we replace the relational consistency loss with embedding consistency loss, which minimizes the difference between the embedding of image inputs and generated sentences; 6) Semantic+, we replace the relational consistency loss with prediction consistency loss, which minimizes the difference between the predictions of image inputs and generated sentences; 7) Strong+, we replace the weak augmentation with strong augmentation for CPRC; 8) w/o Prediction, CPRC only retains the relation consistency loss in Eq. \ref{eq:e7}; 9) w/o Relation, CPRC only retains the prediction consistency in Eq. \ref{eq:e7}; and 10) w/o $\tau$, CPRC removes the confidence threshold as Eq. \ref{eq:e6}. For evaluation, we use different metrics, including BLEU~\cite{PapineniRWZ02}, METEOR~\cite{BanerjeeL05}, ROUGE-L~\cite{HuangWCW19}, CIDEr-D~\cite{VedantamZP15} and SPICE~\cite{AndersonFJG16}, to evaluate the proposed method and comparison methods. All the metrics are computed with the publicly released code\footnote{https://github.com/tylin/coco-caption}.

In fact, the CIDer-D and SPICE metric is more suitable for the image caption task~\cite{AndersonFJG16,VedantamZP15}. One of the problems with using metrics such as BlEU, ROUGE, CIDEr and METEOR is that these metrics are primarily sensitive to n-gram overlap. However, n-gram overlap is neither necessary nor sufficient for two sentences to convey the same meaning~\cite{GimenezM07a}. As shown in the example provided by~\cite{AndersonFJG16}, consider the following two captions (a,b) from the MS COCO dataset:
\begin{itemize}
	\item[(a)] A young girl standing on top of a tennis court.
	\item[(b)] A giraffe standing on top of a green field.
\end{itemize}
The captions describe two different images. However, the mentioned n-gram metrics produces a high similarity score due to the presence of the long 5-gram phrase ``standing on top of a'' in both captions. Meanwhile, the following captions (c,d) obtained from the same image:
\begin{itemize}
	\item[(c)] A shiny metal pot filled with some diced veggies.
	\item[(d)] The pan on the stove has chopped vegetables in it.
\end{itemize}
These captions convey almost the same meaning, whereas exhibit low n-gram similarity as they have no words in common. 

To solve this problem, SPICE~\cite{AndersonFJG16} estimated caption quality by transforming both candidate and reference captions into a graph-based semantic representation (i.e., scene graph). The scene graph can explicitly encodes the objects, attributes and relationships found in image captions, abstracting away most of the lexical and syntactic idiosyncrasies of natural language in the process. CIDer-D~\cite{VedantamZP15} measured the similarity of a candidate sentence to a majority of how most people describe the image (i.e. the reference sentences). 

\subsection{Qualitative Analysis}
Table \ref{tab:tab1} presents the quantitative comparison results with state-of-the-art methods (i.e., 1$\%$ supervised data and 99$\%$ unsupervised in the training set), it is notable that supervised captioning methods can only develop the mapping functions with supervised data, and leave out the unsupervised data. For fairness, all the models are first trained under cross-entropy loss and then optimized for CIDEr-D score as~\cite{HuangWCW19}. ``-'' represents the results have not given in the raw paper. The results reveal that: 1) AoANet achieves the best scores on most metrics compared with the existing supervised methods. Therefore, CPRC adopts AoANet as the base image-sentence mapping function. 2) Unsupervised approach, i.e., UIC, achieve the worst performance on all metrics under different loss. This verifies that the generated sentence may mismatch the image with a high probability when only considering the domain discriminator. Graph-align performs better than supervised approaches, but worse than A3VSE on most metrics, because it ignores to measure specific example matching. 3) Semi-Supervised method, i.e., A3VSE, has little effect on improving the captioning performance, e.g., cross-entropy loss/CIDEr-D score optimization only improves 0.4/2.0 and 0.2/0.1 on CIDEr-D and SPICE scores comparing with AoANet, because it is more difficult to ensure the quality of generated sentences. 4) CPRC achieves the highest scores among all compared methods in terms of all metrics, on both the cross-entropy loss and CIDEr-D score optimization stage, except ROUGE-L on cross-entropy loss. For example, CPRC achieves a state-of-the-art performance of 77.9/78.8 (CIDEr-D score) and 16.2/16.8 (SPICE score) under two losses (cross-entropy and CIDEr-D score), that acquires 8.7/8.4 and 1.9/1.6 improvements comparing with AoANet. The phenomena indicates that, with limited amount of supervised data, existing methods cannot construct a well mapping function, whereas CPRC can reliably utilize the undescribed image to enhance the model; and 5) CPRC performs better than w/o $\tau$ on all metrics, which indicates the effectiveness of threshold confidence. 

\subsection{Ablation Study}
To quantify the impact of proposed CPRC modules, we compare CPRC against other ablated models with various settings. The bottom half of Table \ref{tab:tab1} presents the results: 1) AoANet+P and AoANet+C achieve better performance than AoANet, which indicates that the prediction loss and relation consistency loss can improve the generator learning, because the labels can provide extra semantic information; meanwhile, AoANet+P performs better than AoANet+C on most metric, which indicates that prediction loss is more significant than relation consistency; 2) PL and AC perform worse than the w/o Prediction and w/o Relation, which verifies that traditional semi-supervised techniques considering pseudo labeling are not as good as cross-modal semi-supervised techniques considering raw image as pseudo supervision; 3) Embedding+ performs worse than the Semantic+, which reveals that embeddings are more difficult to compare than predictions since image and text have heterogeneous representations; 4) Strong+ performs worse than CPRC, which validates that the strong augmentation may impact the generated sentence, and further affect the prediction as well as causing the noise accumulation; 5) Both the w/o Prediction and w/o Relation can improve the captioning performance on most criteria, especially on the important criteria, i.e., CIDEr-D and SPICE. The results indicate that both the prediction and relation consistencies can provide effective supervision to ensure the quality of generated sentences; 6) The effect of w/o Relation is more obvious, which shows that prediction loss can further improve the scores by comprehensively considering the semantic information; and 7) CPRC achieves the best scores on most metrics, which indicates that it is better to combine the content and relation information.

\subsection{CPRC with Different Captioning Model}
\begin{table}[htb]{
		\centering
		\caption{Performance of CPRC with different caption model on MS-COCO “Karpathy” test split, where B$@$N, M, R, C and S are short for BLEU@N, METEOR, ROUGE-L, CIDEr-D and SPICE scores. }
		\label{tab:tab2}
		\begin{tabular}{@{}l@{}|@{}c|@{}c|c|c|c|c|c|c}
			\toprule
			\multirow{2}{*}{Methods} & \multicolumn{8}{c}{Cross-Entropy Loss}  \\
			\cmidrule(l){2-9}
			& B$@$1 & B$@$2 & B$@$3 & B$@$4 & M & R & C & S \\
			\midrule
			SCST &56.8&38.6&25.4&16.3&16.0 &42.4&38.9&9.3\\
			GIC &63.0&46.8&33.2 &20.0&19.2&50.3&50.5&12.3\\
			\midrule
			SCST+CPRC &\bf 63.5&\bf 45.9&\bf 31.7&\bf 21.6&\bf 19.4&\bf 45.8& \bf48.1&\bf 10.2\\
			GIC+CPRC &\bf 66.8&\bf 47.5&\bf 34.5&\bf 21.4&\bf 19.2&\bf 50.8&\bf 57.7&\bf 13.4\\
			\midrule
			\multirow{2}{*}{Methods}  & \multicolumn{8}{c}{CIDEr-D Score Optimization} \\
			\cmidrule(l){2-9}
			& B$@$1 & B$@$2 & B$@$3 & B$@$4 & M & R & C & S \\
			\midrule
			SCST &59.4&39.5&25.3 &16.3&17.0&42.9&43.7&9.9\\
			GIC &64.7&46.9&32.0&20.7&19.0 &47.8&55.7&12.5\\
			\midrule
			SCST+CPRC &\bf66.5&\bf48.0&\bf33.7&\bf22.7&\bf20.4&\bf47.9&\bf48.7&\bf10.7\\
			GIC+CPRC &\bf66.9&\bf47.9&\bf34.8&\bf21.8&\bf19.8&\bf48.2&\bf58.9&\bf13.6\\
			\bottomrule
	\end{tabular}}
\end{table}
To explore the generality of CPRC, we conduct more experiments by incorporating CPRC with different supervised captioning approaches, i.e., SCST (encoder-decoder based model), GIC (attention based model). Note that we have not adopted the editing based method considering the reproducibility, the results are recorded in Table \ref{tab:tab11}. We find that all the methods, i.e., SCST, GIC and AoANet (results can refer to the Table \ref{tab:tab1}), have improved the performance after combing the CPRC framework. This phenomena validates that CPRC can well combine the undescribed images for existing supervised captioning models.

\subsection{Influence of the Supervised and Unsupervised Images}
To explore the influence of supervised data, we tune the ratio of supervised data, and the results are recorded in Figure \ref{fig:f1} and Figure \ref{fig:f11} with different metrics. Here, we find that with the percentage of supervised data increase, the performance of CPRC improves faster than other state-of-the-art methods. This indicates that CPRC can reasonably utilize the undescribed images to improve the learning of generator. Furthermore, we validate the influence of unsupervised data, i.e., we fix the supervised ratio to 1$\%$, and tune the ratio of unsupervised data in $\{10\%, 40\%, 70\%, 100\%\}$, the results are recorded in Figure \ref{fig:f2}. Note that one of the problems by using metrics, such as BlEU, ROUGE, CIDEr-D and METEOR to evaluate captions, is that these metrics are primarily sensitive to n-gram overlap~\cite{HuangWCW19,AndersonFJG16}. Therefore, we only give the results of CIDer-D and SPICE here (refer to the supplementary for more details). We find that with the percentage of unsupervised data increases, the performance of CPRC also improves. This indicates that CPRC can make full use of undescribed images for positive training.

\begin{table}[htb]{
		\centering
		\caption{Performance of CPRC with different augmentation number on MS-COCO “Karpathy” test split, where B$@$N, M, R, C and S are short for BLEU@N, METEOR, ROUGE-L, CIDEr-D and SPICE scores. }
		\label{tab:tab3}
		\begin{tabular}{@{}l@{}|c|c|c|c|c|c|c|c}
			\toprule
			\multirow{2}{*}{Methods} & \multicolumn{8}{c}{Cross-Entropy Loss}  \\
			\cmidrule(l){2-9}
			& B$@$1 & B$@$2 & B$@$3 & B$@$4 & M & R & C & S \\
			\midrule
			K=1 &67.5&48.9&34.6&22.5&21.1&48.4&74.7&15.5\\
			K=2 &67.8&49.5&34.9&23.4&21.7&49.5&75.9&15.8\\
			K=3 &\bf 68.8&\bf51.1&\bf35.5&\bf24.9&\bf22.8&\bf50.4&\bf77.9&\bf16.2\\
			K=4 &67.9&49.8&34.8&24.2&22.2&50.1&76.8&16.0\\
			K=5 &67.6&49.7&34.5&23.8&22.0&49.8&76.2&16.0\\
			\midrule
			\multirow{2}{*}{Methods}  & \multicolumn{8}{c}{CIDEr-D Score Optimization} \\
			\cmidrule(l){2-9}
			& B$@$1 & B$@$2 & B$@$3 & B$@$4 & M & R & C & S \\
			\midrule
			K=1 &68.0&50.1&35.7&24.8&22.0&49.5&77.1&16.1\\
			K=2 &68.3&50.5&35.9&25.3&22.1&49.7&77.7&16.5\\
			K=3 &\bf69.9&\bf51.8&\bf36.7&\bf25.5&\bf23.4&\bf50.7&\bf78.8&\bf16.8\\
			K=4 &68.7&51.4&36.5&25.2&22.8&49.7&77.4&16.3\\
			K=5 &68.3&50.8&35.9&25.1&22.7&49.4&77.3&16.2\\
			\bottomrule
	\end{tabular}}
\end{table}

\subsection{Influence of the Augmentation Number}
To explore the influence of augmentation number, i.e., $K$, we conduct more experiments. In detail, we tune the $K$ in $\{1,2,3,4,5\}$ and recorded the results in Table \ref{tab:tab3}. The results reveal that the CPRC achieves the best performance with $K = 3$, for the reason that additional inconsistent noises between image and sentence may be introduced with the the number of augmentations increase.

\begin{table}[htb]{
		\centering
		\caption{Performance of CPRC with different $\tau$ on MS-COCO “Karpathy” test split, where B$@$N, M, R, C and S are short for BLEU@N, METEOR, ROUGE-L, CIDEr-D and SPICE scores. }
		\label{tab:tab4}
		\begin{tabular}{@{}l|c|c|c|c|c|c|c|c}
			\toprule
			\multirow{2}{*}{Methods} & \multicolumn{8}{c}{Cross-Entropy Loss}  \\
			\cmidrule(l){2-9}
			& B$@$1 & B$@$2 & B$@$3 & B$@$4 & M & R & C & S \\
			\midrule
			$\tau = 0$ &66.9&49.8&34.5&24.2&21.5&49.5&76.2&15.4\\
			$\tau = 0.1$ &\bf68.8&\bf51.1&\bf35.5&\bf24.9&\bf22.8&\bf50.4&\bf77.9&\bf16.2\\
			$\tau = 0.4$ &66.4&49.5&34.3&24.0&21.1&48.8&75.8&15.2\\
			$\tau = 0.7$ &64.2&48.1&33.4&22.9&20.4&46.5&73.3&15.0\\
			\midrule
			\multirow{2}{*}{Methods}  & \multicolumn{8}{c}{CIDEr-D Score Optimization} \\
			\cmidrule(l){2-9}
			& B$@$1 & B$@$2 & B$@$3 & B$@$4 & M & R & C & S \\
			\midrule
			$\tau = 0$ &68.5 &50.8&36.2&25.0&22.5&49.8&77.5&16.2\\
			$\tau = 0.1$ &\bf69.9&\bf51.8&\bf36.7&\bf25.5&\bf23.4&\bf50.7&\bf78.8&\bf16.8\\
			$\tau = 0.4$ &68.4&50.2&36.1&24.8&22.1&49.5&77.1&16.1\\
			$\tau = 0.7$ &64.8&48.6&34.2&23.5&20.8&47.3&73.7&15.1\\
			\bottomrule
	\end{tabular}}
\end{table}

\subsection{Influence of the Confidence Threshold}
To explore the influence of confidence threshold, i.e., $\tau$, we conduct more experiments. In detail, we tune the $\tau$ in $\{0, 0.1,0.4,0.7\}$ and recorded the results in Table \ref{tab:tab4}. The results reveal that the performance of CPRC increases firstly, then decreases with the increasing of $\tau$. The reason is that fewer undescribed images are used with the increasing of $\tau$, thereby the generator training has not fully explored the unsupervised data.

\subsection{Visualization and Analysis}
Figure~\ref{fig:f3} shows a few examples with captions generated by our CPRC and two baselines, A3VSE and AoANet, as well as the human-annotated ground truths. From these examples, we find that the generated captions of baseline models lack the logic of language and lose accurate for the image content, while CPRC can generate accurate captions in high quality. 

Figure~\ref{fig:f4} shows an example of augmented images and corresponding generated captions. From these examples, we find that the generated captions basically have similar semantic information, which can help the prediction and relation consistencies for the undescribed images.

\section{Influence of Label Prediction}
To explore the effect of prediction loss, we conduct more experiments and exhibit several cases. Figure~\ref{fig:f12} shows a few examples with captions generated by our CPRC and two baselines, A3VSE and AoA, as well as the human-annotated ground truths. From these examples, we find that the generated captions of baseline models lack the logic of language and inaccurate for the image content, while CPRC can generate accurate captions in high quality. Meanwhile, it can be clearly seen that the label prediction helps the generator to understand the image from the red part of the sentence generated by CPRC, for example, in figure~\ref{fig:f12} (a), the content of the image is complicated and the part of bird is not obvious, which causes the sentences generated by AoANet and A3VSE inconsistent with the ground-truths. But CPRC can generate a good description of ``bird'' and ``umbrella'' by combining label prediction information.

\begin{figure*}[htb]
	\begin{center}
		\begin{minipage}[h]{40mm}
			\centering
			\includegraphics[width=40mm]{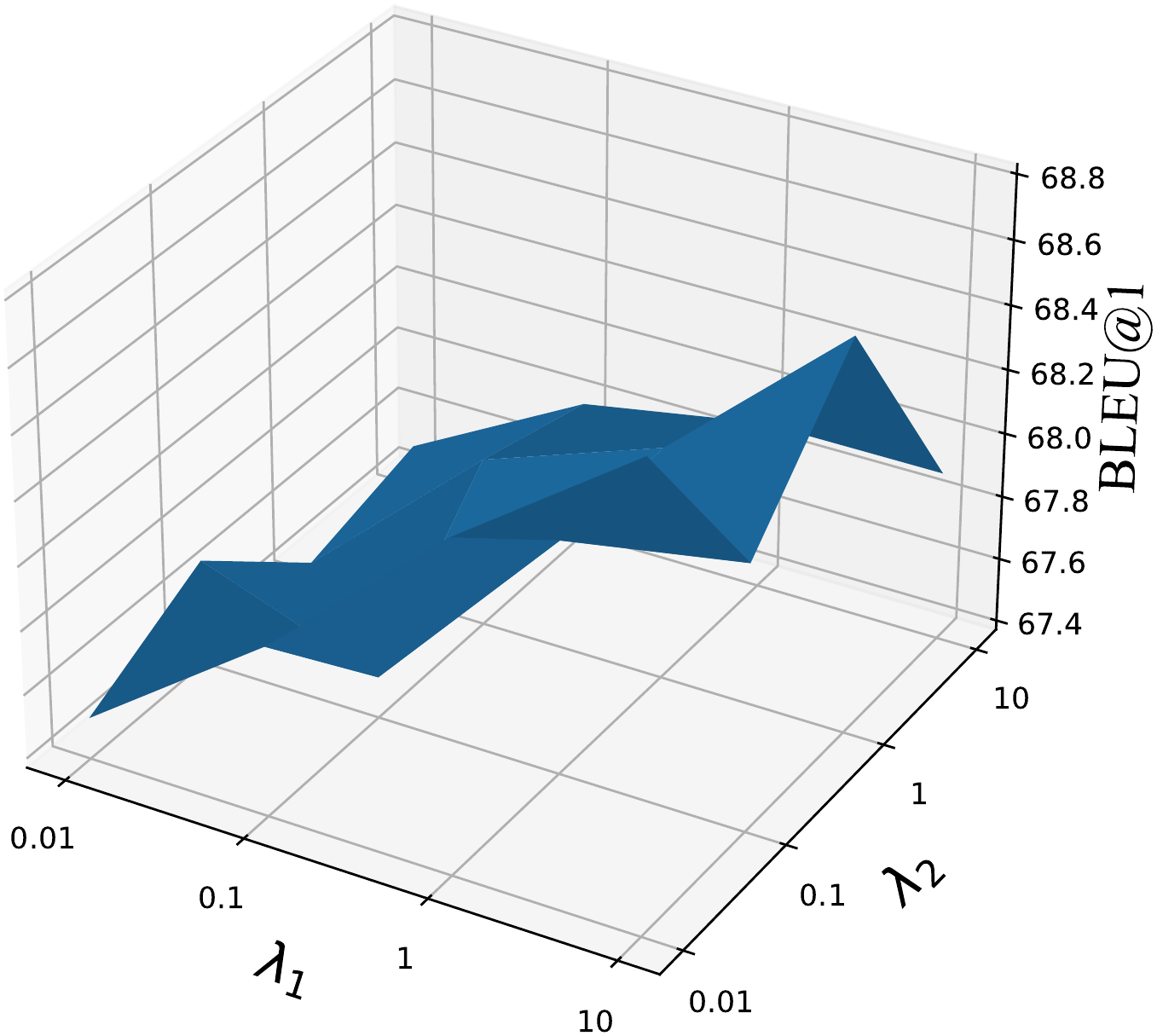}\\
			\mbox{ \;\;\;\; ({\it a}) {BLEU$@$1}}
		\end{minipage}
		\begin{minipage}[h]{40mm}
			\centering
			\includegraphics[width=40mm]{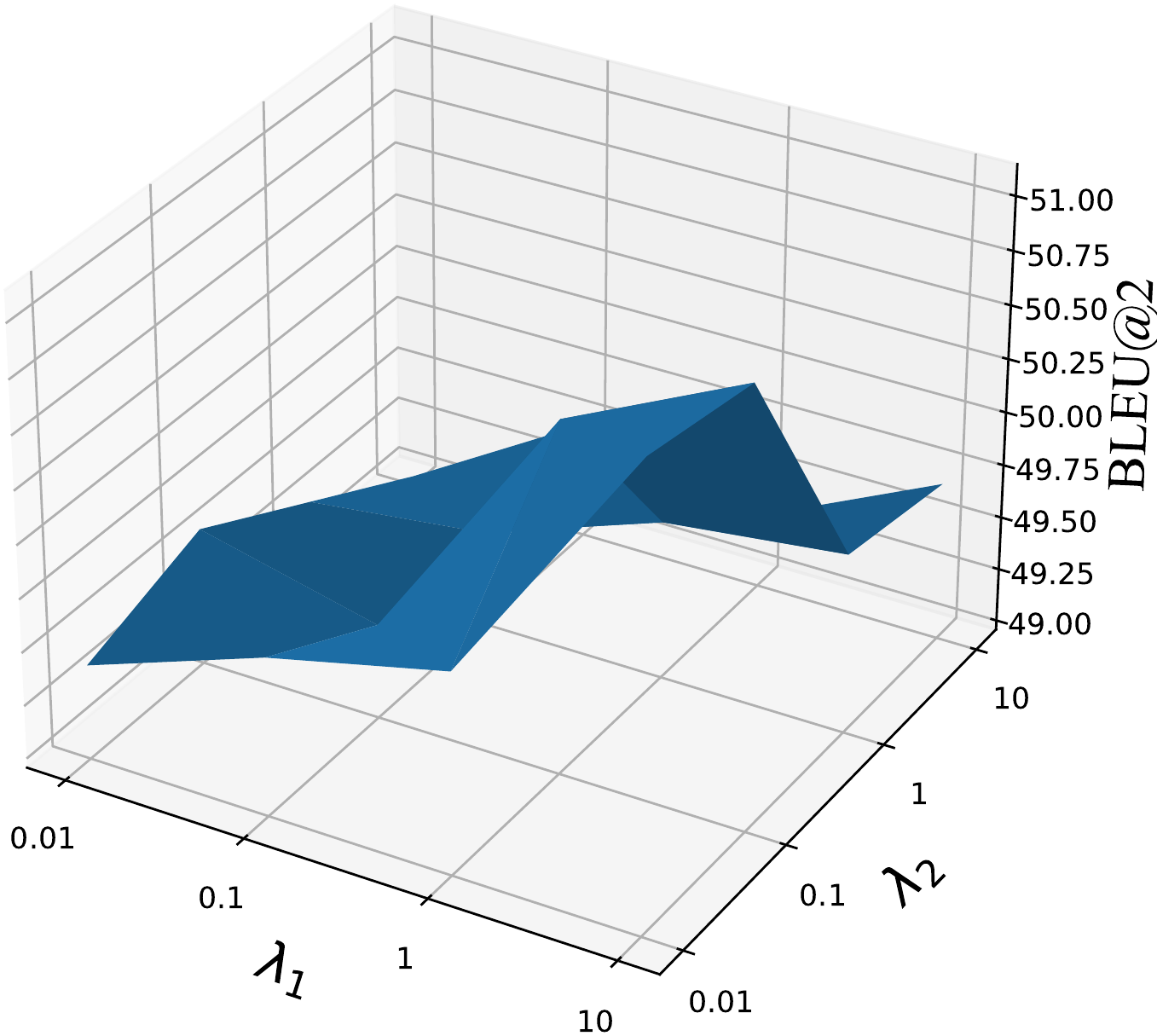}\\
			\mbox{ \;\;\;\; ({\it b}) {BLEU$@$2}}
		\end{minipage}
		\begin{minipage}[h]{40mm}
			\centering
			\includegraphics[width=40mm]{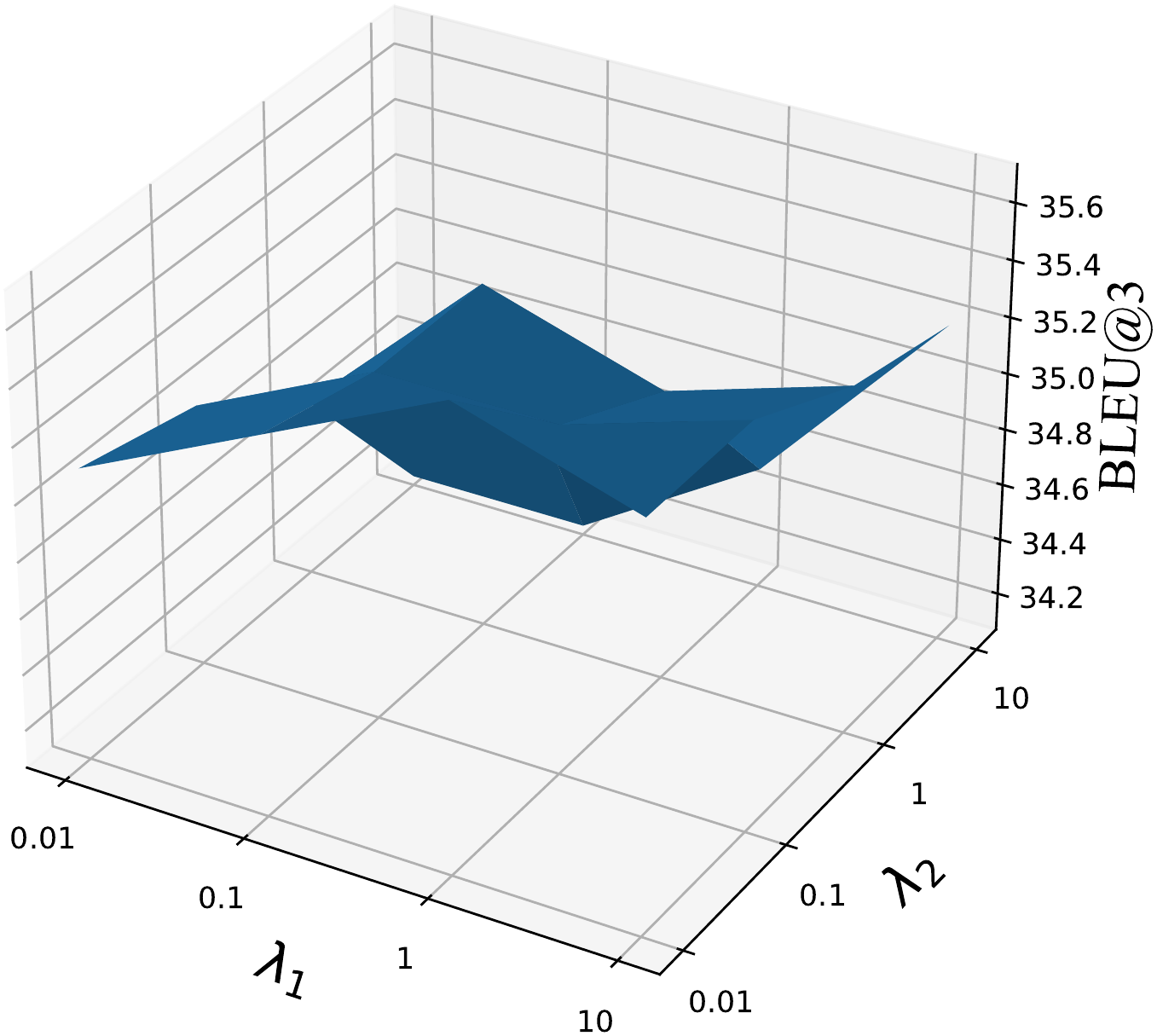}\\
			\mbox{ \;\;\;\; ({\it c}) {BLEU$@$3}}
		\end{minipage}
		\begin{minipage}[h]{40mm}
			\centering
			\includegraphics[width=40mm]{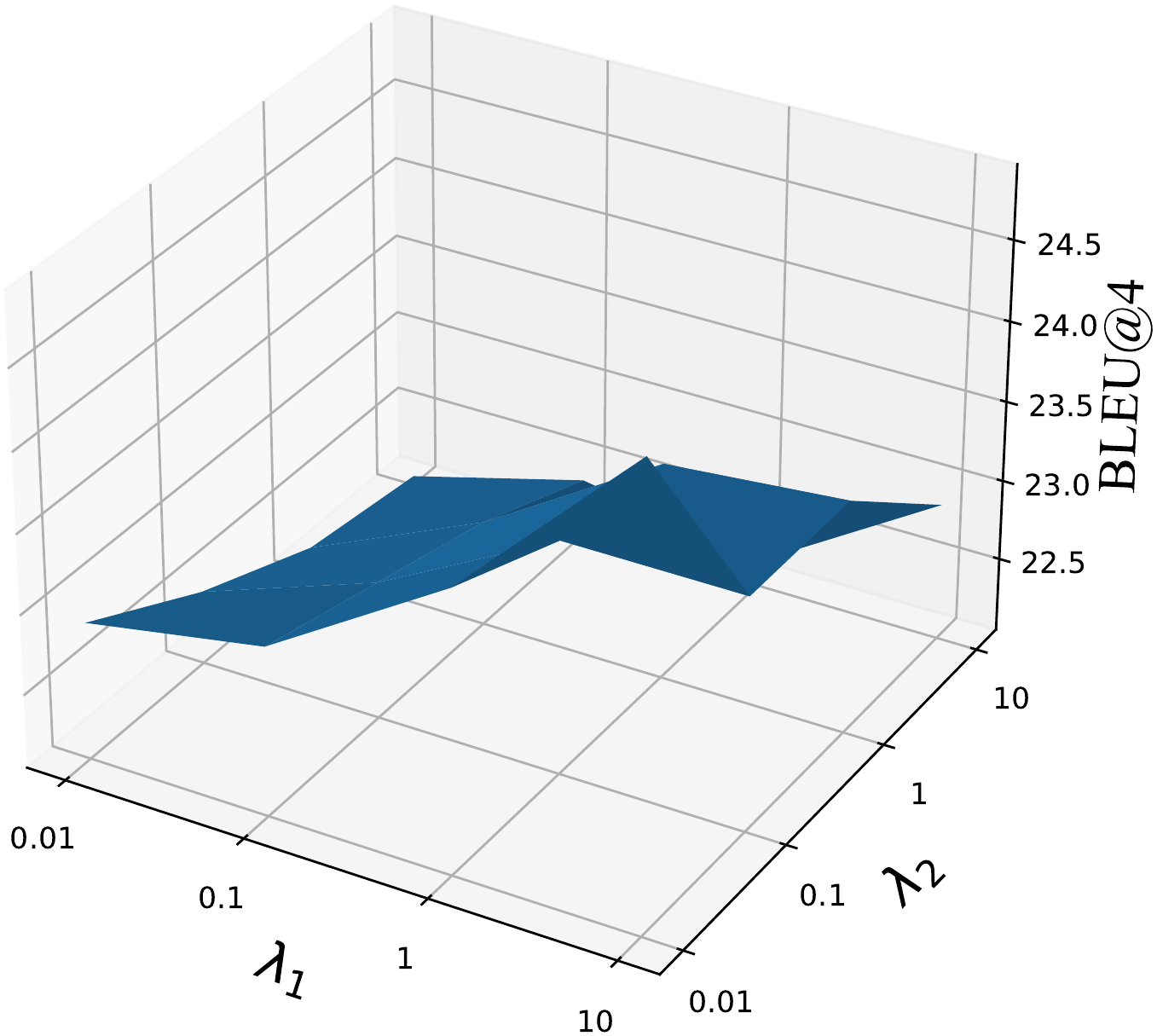}\\
			\mbox{ \;\;\;\; ({\it d}) {BLEU$@$4}}
		\end{minipage}\\
		\begin{minipage}[h]{40mm}
			\centering
			\includegraphics[width=40mm]{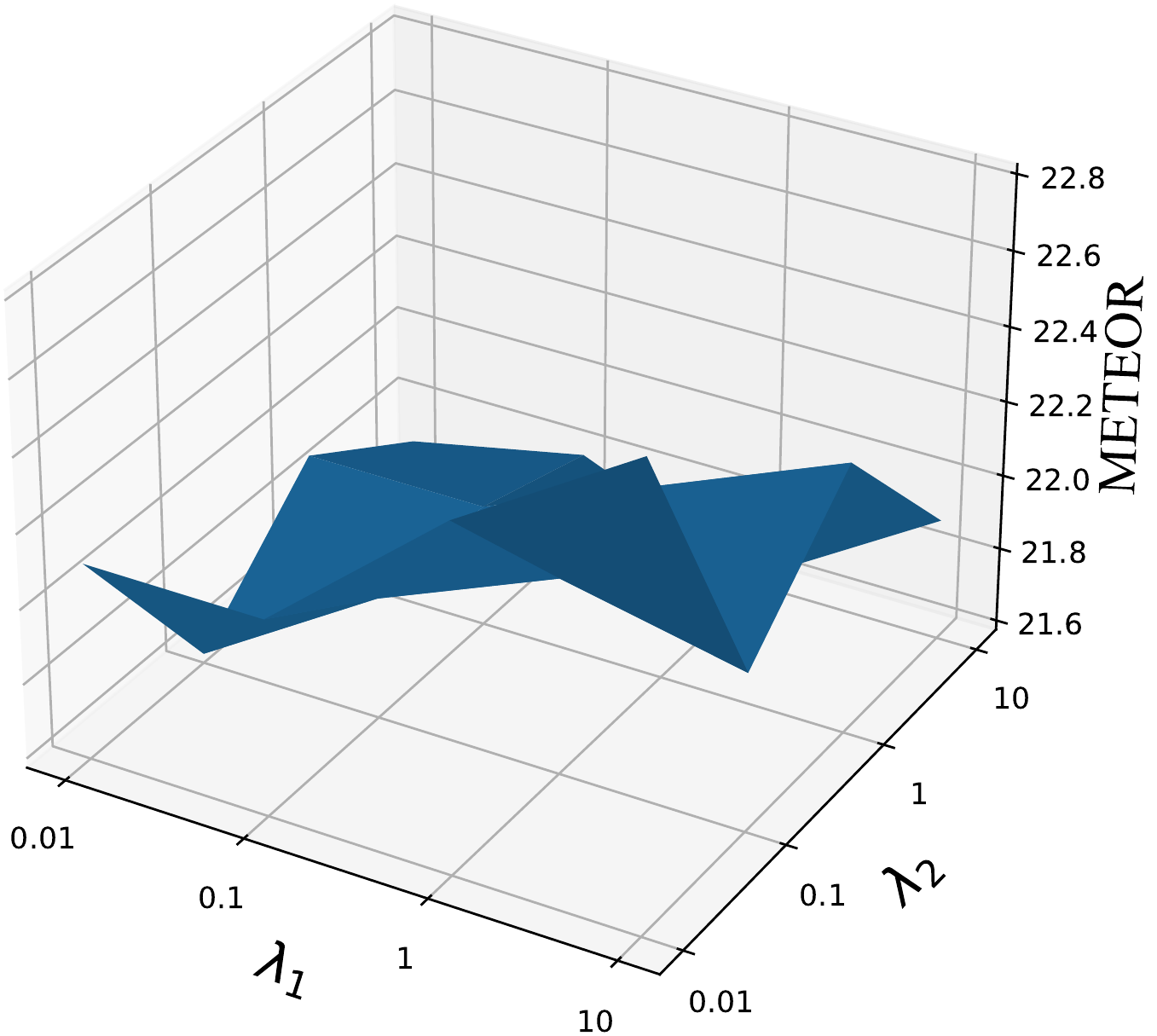}\\
			\mbox{ \;\;\;\; ({\it e}) {METEOR}}
		\end{minipage}
		\begin{minipage}[h]{40mm}
			\centering
			\includegraphics[width=40mm]{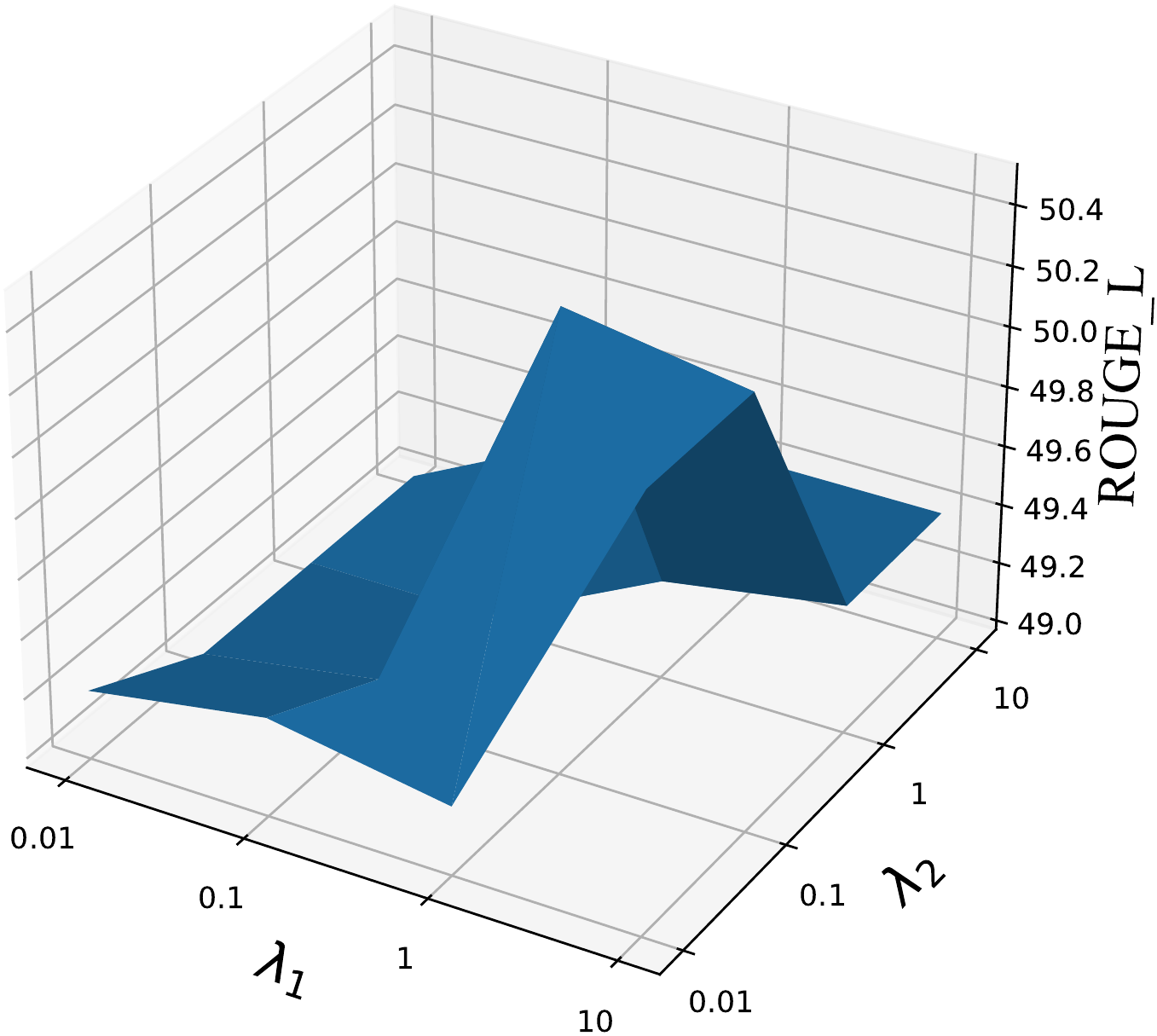}\\
			\mbox{ \;\;\;\; ({\it f}) {ROUGE-L}}
		\end{minipage}
		\begin{minipage}[h]{40mm}
			\centering
			\includegraphics[width=40mm]{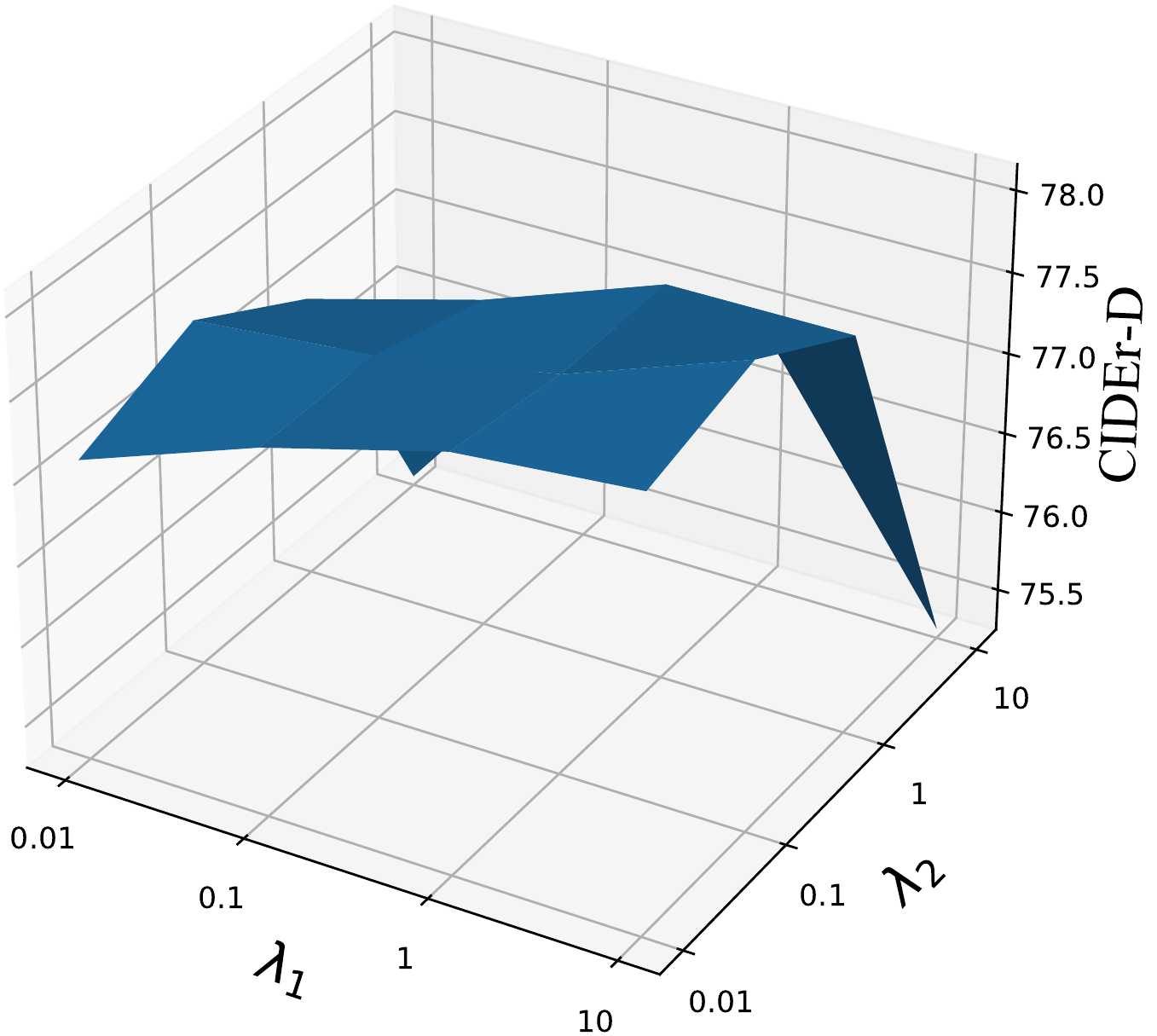}\\
			\mbox{ \;\;\;\; ({\it g}) {CIDEr-D}}
		\end{minipage}
		\begin{minipage}[h]{40mm}
			\centering
			\includegraphics[width=40mm]{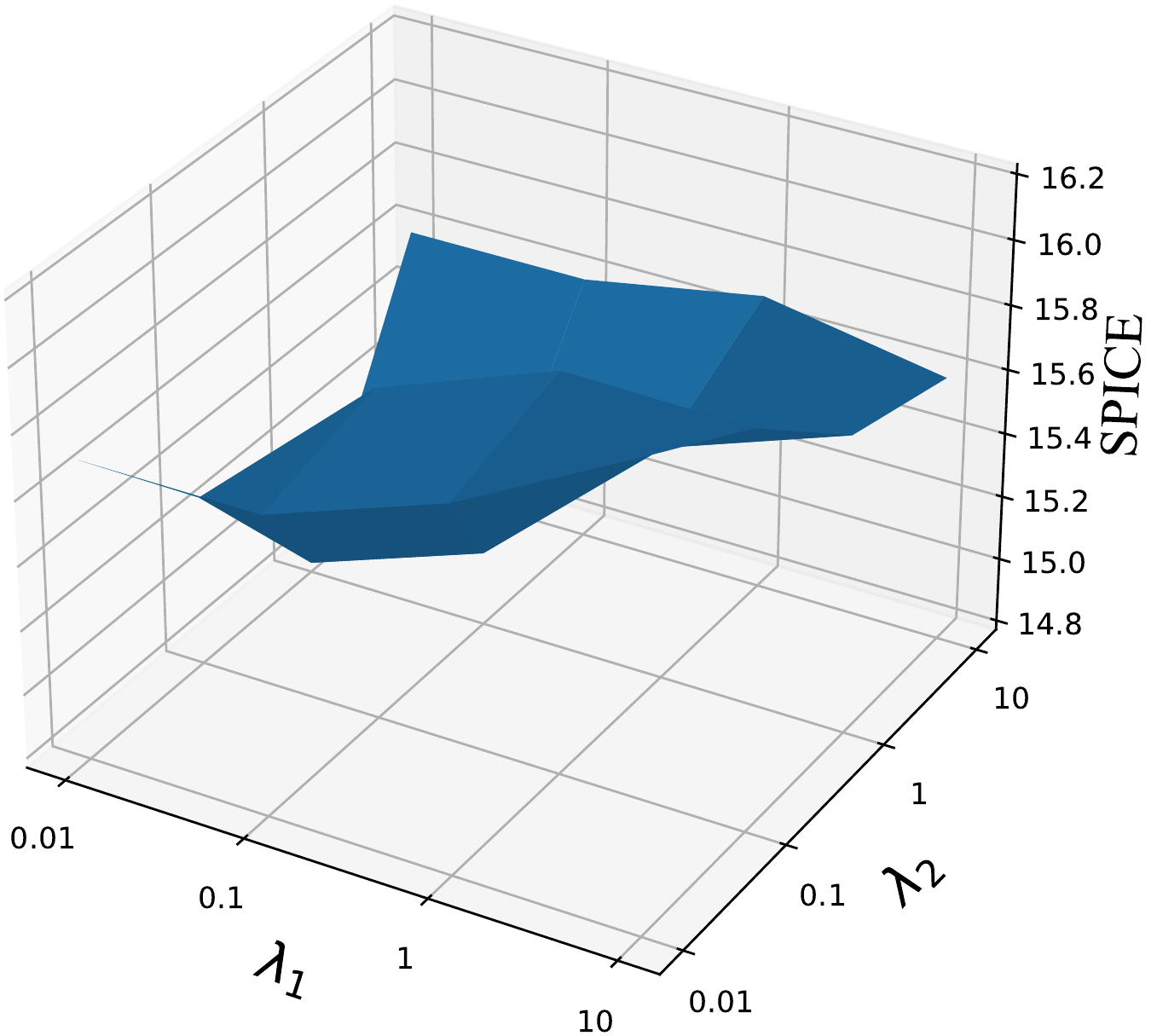}\\
			\mbox{ \;\;\;\; ({\it h}) {SPICE}}
		\end{minipage}
	\end{center}
	\caption{Parameter sensitivity of $\lambda_1$ and $\lambda_2$ with Cross-Entropy Loss.}\label{fig:f13}
	\begin{center}
		\begin{minipage}[h]{40mm}
			\centering
			\includegraphics[width=40mm]{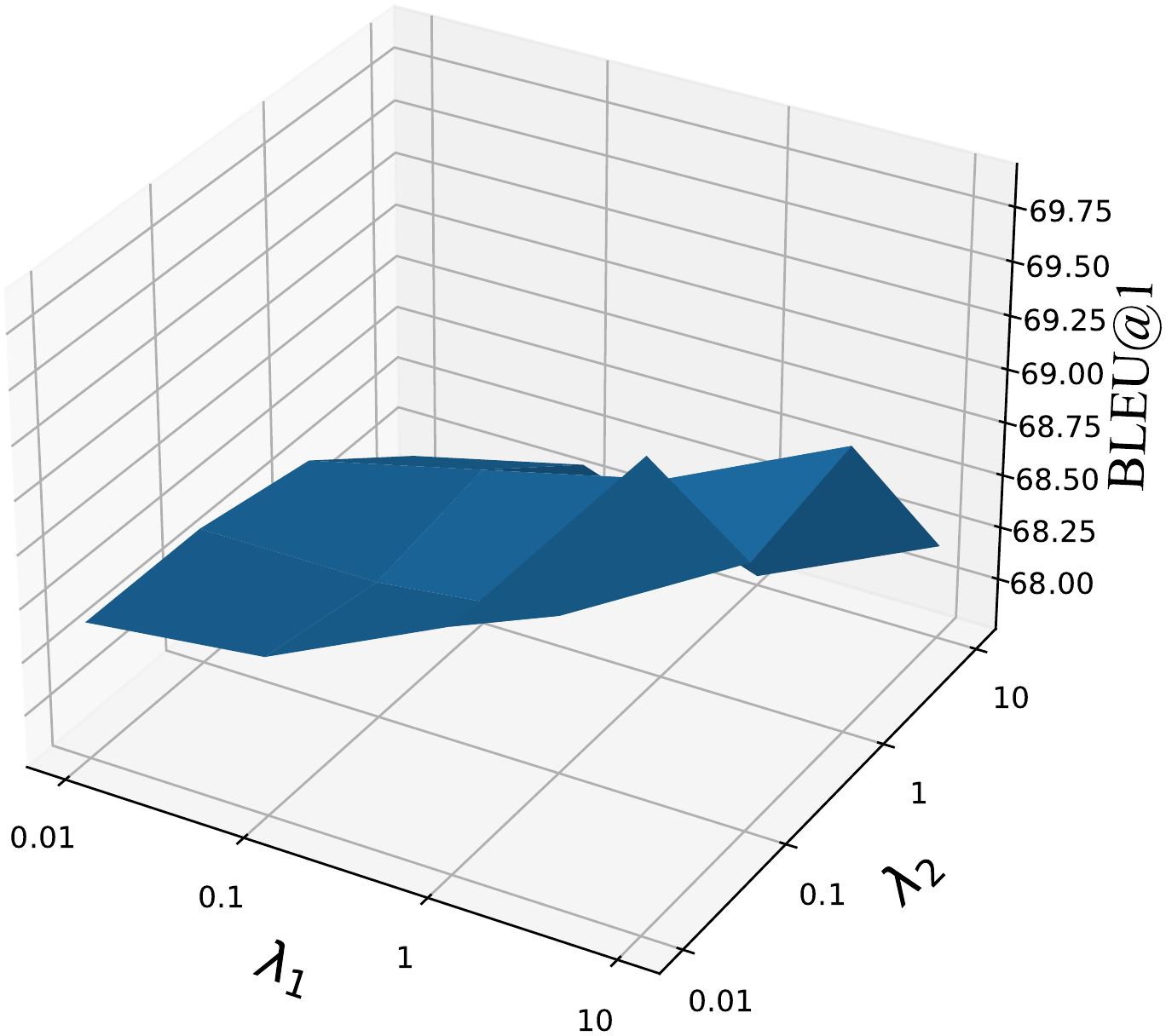}\\
			\mbox{ \;\;\;\; ({\it a}) {BLEU$@$1}}
		\end{minipage}
		\begin{minipage}[h]{40mm}
			\centering
			\includegraphics[width=40mm]{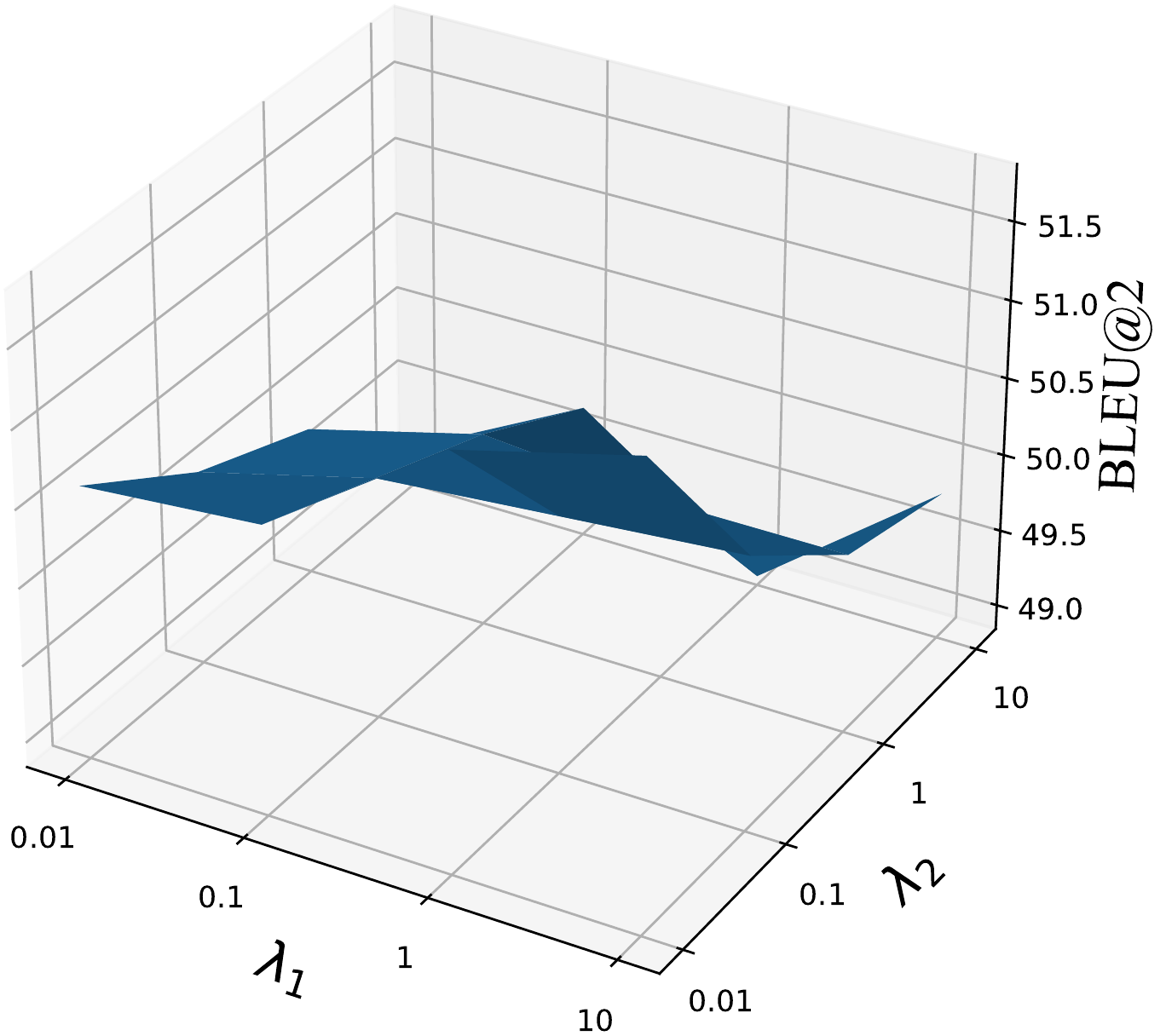}\\
			\mbox{ \;\;\;\; ({\it b}) {BLEU$@$2}}
		\end{minipage} 
		\begin{minipage}[h]{40mm}
			\centering
			\includegraphics[width=40mm]{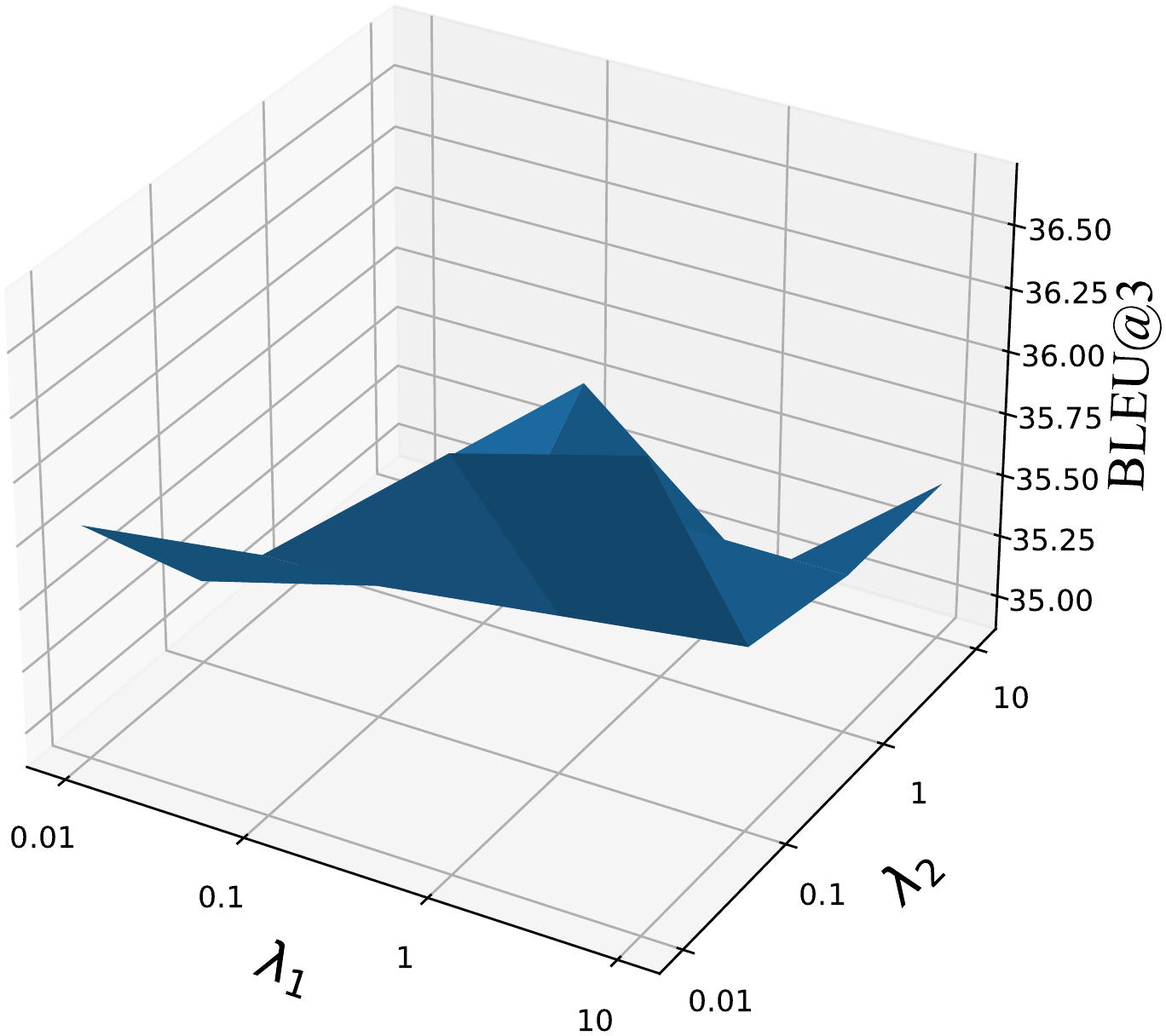}\\
			\mbox{ \;\;\;\; ({\it c}) {BLEU$@$3}}
		\end{minipage}
		\begin{minipage}[h]{40mm}
			\centering
			\includegraphics[width=40mm]{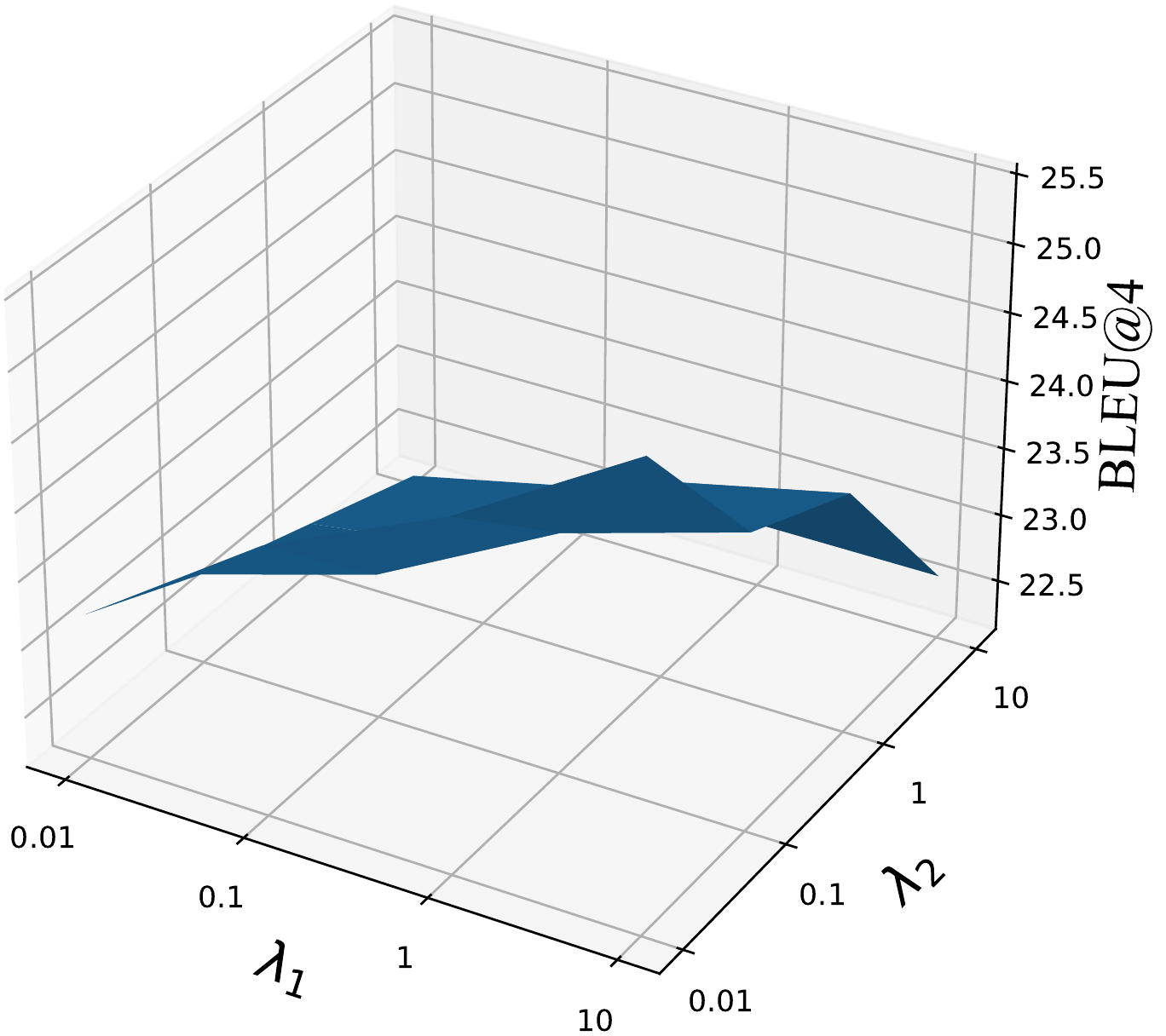}\\
			\mbox{ \;\;\;\; ({\it d}) {BLEU$@$4}}
		\end{minipage}\\
		\begin{minipage}[h]{40mm}
			\centering
			\includegraphics[width=40mm]{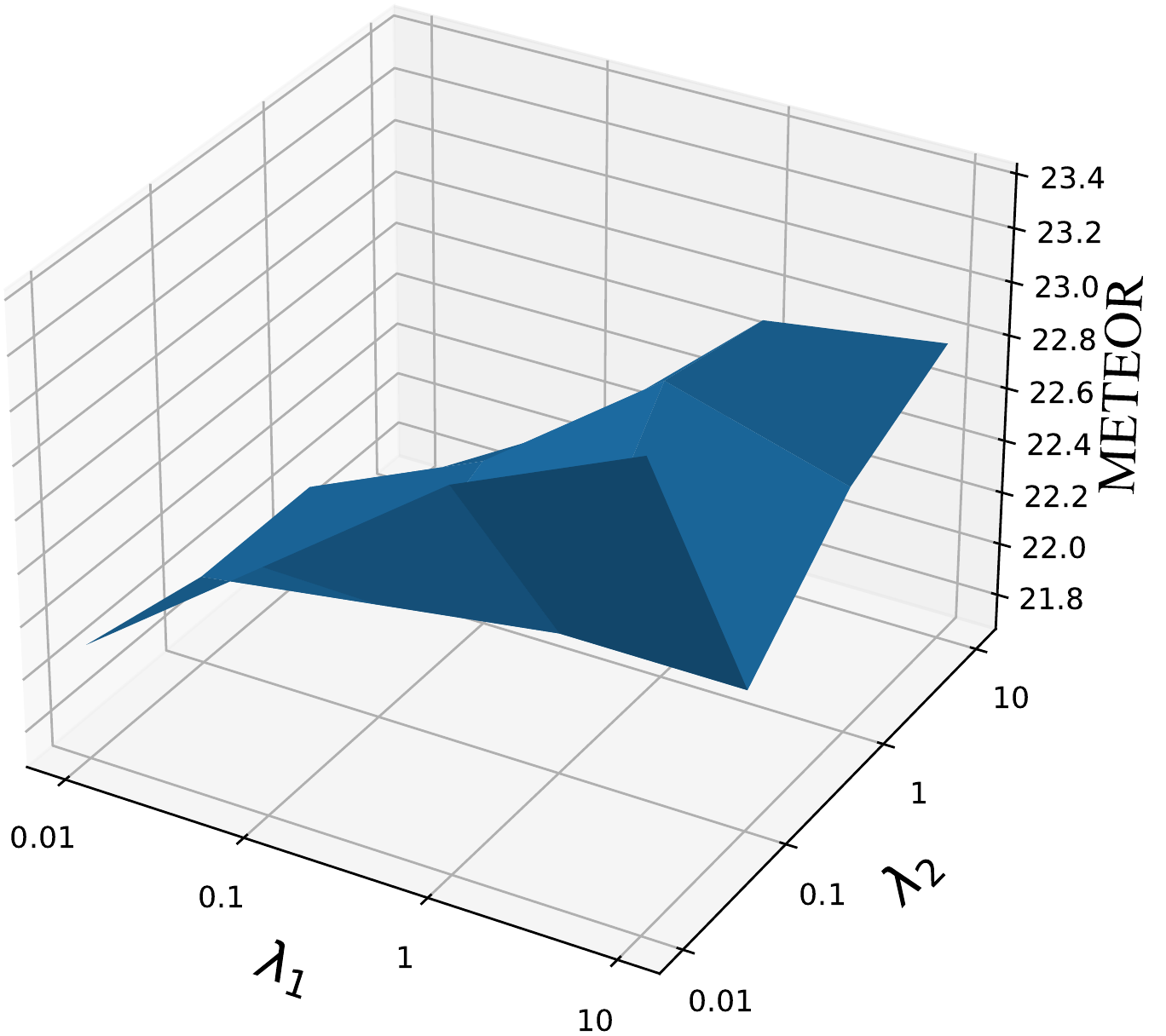}\\
			\mbox{ \;\;\;\; ({\it e}) {METEOR}}
		\end{minipage}
		\begin{minipage}[h]{40mm}
			\centering
			\includegraphics[width=40mm]{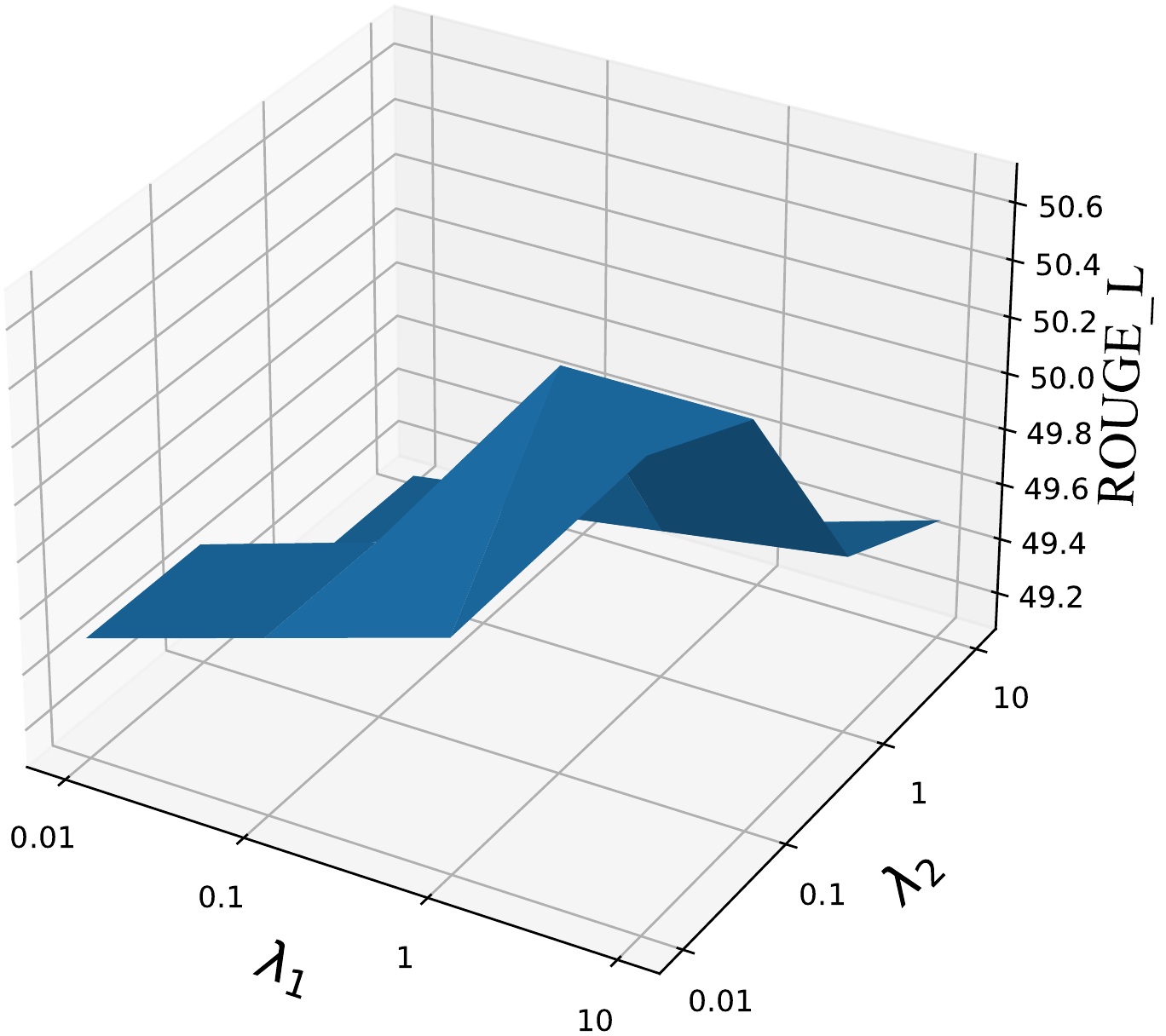}\\
			\mbox{ \;\;\;\; ({\it f}) {ROUGE-L}}
		\end{minipage}
		\begin{minipage}[h]{40mm}
			\centering
			\includegraphics[width=40mm]{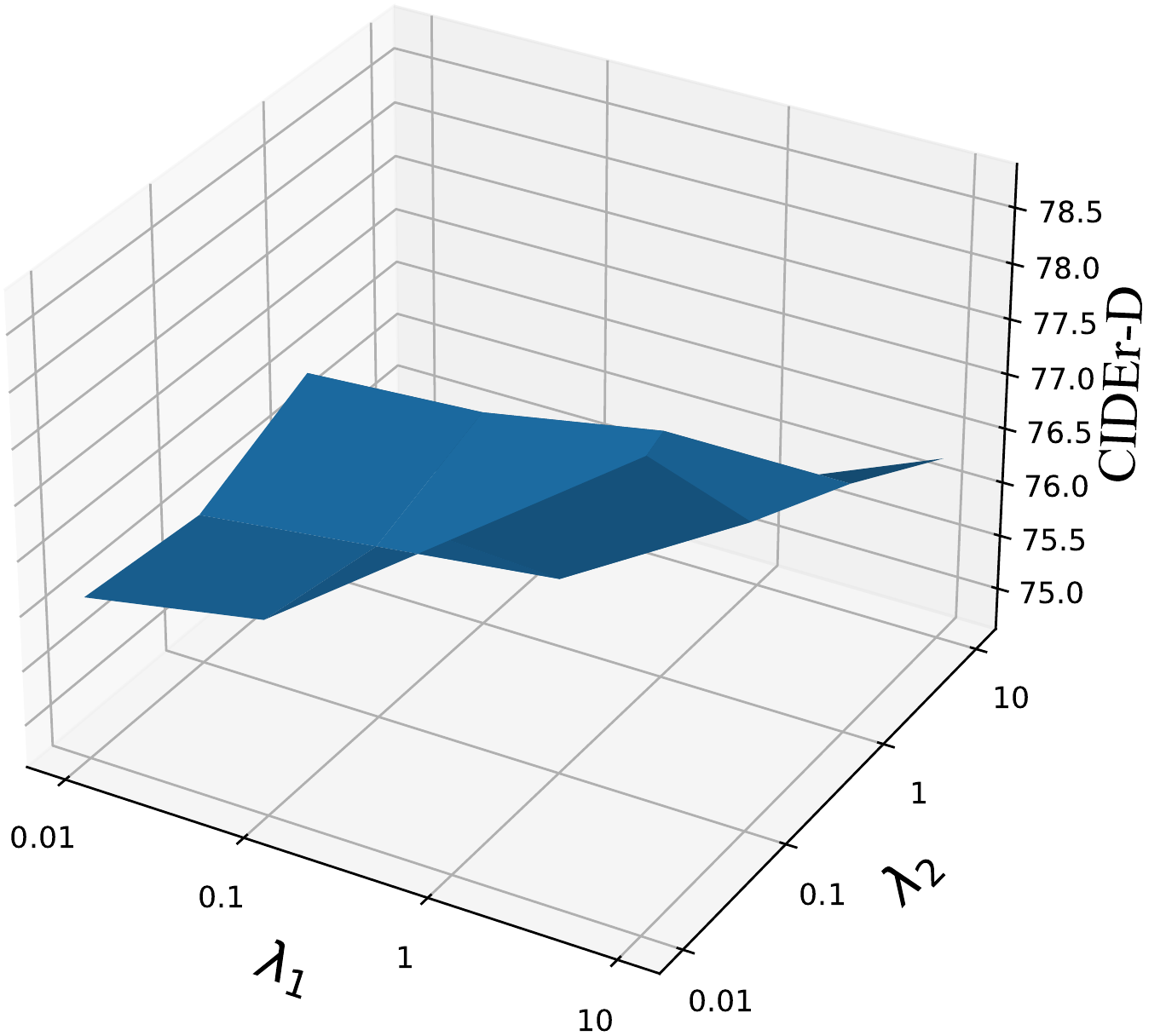}\\
			\mbox{ \;\;\;\; ({\it g}) {CIDEr-D}}
		\end{minipage}
		\begin{minipage}[h]{40mm}
			\centering
			\includegraphics[width=40mm]{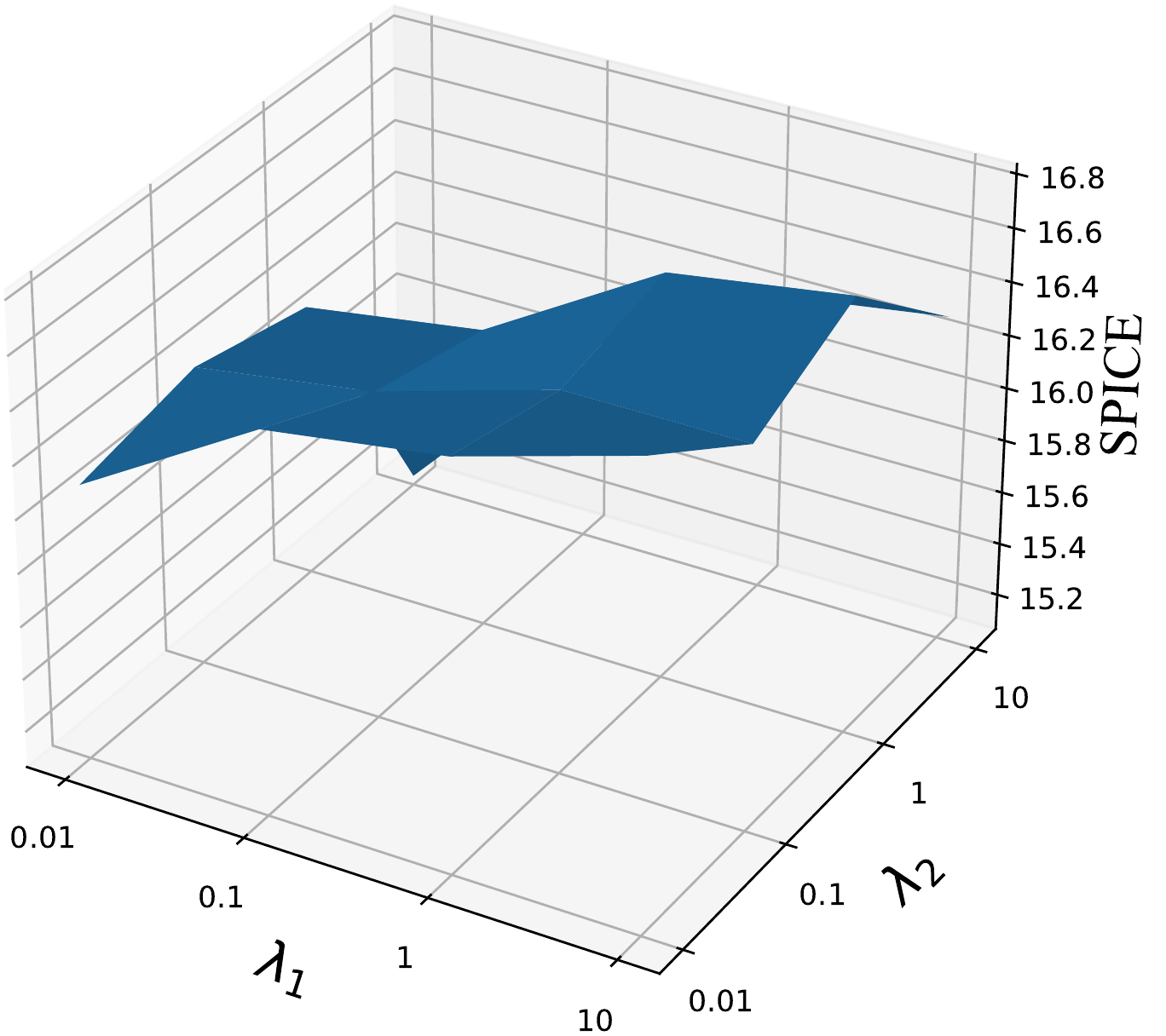}\\
			\mbox{ \;\;\;\; ({\it h}) {SPICE}}
		\end{minipage}
	\end{center}
	\caption{Parameter sensitivity of $\lambda_1$ and $\lambda_2$ with CIDEr-D Score Optimization.}\label{fig:f14}
\end{figure*}

\subsection{Sensitivity to Parameters}
The main parameters are the $\lambda_1$ and $\lambda_2$ in Eq. 5 of the main body. We vary the parameters in $\{0.01, 0.1, 1, 10\}$ to study its sensitivity for different performance, and record the results in Figure \ref{fig:f13} and Figure \ref{fig:f14}. We find that CPRC always achieves the best performance with small $\lambda_1$ (i.e., $\lambda_1=0.01$) and large $\lambda_2$ (i.e., $\lambda_2=10$) in terms of all metrics, on both cross-entropy and CIDEr-D score optimization. This phenomenon also validates that the relation consistency loss plays an important role in enhancing the generator.

\begin{table}[!htb]{
		\centering
		\caption{Performance with different ratio data from unsupervised data (i.e., the supervised is fixed with $1\%$) on MS-COCO “Karpathy” test split, where B$@$N, M, R, C and S are short for BLEU@N, METEOR, ROUGE-L, CIDEr-D and SPICE scores. }
		\label{tab:tab11}
		\begin{tabular}{@{}l|c|c|c|c|c|c|c|c}
			\toprule
			\multirow{2}{*}{Methods} & \multicolumn{8}{c}{Cross-Entropy Loss}  \\
			\cmidrule(l){2-9}
			& B$@$1 & B$@$2 & B$@$3 & B$@$4 & M & R & C & S \\
			\midrule
			10$\%$  &68.3&49.5&34.9&23.3&21.4&49.6&71.7&14.6\\
			40$\%$ &66.9&48.7&34.2 &23.4& 22.9&49.6&72.9&15.6\\
			70$\%$ &68.4&50.6&35.6 &24.4& 22.9&\bf 50.5&74.4&15.9\\
			100$\%$ &\bf 68.8&\bf 51.1&\bf 35.7&\bf 24.9&\bf 22.9&50.4&\bf 77.9&\bf 16.2\\
			\midrule
			\multirow{2}{*}{Methods}  & \multicolumn{8}{c}{CIDEr-D Score Optimization} \\
			\cmidrule(l){2-9}
			& B$@$1 & B$@$2 & B$@$3 & B$@$4 & M & R & C & S \\
			\midrule
			10$\%$  &68.7&51.0&25.6&23.9&22.4&50.6&74.1&14.9\\
			40$\%$ &69.2&50.2&35.6 &24.1&22.9&\bf 50.8&75.7&15.9\\
			70$\%$ &69.4&51.3&36.5 &24.8&22.8&50.7&76.5&16.2\\
			100$\%$ &\bf 69.9&\bf 51.8&\bf 36.7&\bf 25.5&\bf 23.4&50.7&\bf 78.8&\bf 16.8\\
			\bottomrule
	\end{tabular}}
\end{table}

\section{Conclusion}\label{sec:s3}
Since traditional image captioning methods are usually working on supervised multi-modal data, in this paper, we investigated how to use undescribed images for semi-supervised image captioning. Specifically, our method can take Cross-modal Prediction and Relation Consistency (CPRC) into consideration.  CPRC employs prediction distillation for the predictions of sentences generated from undescribed images, and develops a novel relation consistency between augmented images and generated sentences to retain the important relational knowledge. As demonstrated by the experiments on the MS-COCO dataset,  CPRC outperforms state-of-the-art methods in various complex semi-supervised scenarios.

\appendices
\section{Influence of Unsupervised Data}
Furthermore, we explore the influence of unsupervised data, i.e., we fix the supervised ratio to 1$\%$, and tune the data ratio from unsupervised data in $\{10\%, 40\%, 70\%, 100\%\}$, the results are recorded in Table \ref{tab:tab11}. We find that with the percentage of unsupervised data increases, the performance of CPRC also improves in terms of all metrics. This indicates that CPRC can make full use of undescribed images for positive training. But the growth rate slows down in the later period (i.e., after $70\%$), probably owing to the interference of pseudo label noise.

%

%

\section*{Acknowledgment}
This research was supported by NSFC (62006118), Natural Science Foundation of Jiangsu Province of China under Grant (BK20200460). NSFC-NRF Joint Research Project under Grant 61861146001, CCF- Baidu Open Fund (CCF-BAIDU OF2020011), Baidu TIC Open Fund.

\ifCLASSOPTIONcaptionsoff
  \newpage
\fi



%
%
%
\bibliographystyle{IEEEtranN}\small
\bibliography{pcrc}

\begin{thebibliography}{46}
\providecommand{\natexlab}[1]{#1}
\providecommand{\url}[1]{#1}
\csname url@samestyle\endcsname
\providecommand{\newblock}{\relax}
\providecommand{\bibinfo}[2]{#2}
\providecommand{\BIBentrySTDinterwordspacing}{\spaceskip=0pt\relax}
\providecommand{\BIBentryALTinterwordstretchfactor}{4}
\providecommand{\BIBentryALTinterwordspacing}{\spaceskip=\fontdimen2\font plus
\BIBentryALTinterwordstretchfactor\fontdimen3\font minus
  \fontdimen4\font\relax}
\providecommand{\BIBforeignlanguage}[2]{{%
\expandafter\ifx\csname l@#1\endcsname\relax
\typeout{** WARNING: IEEEtranN.bst: No hyphenation pattern has been}%
\typeout{** loaded for the language `#1'. Using the pattern for}%
\typeout{** the default language instead.}%
\else
\language=\csname l@#1\endcsname
\fi
#2}}
\providecommand{\BIBdecl}{\relax}
\BIBdecl

\bibitem[Baltrusaitis et~al.(2019)Baltrusaitis, Ahuja, and
  Morency]{BaltrusaitisAM19}
T.~Baltrusaitis, C.~Ahuja, and L.~Morency, ``Multimodal machine learning: {A}
  survey and taxonomy,'' \emph{{IEEE} TPAMI}, vol.~41, no.~2, pp. 423--443,
  2019.

\bibitem[Debie et~al.(2021)Debie, Rojas, Fidock, Barlow, Kasmarik, Anavatti,
  Garratt, and Abbass]{DebieRFBKAGA21}
E.~S. Debie, R.~F. Rojas, J.~Fidock, M.~Barlow, K.~Kasmarik, S.~G. Anavatti,
  M.~Garratt, and H.~A. Abbass, ``Multimodal fusion for objective assessment of
  cognitive workload: {A} review,'' \emph{{IEEE} Trans. Cybern.}, vol.~51,
  no.~3, pp. 1542--1555, 2021.

\bibitem[Karpathy and Li(2015)]{KarpathyL15}
A.~Karpathy and F.~Li, ``Deep visual-semantic alignments for generating image
  descriptions,'' in \emph{CVPR}, 2015, pp. 3128--3137.

\bibitem[Xu et~al.(2015)Xu, Ba, Kiros, Cho, Courville, Salakhutdinov, Zemel,
  and Bengio]{XuBKCCSZB15}
K.~Xu, J.~Ba, R.~Kiros, K.~Cho, A.~C. Courville, R.~Salakhutdinov, R.~S. Zemel,
  and Y.~Bengio, ``Show, attend and tell: Neural image caption generation with
  visual attention,'' in \emph{ICML}, 2015, pp. 2048--2057.

\bibitem[Sammani and Elsayed(2019)]{SammaniE19}
F.~Sammani and M.~Elsayed, ``Look and modify: Modification networks for image
  captioning,'' in \emph{BMVC}, 2019, p.~75.

\bibitem[Bin et~al.(2019)Bin, Yang, Shen, Xie, Shen, and Li]{BinYSXSL19}
Y.~Bin, Y.~Yang, F.~Shen, N.~Xie, H.~T. Shen, and X.~Li, ``Describing video
  with attention-based bidirectional {LSTM},'' \emph{{IEEE} Trans. Cybern.},
  vol.~49, no.~7, pp. 2631--2641, 2019.

\bibitem[Yang et~al.(2016)Yang, Yuan, Wu, Cohen, and Salakhutdinov]{YangYWCS16}
Z.~Yang, Y.~Yuan, Y.~Wu, W.~W. Cohen, and R.~Salakhutdinov, ``Review networks
  for caption generation,'' in \emph{NeurIPS}, 2016, pp. 2361--2369.

\bibitem[Lu et~al.(2017)Lu, Xiong, Parikh, and Socher]{LuXPS17}
J.~Lu, C.~Xiong, D.~Parikh, and R.~Socher, ``Knowing when to look: Adaptive
  attention via a visual sentinel for image captioning,'' in \emph{CVPR}, 2017,
  pp. 3242--3250.

\bibitem[Huang et~al.(2019{\natexlab{a}})Huang, Wang, Chen, and
  Wei]{HuangWCW19}
L.~Huang, W.~Wang, J.~Chen, and X.~Wei, ``Attention on attention for image
  captioning,'' in \emph{ICCV}, 2019, pp. 4633--4642.

\bibitem[Hashimoto et~al.(2018)Hashimoto, Guu, Oren, and Liang]{HashimotoGOL18}
T.~B. Hashimoto, K.~Guu, Y.~Oren, and P.~Liang, ``A retrieve-and-edit framework
  for predicting structured outputs,'' in \emph{NeurIPS}, 2018, pp.
  10\,073--10\,083.

\bibitem[Feng et~al.(2019)Feng, Ma, Liu, and Luo]{Feng00L19a}
Y.~Feng, L.~Ma, W.~Liu, and J.~Luo, ``Unsupervised image captioning,'' in
  \emph{CVPR}, Long Beach, CA, 2019, pp. 4125--4134.

\bibitem[Gu et~al.(2019)Gu, Joty, Cai, Zhao, Yang, and Wang]{GuJCZYW19}
J.~Gu, S.~R. Joty, J.~Cai, H.~Zhao, X.~Yang, and G.~Wang, ``Unpaired image
  captioning via scene graph alignments,'' in \emph{ICCV}, Seoul, Korea, 2019,
  pp. 10\,322--10\,331.

\bibitem[Goodfellow et~al.(2014)Goodfellow, Pouget{-}Abadie, Mirza, Xu,
  Warde{-}Farley, Ozair, Courville, and Bengio]{GoodfellowPMXWOCB14}
I.~J. Goodfellow, J.~Pouget{-}Abadie, M.~Mirza, B.~Xu, D.~Warde{-}Farley,
  S.~Ozair, A.~C. Courville, and Y.~Bengio, ``Generative adversarial nets,'' in
  \emph{NeurIPS}, Montreal, Canada, 2014, pp. 2672--2680.

\bibitem[Mithun et~al.(2018)Mithun, Panda, Papalexakis, and
  Roy{-}Chowdhury]{MithunPPR18}
N.~C. Mithun, R.~Panda, E.~E. Papalexakis, and A.~K. Roy{-}Chowdhury, ``Webly
  supervised joint embedding for cross-modal image-text retrieval,'' in
  \emph{ACMMM}, 2018, pp. 1856--1864.

\bibitem[Huang et~al.(2019{\natexlab{b}})Huang, Kang, Liu, Chang, and
  Hauptmann]{HuangKLCH19}
P.~Huang, G.~Kang, W.~Liu, X.~Chang, and A.~G. Hauptmann, ``Annotation
  efficient cross-modal retrieval with adversarial attentive alignment,'' in
  \emph{ACMMM}, 2019, pp. 1758--1767.

\bibitem[Park et~al.(2019)Park, Kim, Lu, and Cho]{ParkKLC19}
W.~Park, D.~Kim, Y.~Lu, and M.~Cho, ``Relational knowledge distillation,'' in
  \emph{CVPR}, 2019, pp. 3967--3976.

\bibitem[Rennie et~al.(2017)Rennie, Marcheret, Mroueh, Ross, and
  Goel]{RennieMMRG17}
S.~J. Rennie, E.~Marcheret, Y.~Mroueh, J.~Ross, and V.~Goel, ``Self-critical
  sequence training for image captioning,'' in \emph{CVPR}, 2017, pp.
  1179--1195.

\bibitem[Zhou et~al.(2020)Zhou, Wang, Liu, Hu, and Zhang]{ZhouWLHZ20}
Y.~Zhou, M.~Wang, D.~Liu, Z.~Hu, and H.~Zhang, ``More grounded image captioning
  by distilling image-text matching model,'' in \emph{CVPR}, 2020, pp.
  4776--4785.

\bibitem[Yao et~al.(2010)Yao, Yang, Lin, Lee, and Zhu]{YaoYLLZ10}
B.~Z. Yao, X.~Yang, L.~Lin, M.~W. Lee, and S.~C. Zhu, ``{I2T:} image parsing to
  text description,'' \emph{Proceedings of the {IEEE}}, vol.~98, no.~8, pp.
  1485--1508, 2010.

\bibitem[Cho et~al.(2014)Cho, van Merrienboer, cCaglar Gulccehre, Bahdanau,
  Bougares, Schwenk, and Bengio]{ChoMGBBSB14}
K.~Cho, B.~van Merrienboer, cCaglar Gulccehre, D.~Bahdanau, F.~Bougares,
  H.~Schwenk, and Y.~Bengio, ``Learning phrase representations using {RNN}
  encoder-decoder for statistical machine translation,'' in \emph{EMNLP}, 2014,
  pp. 1724--1734.

\bibitem[Vinyals et~al.(2015)Vinyals, Toshev, Bengio, and Erhan]{VinyalsTBE15}
O.~Vinyals, A.~Toshev, S.~Bengio, and D.~Erhan, ``Show and tell: {A} neural
  image caption generator,'' in \emph{CVPR}, 2015, pp. 3156--3164.

\bibitem[Grandvalet and Bengio(2004)]{GrandvaletB04}
Y.~Grandvalet and Y.~Bengio, ``Semi-supervised learning by entropy
  minimization,'' in \emph{NeurIPS}, 2004, pp. 529--536.

\bibitem[Arazo et~al.(2020)Arazo, Ortego, Albert, O'Connor, and
  McGuinness]{ArazoOAOM20}
E.~Arazo, D.~Ortego, P.~Albert, N.~E. O'Connor, and K.~McGuinness,
  ``Pseudo-labeling and confirmation bias in deep semi-supervised learning,''
  in \emph{IJCNN}, 2020, pp. 1--8.

\bibitem[Bachman et~al.(2014)Bachman, Alsharif, and Precup]{BachmanAP14}
P.~Bachman, O.~Alsharif, and D.~Precup, ``Learning with pseudo-ensembles,'' in
  \emph{NeurIPS}, 2014, pp. 3365--3373.

\bibitem[Tarvainen and Valpola(2017)]{TarvainenV17}
A.~Tarvainen and H.~Valpola, ``Mean teachers are better role models:
  Weight-averaged consistency targets improve semi-supervised deep learning
  results,'' in \emph{NeurIPS}, Long Beach, CA, 2017, pp. 1195--1204.

\bibitem[Laine and Aila(2017)]{LaineA17}
S.~Laine and T.~Aila, ``Temporal ensembling for semi-supervised learning,'' in
  \emph{ICLR}, Toulon, France, 2017.

\bibitem[Xie et~al.(2020)Xie, Dai, Hovy, Luong, and Le]{XieDHL020}
Q.~Xie, Z.~Dai, E.~H. Hovy, T.~Luong, and Q.~Le, ``Unsupervised data
  augmentation for consistency training,'' in \emph{NeurIPS}, 2020.

\bibitem[Berthelot et~al.(2020)Berthelot, Carlini, Cubuk, Kurakin, Sohn, Zhang,
  and Raffel]{BerthelotCCKSZR20}
D.~Berthelot, N.~Carlini, E.~D. Cubuk, A.~Kurakin, K.~Sohn, H.~Zhang, and
  C.~Raffel, ``Remixmatch: Semi-supervised learning with distribution matching
  and augmentation anchoring,'' in \emph{ICLR}, 2020.

\bibitem[French et~al.(2018)French, Mackiewicz, and Fisher]{FrenchMF18}
G.~French, M.~Mackiewicz, and M.~H. Fisher, ``Self-ensembling for visual domain
  adaptation,'' in \emph{ICLR}, 2018.

\bibitem[Sohn et~al.(2020)Sohn, Berthelot, Li, Zhang, Carlini, Cubuk, Kurakin,
  Zhang, and Raffel]{Kihyuk2020}
K.~Sohn, D.~Berthelot, C.~Li, Z.~Zhang, N.~Carlini, E.~D. Cubuk, A.~Kurakin,
  H.~Zhang, and C.~Raffel, ``Fixmatch: Simplifying semi-supervised learning
  with consistency and confidence,'' \emph{CoRR}, vol. abs/2001.07685, 2020.

\bibitem[He et~al.(2016)He, Zhang, Ren, and Sun]{HeZRS16}
K.~He, X.~Zhang, S.~Ren, and J.~Sun, ``Deep residual learning for image
  recognition,'' in \emph{Proceedings of the 2016 {IEEE} Conference on Computer
  Vision and Pattern Recognition}, Las Vegas, NV, 2016, pp. 770--778.

\bibitem[Ren et~al.(2017)Ren, He, Girshick, and Sun]{RenHG017}
S.~Ren, K.~He, R.~B. Girshick, and J.~Sun, ``Faster {R-CNN:} towards real-time
  object detection with region proposal networks,'' \emph{{IEEE} Trans. Pattern
  Anal. Mach. Intell.}, vol.~39, no.~6, pp. 1137--1149, 2017.

\bibitem[Bahdanau et~al.(2015)Bahdanau, Cho, and Bengio]{BahdanauCB14}
D.~Bahdanau, K.~Cho, and Y.~Bengio, ``Neural machine translation by jointly
  learning to align and translate,'' in \emph{ICLR}, San Diego, CA, 2015.

\bibitem[Vedantam et~al.(2015)Vedantam, Zitnick, and Parikh]{VedantamZP15}
R.~Vedantam, C.~L. Zitnick, and D.~Parikh, ``Cider: Consensus-based image
  description evaluation,'' in \emph{CVPR}, 2015, pp. 4566--4575.

\bibitem[Ranzato et~al.(2016)Ranzato, Chopra, Auli, and Zaremba]{RanzatoCAZ15}
M.~Ranzato, S.~Chopra, M.~Auli, and W.~Zaremba, ``Sequence level training with
  recurrent neural networks,'' in \emph{ICLR}, Y.~Bengio and Y.~LeCun, Eds.,
  San Juan, Puerto Rico, 2016.

\bibitem[Cubuk et~al.(2018)Cubuk, Zoph, Man{\'{e}}, Vasudevan, and Le]{Cubuk20}
E.~D. Cubuk, B.~Zoph, D.~Man{\'{e}}, V.~Vasudevan, and Q.~V. Le, ``Autoaugment:
  Learning augmentation policies from data,'' \emph{CoRR}, vol. abs/1805.09501,
  2018.

\bibitem[Lin et~al.(2021)Lin, Wang, Chang, and Sun]{LinWCS21}
Y.~Lin, C.~Wang, C.~Chang, and H.~Sun, ``An efficient framework for counting
  pedestrians crossing a line using low-cost devices: the benefits of
  distilling the knowledge in a neural network,'' \emph{Multim. Tools Appl.},
  vol.~80, no.~3, pp. 4037--4051, 2021.

\bibitem[Matthews(2001)]{Matthews2001A}
P.~Matthews, ``A short history of structural linguistics,'' 2001.

\bibitem[Lin et~al.(2014)Lin, Maire, Belongie, Hays, Perona, Ramanan, Dollar,
  and Zitnick]{LinMBHPRDZ14}
T.~Lin, M.~Maire, S.~J. Belongie, J.~Hays, P.~Perona, D.~Ramanan, P.~Dollar,
  and C.~L. Zitnick, ``Microsoft coco: Common objects in context,'' in
  \emph{ECCV}, 2014, pp. 740--755.

\bibitem[Huang et~al.(2019{\natexlab{c}})Huang, Wang, Xia, and
  Chen]{HuangWXC19}
L.~Huang, W.~Wang, Y.~Xia, and J.~Chen, ``Adaptively aligned image captioning
  via adaptive attention time,'' in \emph{NeurIPS}, 2019, pp. 8940--8949.

\bibitem[Herdade et~al.(2019)Herdade, Kappeler, Boakye, and
  Soares]{HerdadeKBS19}
S.~Herdade, A.~Kappeler, K.~Boakye, and J.~Soares, ``Image captioning:
  Transforming objects into words,'' in \emph{NeurIPS}, 2019, pp.
  11\,135--11\,145.

\bibitem[Karpathy and Fei{-}Fei(2017)]{KarpathyF17}
A.~Karpathy and L.~Fei{-}Fei, ``Deep visual-semantic alignments for generating
  image descriptions,'' \emph{TPAMI}, vol.~39, no.~4, pp. 664--676, 2017.

\bibitem[Kingma and Ba(2015)]{KingmaB14}
D.~P. Kingma and J.~Ba, ``Adam: {A} method for stochastic optimization,'' in
  \emph{ICLR}, 2015.

\bibitem[Papineni et~al.(2002)Papineni, Roukos, Ward, and Zhu]{PapineniRWZ02}
K.~Papineni, S.~Roukos, T.~Ward, and W.~Zhu, ``Bleu: a method for automatic
  evaluation of machine translation,'' in \emph{ACL}, 2002, pp. 311--318.

\bibitem[Banerjee and Lavie(2005)]{BanerjeeL05}
S.~Banerjee and A.~Lavie, ``{METEOR:} an automatic metric for {MT} evaluation
  with improved correlation with human judgments,'' in \emph{IEEMMT}, 2005, pp.
  65--72.

\bibitem[Anderson et~al.(2016)Anderson, Fernando, Johnson, and
  Gould]{AndersonFJG16}
P.~Anderson, B.~Fernando, M.~Johnson, and S.~Gould, ``{SPICE:} semantic
  propositional image caption evaluation,'' in \emph{ECCV}, 2016, pp. 382--398.

\end{thebibliography}

%

\vspace{-1.2cm}
\begin{IEEEbiography}[{\includegraphics[width=1in,height=1.25in,clip,keepaspectratio]{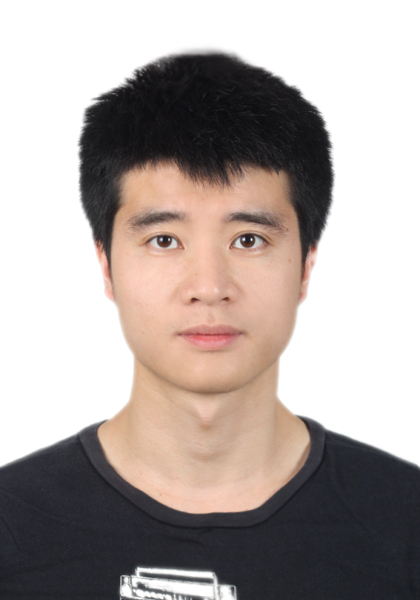}}]{Yang Yang}
	received the Ph.D. degree in computer science, Nanjing University, China in 2019. At the same year, he became a faculty member at Nanjing University of Science and Technology, China. He is currently a Professor with the School of Computer Science and Engineering. His research interests lie primarily in machine learning and data mining, including heterogeneous learning, model reuse, and incremental mining. He has published over 10 papers in leading international journal/conferences. He serves as PC in leading conferences such as IJCAI, AAAI, ICML, NIPS, etc.
\end{IEEEbiography}
\vspace{-1.2cm}
\begin{IEEEbiography}[{\includegraphics[width=1in,height=1.25in,clip,keepaspectratio]{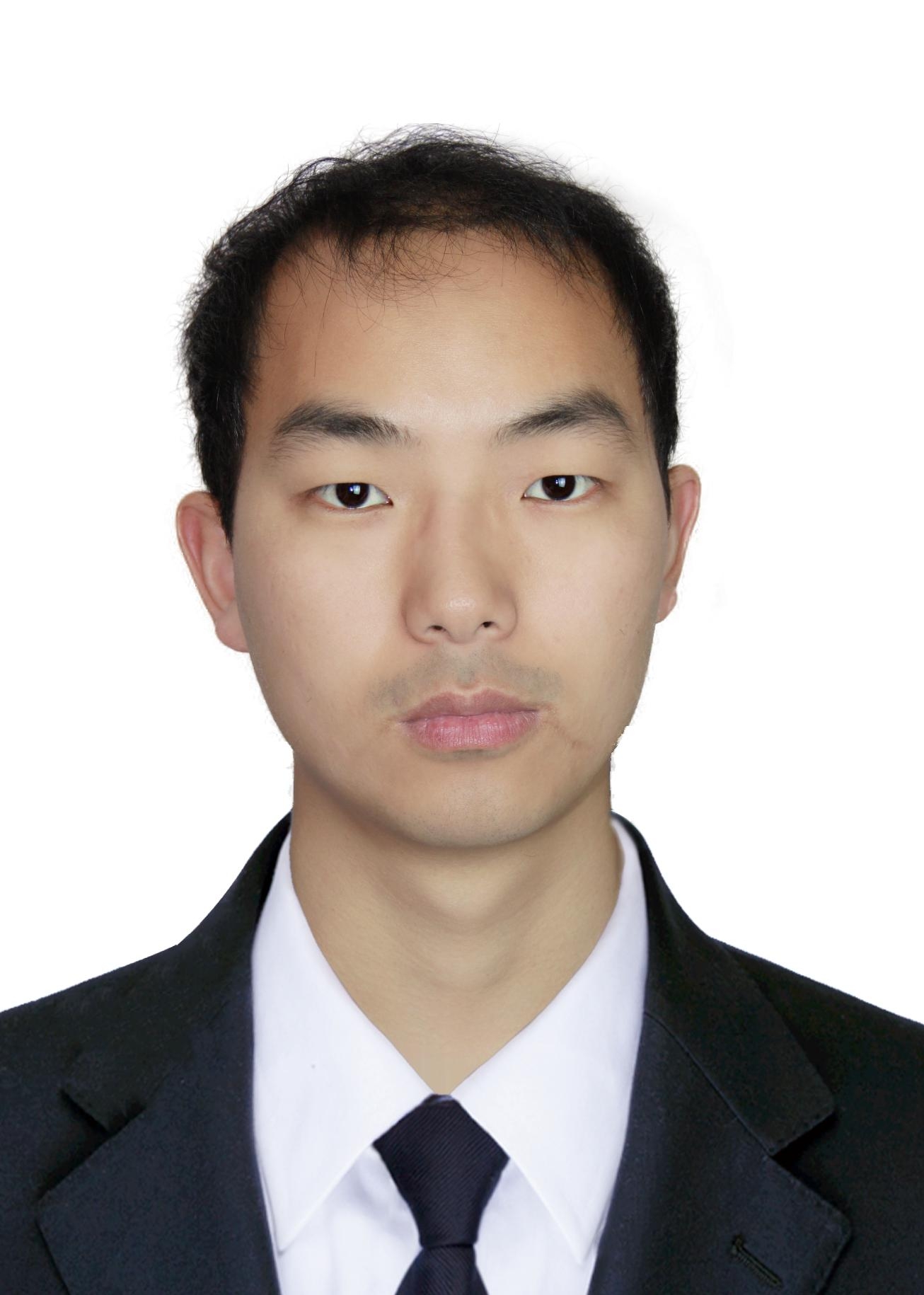}}]{Hongchen Wei}
	is working towards the M.Sc. degree with the School of Computer Science and Engineering, in Nanjing University of Science and Technology, China. His research interests lie primarily in machine learning and data mining, including cross-modal learning.
\end{IEEEbiography}
\vspace{-1.2cm}
\begin{IEEEbiography}[{\includegraphics[width=1in,height=1.25in,clip,keepaspectratio]{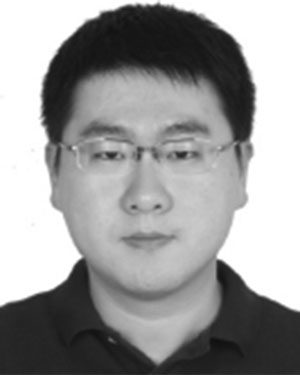}}]{Hengshu Zhu}(SM’19)
	is currently a principal data scientist $\&$ architect at Baidu Inc. He received the Ph.D. degree in 2014 and B.E. degree in 2009, both in Computer Science from University of Science and Technology of China (USTC), China. His general area of research is data mining and machine learning, with a focus on developing advanced data analysis techniques for innovative business applications. He has published prolifically in refereed journals and conference proceedings, including IEEE Transactions on Knowledge and Data Engineering (TKDE), IEEE Transactions on Mobile Computing (TMC), ACM Transactions on Information Systems (ACM TOIS), ACM Transactions on Knowledge Discovery from Data (TKDD), ACM SIGKDD, ACM SIGIR, WWW, IJCAI, and AAAI. He has served regularly on the organization and program committees of numerous conferences, including as a program co-chair of the KDD Cup-2019 Regular ML Track, and a founding co-chair of the first International Workshop on Organizational Behavior and Talent Analytics (OBTA) and the International Workshop on Talent and Management Computing (TMC), in conjunction with ACM SIGKDD. He was the recipient of the Distinguished Dissertation Award of CAS (2016), the Distinguished Dissertation Award of CAAI (2016), the Special Prize of President Scholarship for Postgraduate Students of CAS (2014), the Best Student Paper Award of KSEM-2011, WAIM-2013, CCDM-2014, and the Best Paper Nomination of ICDM-2014. He is the senior member of IEEE, ACM, and CCF. 
\end{IEEEbiography}
\vspace{-1.2cm}
\begin{IEEEbiography}[{\includegraphics[width=1in,height=1.25in,clip,keepaspectratio]{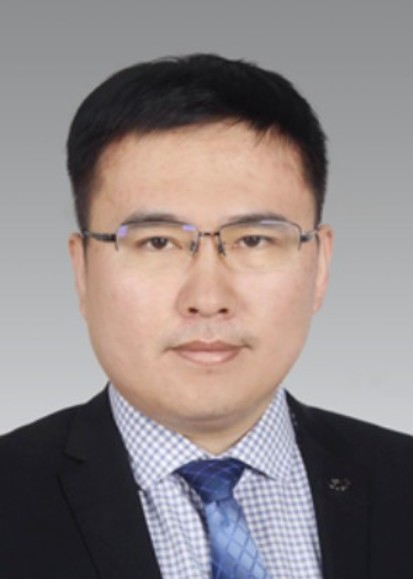}}]{Dianhai Yu}
is currently a chief machine learning architect at Baidu Inc. and in charge of the open-source deep learning platform PaddlePaddle. He joined Baidu in 2008 after graduating from Peking University. His research interests lie primarily in machine learning and natural language processing, including machine learning systems, scalable distributed deep learning, semantic computing, and human-machine dialogue systems. He has published over 10 papers in the field of Artificial Intelligence. He is the senior member of CCF. In 2019, He received the CCF Outstanding Engineer Award.
\end{IEEEbiography}
\vspace{-1.2cm}
\begin{IEEEbiography}[{\includegraphics[width=1in,height=1.25in,clip,keepaspectratio]{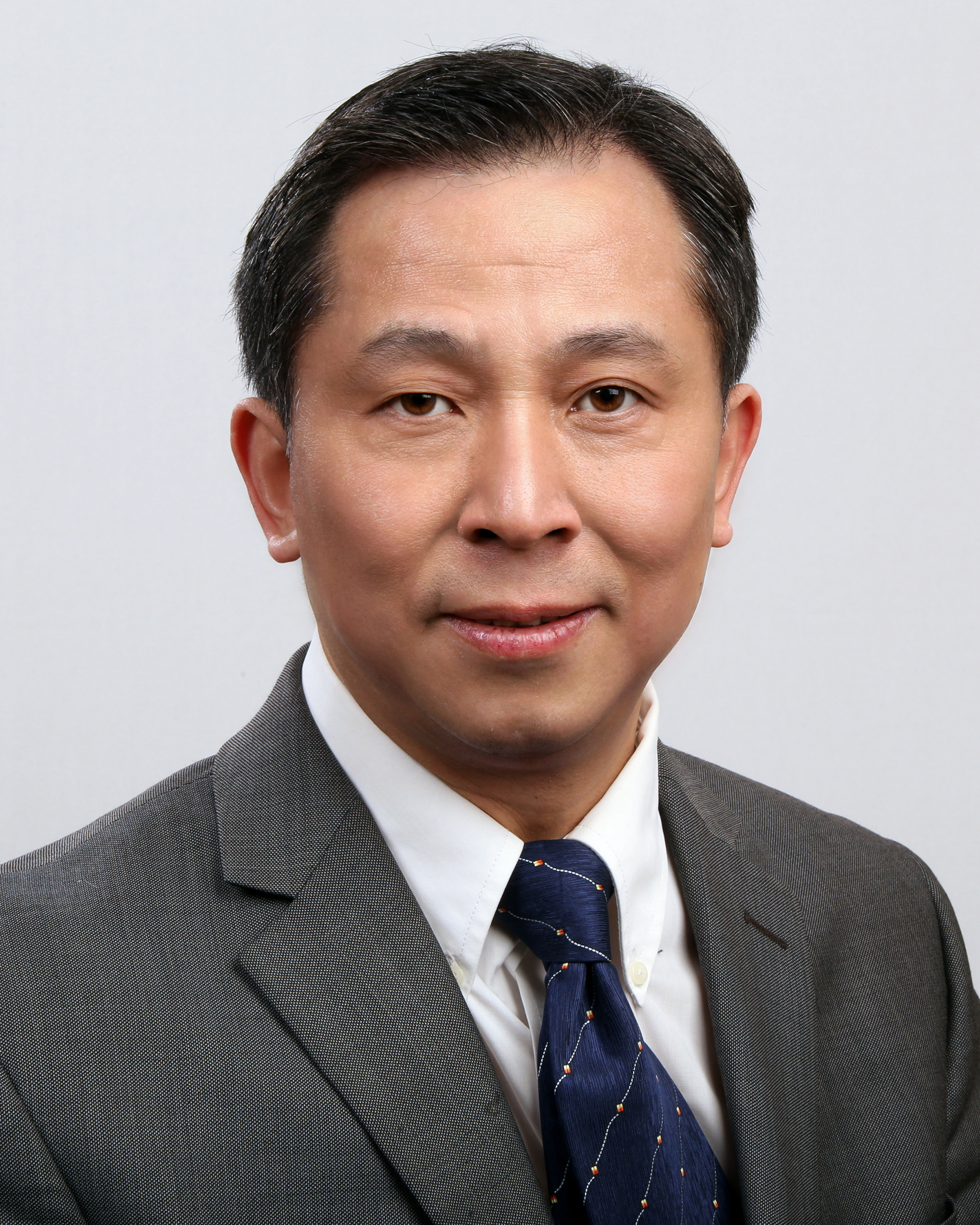}}]{Hui Xiong} (IEEE Fellow)
	is currently a Full Professor at the Rutgers, the State University of New Jersey, where he received the 2018 Ram Charan Management Practice Award as the Grand Prix winner from the Harvard Business Review, RBS Dean’s Research Professorship (2016), the Rutgers University Board of Trustees Research Fellowship for Scholarly Excellence (2009), the ICDM Best Research Paper Award (2011), and the IEEE ICDM Outstanding Service Award (2017). He received the Ph.D. degree from the University of Minnesota (UMN), USA. He is a co-Editor-in-Chief of Encyclopedia of GIS, an Associate Editor of IEEE Transactions on Big Data (TBD), ACM Transactions on Knowledge Discovery from Data (TKDD), and ACM Transactions on Management Information Systems (TMIS). He has served regularly on the organization and program committees of numerous conferences, including as a Program Co-Chair of the Industrial and Government Track for the 18th ACM SIGKDD International Conference on Knowledge Discovery and Data Mining (KDD), a Program Co-Chair for the IEEE 2013 International Conference on Data Mining (ICDM), a General Co-Chair for the IEEE 2015 International Conference on Data Mining (ICDM), and a Program Co-Chair of the Research Track for the 2018 ACM SIGKDD International Conference on Knowledge Discovery and Data Mining. He is an IEEE Fellow and an ACM Distinguished Scientist.
\end{IEEEbiography}
\vspace{-1.2cm}
\begin{IEEEbiography}[{\includegraphics[width=1in,height=1.25in,clip,keepaspectratio]{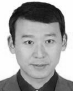}}]{Jian Yang} (M'08)
	received the Ph.D. degree in pattern recognition and intelligence systems from the Nanjing University of Science and Technology 	(NUST), Nanjing, China, in 2002. In 2003, he was a Post-Doctoral Researcher with the University of Zaragoza, Zaragoza, Spain. From 	2004 to 2006, he was a Post-Doctoral Fellow with the Biometrics Centre, The Hong Kong Polytechnic University, Hong Kong. From 2006 to 2007, he was a Post-Doctoral Fellow with the Department of Computer Science, New Jersey Institute of Technology, Newark, NJ, USA. He is currently a Chang-Jiang Professor with the School of Computer Science and Engineering, NUST. He has authored more than 200 scientific papers in pattern recognition and computer vision. His papers have been cited more than 6000 times in the Web of Science and 15,000 times in the Scholar Google. His current research interests include pattern recognition, computer vision, and machine learning. Dr. Yang is a Fellow of IAPR. He is currently an Associate Editor of Pattern Recognition, Pattern Recognition Letters, the IEEE TRANSACTIONS ON NEURAL NETWORKS AND LEARNING SYSTEMS, and Neurocomputing.
\end{IEEEbiography}

\end{document}